\documentclass[preprint,12pt]{elsarticle}
\usepackage{amssymb}    
\usepackage{amsmath}    

\usepackage[colorlinks=true, linkcolor=blue, urlcolor=blue, citecolor=blue]{hyperref}

\usepackage{graphicx}
\usepackage{subcaption}

\usepackage[T1]{fontenc}

\usepackage{diffcoeff} 

\usepackage{cleveref}

\usepackage{booktabs}  
\usepackage{colortbl}  
\usepackage{xcolor}    

\usepackage{circuitikz, tikz} 
\usetikzlibrary{shapes}  

\journal{Computer Methods in Applied Mechanics and Engineering }
\begin{document}

\begin{frontmatter}

\title{eXtended Physics Informed Neural Network Method for Fracture Mechanics Problems}

\author{A. Lotfalian \corref{cor1}}
\ead{Lotfalian9930116@shirazu.ac.ir}
\cortext[cor1]{Corresponding author}
\author{M. R. Banan}
\author{P. Broumand}

\affiliation{organization={Department of Civil and Environmental Engineering, Shiraz University, Shiraz, Iran}}

\begin{abstract}
This paper presents eXtended Physics-Informed Neural Network (X-PINN), a novel and robust framework for addressing fracture mechanics problems involving multiple cracks in fractured media. To address this, an energy-based loss function, customized integration schemes, and domain decomposition procedures are proposed. Inspired by the Extended Finite Element Method (XFEM), the neural network solution space is enriched with specialized functions that allow crack body discontinuities and singularities at crack tips to be explicitly captured. Furthermore, a structured framework is introduced in which standard and enriched solution components are modeled using distinct neural networks, enabling flexible and effective simulations of complex multiple-crack problems in 1D and 2D domains, with convenient extensibility to 3D problems. Numerical experiments are conducted to validate the effectiveness and robustness of the proposed method.
\end{abstract}

\begin{keyword}
Fracture Mechanics, PINN, Enrichment, Energy Based Loss, XFEM

\begin{highlights}
\item An extended Physics-Informed Neural Network (PINN) framework is introduced for modeling multiple cracks in solid media.
\item An energy-based loss function and customized integration schemes are incorporated to enhance crack modeling accuracy.
\item The PINN solution space is enriched with specialized functions to capture discontinuities of the crack bodies and singularities at crack tips.
\item Validation is conducted through numerical experiments in 1D and 2D domains, showcasing robustness for future applications.
\end{highlights}
\end{keyword}
\end{frontmatter} 

\section{Introduction} \label{Introduction}
Fracture and damage mechanics are essential tools in engineering, providing the foundation for predicting material failure and designing safer and more resilient structures. The presence of cracks creates significant challenges in predicting material behavior, necessitating advanced approaches for material selection and the implementation of rigorous safety measures to ensure structural integrity. Numerical methods play a crucial role in this process, offering powerful tools to simulate fracture mechanisms, predict crack propagation, and assess damage evolution. These computational approaches reduce reliance on costly and time-consuming experimental testing while improving the accuracy of failure predictions.

The Extended Finite Element Method (XFEM), pioneered by Belytschko and his team in 1999 \cite{belytschko1999elastic, moes1999finite, daux2000arbitrary, dolbow2000discontinuous, dolbow2001extended}, has emerged as a powerful technique for modeling fracture and damage problems. Built on the Partition of Unity (PU) framework \cite{melenk1996partition}, XFEM allows for the local application of enrichment functions, enabling the accurate representation of discontinuities without requiring mesh refinement. This adaptability makes XFEM particularly effective for simulating cracks, interfaces, and evolving material damage across various engineering applications  \cite{khoei2015extended, rabczuk2019extended}.

Over the years, the Extended Finite Element Method (XFEM) has been successfully applied to a wide variety of complex problems across multiple fields. In Linear Elastic Fracture Mechanics (LEFM)\cite{gravouil2002non,sukumar2008three}, Cohesive Fracture Mechanics \cite{moes2002extended,zi2003new}, composite materials \cite{sukumar2004partition,gracie2009concurrently,akhondzadeh2017efficient}, plasticity and damage \cite{broumand2013extended, broumand2015x}, fatigue \cite{chopp2003fatigue}, contact mechanics \cite{liu2008contact, broumand2018general}, porous media and hydraulic fracturing \cite{khoei2014mesh, jafari2022extended, jafari2023extended}, structural health monitoring and crack detection \cite{rabinovich2007xfem, agathos2018multiple, broumand2021inverse}, XFEM has demonstrated remarkable versatility and effectiveness. While several studies have notably advanced its application in these areas, the examples mentioned represent only a portion of the extensive research conducted. 

In recent years, the utilization of machine learning (ML) methodologies grounded in artificial neural networks (ANNs) has experienced a marked increase, particularly in data-intensive domains such as text, image, and audio processing, where these techniques have yielded exceptional results, significantly outperforming the previous state-of-the-art algorithms \cite{goodfellow2016deep, lecun2015deep}. Within the scientific community, there has been a growing emphasis on leveraging the recent advancements in Artificial Neural Networks (ANNs) and Machine Learning (ML) to address the numerical solutions of partial differential equations (PDEs) and other engineering problems of practical significance.

Physics-informed neural networks (PINNs) leverage the universal approximation capability of Neural Networks (NNs) \cite{hornik1989multilayer}, which enables them to approximate any continuous function. This feature allows neural networks to serve as a basis for deriving solutions to PDEs. The key idea behind PINNs is to augment the neural network training process with the underlying physical constraints of the problem, rather than solely relying on data. By incorporating the governing PDEs or other physical laws as additional loss terms in the network's objective function, PINNs ensure that the network's predictions satisfy the physical constraints. This approach enables the network to learn the intrinsic physical properties and behaviors of the system, making it more accurate and generalizable compared to traditional data-driven methods.

The origins of PINNs trace back to the pioneering work of Lagaris et al. in the 1990s, where neural networks were first employed to solve differential equations \cite{lagaris1998artificial, lagaris1997artificial}. This early research was later extended to irregular domains \cite{lagaris2000neural}. With advancements in computational power, PINNs gained renewed interest, notably through the contributions of Sirignano and Spiliopoulos \cite{sirignano2018dgm} and Raissi et al. \cite{raissi2019physics}.

PINNs have since been applied to various engineering problems. Raissi et al. \cite{raissi2020hidden} introduced Hidden Fluid Mechanics (HFM), integrating the Navier-Stokes equations into neural networks for flow visualization. In solid mechanics, Haghighat et al. \cite{haghighat2021physics} incorporated momentum balance and constitutive relations into PINNs, extending applications from linear elasticity to von Mises elastoplasticity. Jeong et al. \cite{jeong2023complete} developed CPINNTO for topology optimization, leveraging Deep Energy Method (DEM) and sensitivity-analysis PINNs to achieve optimal designs without labeled data or FEA. Cai et al. \cite{cai2021physics} demonstrated PINNs' effectiveness in heat transfer problems, including convection with unknown thermal boundaries and two-phase flow modeling.

PINNs have been extensively applied in fracture and damage mechanics, enabling the modeling of crack propagation and damage evolution without requiring a detailed mesh. Their integration of experimental data, boundary conditions, and physical laws enhances predictive accuracy. Goswami et al. \cite{goswami2020transfer} introduced a PINN algorithm for brittle fracture by minimizing variational energy instead of PDE residuals, ensuring exact boundary condition satisfaction. To address computational limitations in brittle fracture analysis, Goswami et al. \cite{goswami2022physics} developed V-DeepONet, a hybrid approach integrating governing equations with labeled data for efficient predictions.

Zheng et al. \cite{zheng2022physics} proposed a PINN method for quasi-brittle materials that reconstructs the displacement field while maintaining thermodynamic consistency, leveraging transfer learning and domain decomposition for improved convergence. Gu et al. \cite{gu2023enriched} introduced a meshless PINN framework for 2D in-plane crack problems, enhancing standard formulations with crack-tip asymptotic functions to capture local behavior without nodal refinement.

Pantidis and Mobasher \cite{pantidis2023integrated} developed the Integrated Finite Element Neural Network (I-FENN), embedding neural networks into the finite element stiffness function for nonlinear computational mechanics, specifically continuum damage analysis. Eghbalpoor and Sheidaei \cite{eghbalpoor2024peridynamic} combined PINNs with peridynamic theory to improve quasi-static damage and crack propagation predictions. Manav et al. \cite{manav2024phase} proposed a deep Ritz method for modeling complex fracture processes within a phase-field framework, ensuring accurate energy minimization through optimized neural network training.

In the studies reviewed, cracks have often been implicitly modeled, which can overlook the complexities of crack behavior during the damage process. This research aims to develop a novel method, termed X-PINN, that enables explicit consideration of the presence of multiple cracks in the analyses. Such an endeavor poses significant challenges for physics-informed neural networks, as these networks are primarily designed to approximate continuous functions. To this end, we propose an energy-based loss function specifically designed for PINNs, complemented by customized integration schemes, and domain decomposition procedures that enhance crack modeling capabilities. The foundational principles of the XFEM are utilized to enrich the PINN solution space by introducing specialized enrichment functions that address crack body discontinuities and singularities at the crack tips. In addition, a structured framework is outlined in which standard and enriched components are modeled using distinct neural networks, enabling more flexible and effective simulation of complex multiple crack problems in 1D and 2D domains, with convenient extensibility to 3D problems. The Python PyTorch framework is utilized for its flexibility, enabling the effective implementation of the proposed PINN formulation and the neural network architecture. Promising results are obtained for complex fracture problems containing multiple cracks. 

The paper is structured as follows. In Section \ref{theoreticalBackground}, the theoretical foundations of cracked media, the Extended Finite Element Method, and Physics-Informed Neural Networks are reviewed. In Section \ref{proposedMethod}, the proposed framework is introduced and its implementation aspects are discussed. Section \ref{numericalExamples} presents several numerical examples that evaluate the accuracy and robustness of the proposed method. Finally, Section \ref{summary} concludes the study by summarizing the key findings.
\section{Theoretical Background} \label{theoreticalBackground}
This section addresses the linear elasticity problem in solid mechanics as it pertains to fractured media. It will introduce the governing equations for linear elastic solid materials and the equivalent energy form, outline the fundamental concepts of XFEM, and examine the standard formulations employed in PINNs for effectively modeling and solving these governing equations.

\subsection {Governing equations for linear elastic solid mechanics}
\begin{figure}[h!]
    \centering
\includegraphics[scale=0.5]{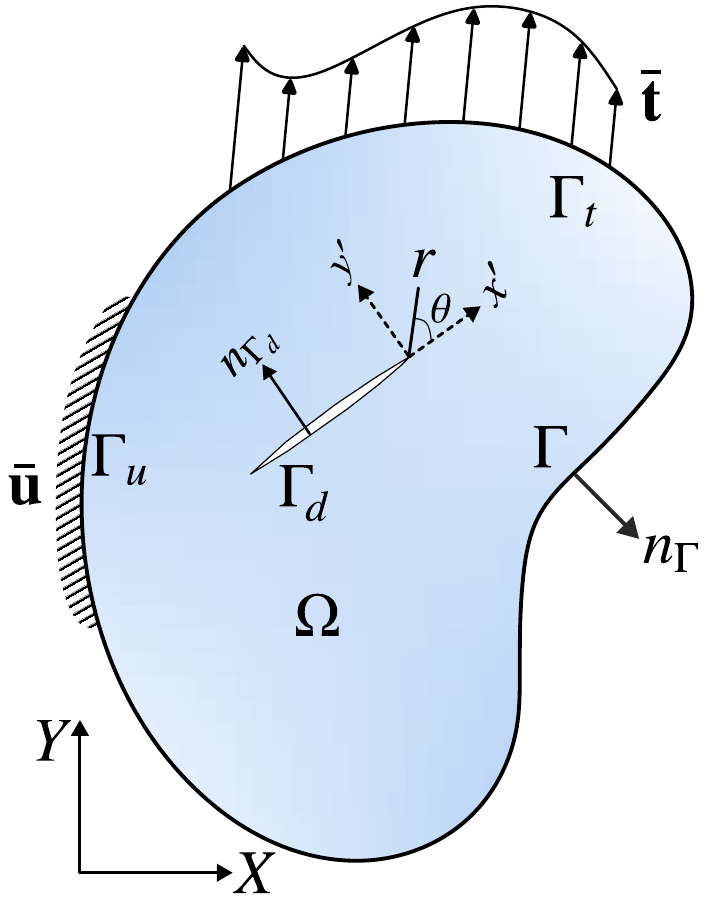}
    \caption{Schematic configuration of the problem domain and boundaries of the fractured media}
    \label{fig:Domain}
\end{figure}

Consider a cracked body $\mathit{\Omega}$ that is bounded by $\mathit{\Gamma = \Gamma_u \cup \Gamma_t}$ and crack surfaces $\mathit{\Gamma_c}$, with $\mathit{\Gamma_u \cap \Gamma_t = \emptyset}$ (Fig.~\hyperref[fig:Domain]{\ref*{fig:Domain}}), the corresponding equilibrium equation can be written in index notation as:
\begin{equation}\label{equilibriumEqu}
    \sigma_{ij,j} + f_i=0 \quad \quad \quad in \quad  
  \Omega, \quad i,j=1,2,3
\end{equation}

\noindent Here, $\sigma_{ij}$ and $f_i$ are the Cauchy stress tensor and the body force vector per unit mass. Under the assumption of infinitely small deformations, the stress tensor is determined by the following relationship:

\begin{equation}\label{sigmaEqu}
\sigma_{ij} = \lambda \delta_{ij} \epsilon_{kk} + 2\mu \epsilon_{ij}
\end{equation}

\begin{equation}\label{epsilonEqu}
\epsilon_{ij} = \frac{1}{2} \left( \frac{\partial u_i}{\partial x_j} + \frac{\partial u_j}{\partial x_i} \right)
\end{equation}
where $\lambda$ and $\mu$ are referred to as Lamé constants, which characterize the elastic properties of the material. The symbol $\delta_{ij}$ denotes the Kronecker delta function, which is used to represent the identity property in tensor operations. Furthermore, $\epsilon$ represents the strain tensor, while $u$ signifies the displacement vector. The governing equations presented above are complemented by Dirichlet and Neumann boundary conditions, which can be expressed as follows:

\begin{subequations}\label{BCs}
\begin{equation}
    u_i = \bar{u_i} \quad \text{on } \Gamma_u \tag{4a} \label{dirichlet}
\end{equation}
\begin{equation}
    \sigma_{ij} n_j = \bar{t_i} \quad \text{on } \Gamma_t \tag{4b} \label{neumann}
\end{equation}
\begin{equation}
    \sigma_{ij} n_j = 0 \quad \text{on } \Gamma_d \tag{4c} \label{crackBC}
\end{equation}
\end{subequations}
where $\bar{u}$ and $\bar{t}$ denote the prescribed displacement and traction boundary conditions, respectively, applied to the boundaries $\Gamma_u$ and $\Gamma_t$. The crack body $\Gamma_d$ is assumed to be traction-free as stated in the third condition.

\subsection{Formulation based on Principle of Minimum Potential Energy}
\label{energyMethod}
In place of the strong-form equations in Eq. \(\eqref{equilibriumEqu}\) with boundary conditions described by Eqs. \(\eqref{dirichlet}\),\(\eqref{neumann}\) and \eqref{crackBC}, the linear elasticity problem can be examined using the Minimum Potential Energy, which states \cite{sadd2009elasticity}: \textit{among all displacements that satisfy the specified boundary conditions, those that also satisfy the equilibrium equations make the potential energy a local minimum}. In mathematical terms:

\begin{equation}\label{totalPotential}
\mathbf{u}^{*} = \arg \min_\mathbf{u} \mathit{\Pi}(\mathbf{u})
\end{equation}
Here, $\mathbf{u}^{*}$ denotes the solution to the problem and $\mathit{\Pi}$ represents the total potential energy defined as follows:
\begin{equation}
\mathit{\Pi} = U + V + T
\end{equation}
where $U$ is the internal energy,
\begin{equation}
U = \frac{1}{2} \int_{\Omega} \boldsymbol{\sigma}(\boldsymbol{\epsilon}) : \boldsymbol{\epsilon} \, dV = \frac{1}{2} \int_{\Omega} \sigma_{ij} \epsilon_{ij}\, dV
\end{equation}
$V$ is the potential energy due to external loads:
\begin{equation}
V = -\int_{\Omega} \mathbf{f} \cdot \mathbf{u} \, dV - \int_{\Gamma_t} \mathbf{\bar{t}} \cdot \mathbf{u} \, dS = -\int_{\Omega} f_i u_i \, dV - \int_{\Gamma_t} \bar{t}_{i} u_i \, dS
\end{equation}
and $T$ acts as a penalty term to enforce the essential boundary conditions weakly as:
\begin{equation}
T = \int_{\Gamma_u} \boldsymbol{\lambda} \cdot |\mathbf{u} - \mathbf{\bar{u}}| \, dS =  \int_{\Gamma_u} \lambda_i |u_i - \bar{u_i}| \, dS
\end{equation}
where $\lambda_i$s denotes the Lagrangian multipliers, each having dimensions of force.

\subsection{Extended Finite Element Method (XFEM)} \label{XFEM}
XFEM enhances the finite element space by introducing enrichment functions designed to capture the behavior of discontinuities, such as cracks or voids. The displacement field, \( \mathbf{u}(\mathbf{x}) \), in XFEM is constituted from three main terms, namely, standard, discontinuity enrichment, and asymptotic singularity enrichment terms, expressed as:
\begin{equation}
    \label{eq:xfem_displacement}
    \mathbf{u}(\mathbf{x}) \approx \underbrace{\sum_{I=1}^{N} \mathbf{N}_I(\mathbf{x}) \mathbf{u}_I}_{\text{Standard FEM term}} + \underbrace{\sum_{J=1}^{N_D} \mathbf{N}_J(\mathbf{x}) \Phi(\mathbf{x}) \boldsymbol{\alpha}_J}_{\text{Discontinuity Enrichment}} + \underbrace{\sum_{K=1}^{N_S} \mathbf{N}_K(\mathbf{x}) \sum_{\beta} \Psi_{\beta}(\mathbf{x}) \boldsymbol{\gamma}_{K,\beta}}_{\text{Asymptotic Singularity Enrichment}}
\end{equation}
where, \(\mathbf{N}_I(\mathbf{x})\), \(\mathbf{N}_J(\mathbf{x})\), and \(\mathbf{N}_K(\mathbf{x})\) are the vector-valued shape functions of the standard FEM; \(N\), \(N_D\) and \(N_S\) are the total number of nodes in the mesh associated with standard FEM,  discontinuity enrichment, and crack tip asymptotic singularity enrichment, respectively. \(\mathbf{u}_I\), \(\boldsymbol{\alpha}_J\), and \(\boldsymbol{\gamma}_{K,\beta}\) are the vectors of degrees of freedom of nodal displacement associated with the three terms, respectively. \(\Phi(\mathbf{x})\) is the enrichment function for discontinuity enrichment, typically a shifted Heaviside step function introducing a displacement jump across the interface, and \(\Psi_{\beta}(\mathbf{x})\) are the crack tip singularity enrichment functions, each identified with an index \(\beta\). For more details on the formulation and implementation, the interested readers are referred to \cite{khoei2015extended,rabczuk2019extended}.

\begin{figure}[h!]
    \centering
\includegraphics[scale=0.27]{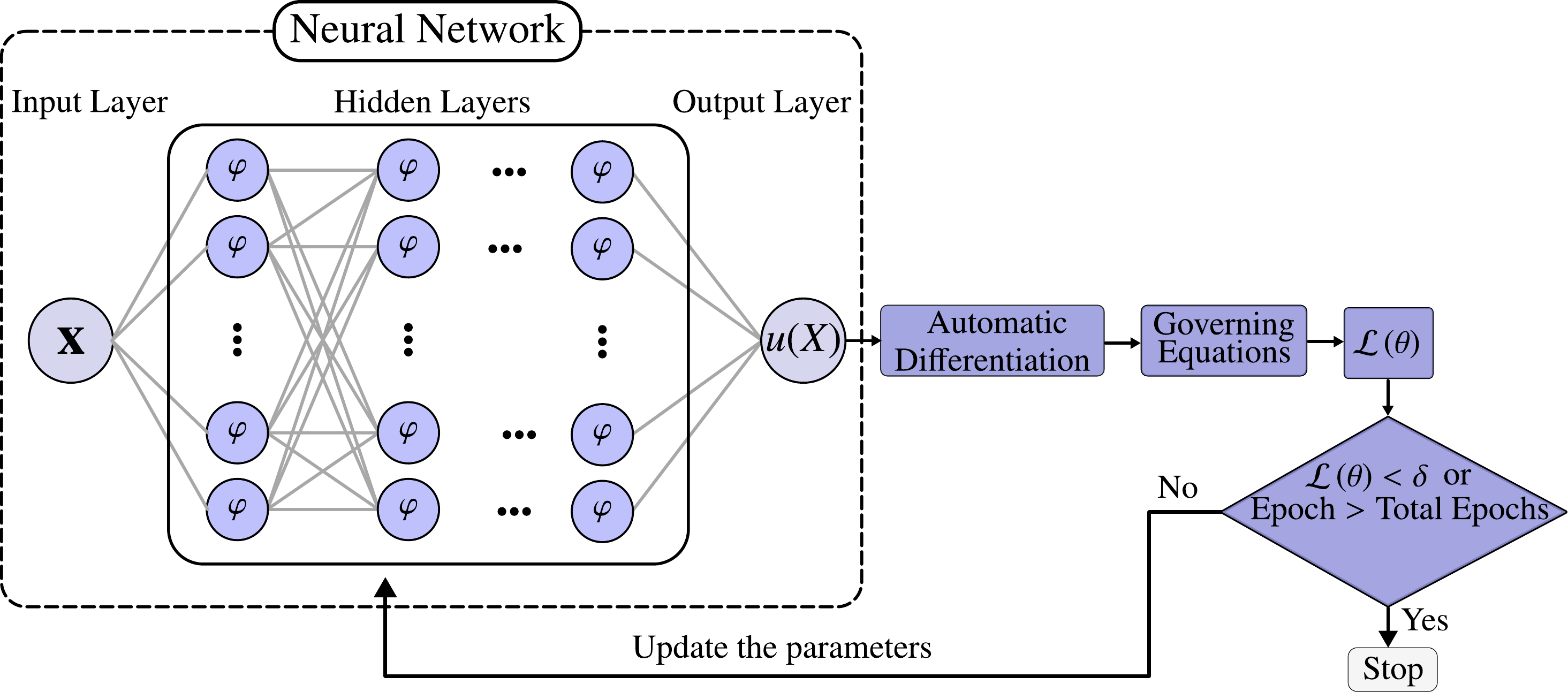}
    \caption{Schematic diagram of a PINN solver}
    \label{fig:neuralNetwork}
\end{figure}

\subsection{Physics Informed Neural Network (PINN)}
In a densely connected ANN, neurons are organized into sequential layers (Fig. \hyperref[fig:neuralNetwork]{\ref{fig:neuralNetwork}}). The first layer is the input layer, responsible for receiving the initial data, while the final layer is the output layer, which produces the network's prediction. Crucially, one or more "hidden layers" lie between the input and output layers. These hidden layers enable the network to learn complex, non-linear relationships within the data.  Adjacent layers are interconnected through weights $\mathbf{W}$, biases $\mathbf{b}$, and activation functions $\varphi$. These elements facilitate information flow and enable the network to learn and make predictions by transforming the input data through successive layers.  Mathematically, this neural network can be expressed as \( \mathbf{y} = \mathcal{N}(\mathbf{x}; \mathbf{W}, \mathbf{b}) \), where $\mathbf{x}$ is the input vector and $\mathbf{y}$ is the output vector. The propagation of information between layers can be described as a recursive transformation:

\begin{equation}
\mathbf{z}_l = \varphi_l \left( \mathbf{W}_l \cdot \mathbf{z}_{l-1} + \mathbf{b}_l \right), \quad l \in \{1, 2, \ldots, L\}
\end{equation}
where $L$ is the total number of layers, $\mathbf{z}_l$ is the output vector of $l$-th layer, thus, $\mathbf{z}_0= \mathbf{x}$ and $\mathbf{z}_L= \mathbf{y}$.

To optimize the performance of an Artificial Neural Network (ANN), a training regimen is implemented wherein an optimization algorithm iteratively refines the network's weights and biases. Central to this process is the formulation of a loss function, $\mathcal{L}$, which quantifies the discrepancy between the network's predicted outputs and the corresponding ground truth data.

In PINNs, as mentioned, specialized loss functions are designed to ensure that the neural network complies with the physical laws governing the problem, such as the relevant PDEs or the total potential of the system. To achieve this, Automatic Differentiation (AD) is utilized to compute precise derivatives of the neural network outputs with respect to their inputs. Compared to symbolic differentiation, which can become overly complex, and numerical differentiation, which may introduce noticeable errors and become expensive, AD is regarded as a more accurate and efficient method. This capability allows for the seamless integration of the governing equations into the loss formulation, enhancing the model's fidelity and performance.

Through AD, spatial and temporal derivatives of the network outputs can be evaluated at any point in the domain. This capability allows the governing PDEs to be enforced directly within the loss function. As a result, a flexible and mesh-free formulation is enabled, which is a key strength of the PINN framework. Note that the differentiation of certain non-standard or non-smooth functions, such as our enrichment functions, is not supported by PyTorch’s default AD implementation. To address this limitation, the enrichment functions were implemented manually and integrated into PyTorch through the custom autograd function capability. 

\subsubsection{Collocation Loss Function}
The Collocation Loss Function is a foundational element in PINNs, engineered to enforce the governing equations of a physical system at discrete spatial locations. In the context of solid mechanics, this encompasses the equilibrium equations, Dirichlet boundary conditions, and Neumann boundary conditions, as defined in Eqs. (\ref{equilibriumEqu}) and (\ref{BCs}). The Collocation Loss Function operates by minimizing the residual errors at specific collocation points. By evaluating the solution at these discrete points, the function ensures that the neural network's predictions adhere to the underlying physical laws. The mathematical representation of this loss function is as follows:
\begin{equation}
\mathcal{L}_{Collocation} = \frac{\eta_1}{n_\Omega} \sum_{\alpha=1}^{n_\Omega}\left( \sigma_{ij}^{\alpha}-f_i^{\alpha} \right)^{2} + \frac{\eta_2}{n_u} \sum_{\alpha=1}^{n_u}\left( u_i^{\alpha}  - \bar{u_i}^{\alpha} \right)^{2} + \frac{\eta_3}{n_t} \sum_{\alpha=1}^{n_t}\left( \sigma_{ij}^{\alpha} n_j  - \bar{t_i}^{\alpha} \right)^{2}
\end{equation}
In this formulation, \( n_\Omega \) represents the number of collocation points within the domain \( \Omega \), while \( n_u \) and \( n_t \) denote the number of collocation points on the Dirichlet boundary \( \Gamma_u \) and Neumann boundary \( \Gamma_t \), respectively. The parameters \( \eta_1 \), \( \eta_2 \), and \( \eta_3 \) are weighting coefficients. Without appropriate weighting, discrepancies in the magnitude of PDE and BC residuals can lead to the optimizer prioritizing the minimization of larger PDE-related terms, potentially neglecting the comparatively smaller BC contributions. To mitigate this, the weights \( \eta_1 \), \( \eta_2 \), and \( \eta_3 \) are introduced to ensure that the optimizer adequately addresses both PDE and BC residuals \cite{bai2023physics,mcclenny2023self}, thus facilitating more stable and efficient convergence during training. Recently, adaptive methodologies for determining these weights have been proposed, as detailed by Wang et al. \cite{wang2022and}, providing a sophisticated approach to dynamically balance the influence of PDEs and BCs within the loss function.

\subsubsection{Energy-Based Loss Function}
Energy-based loss functions offer a powerful alternative to collocation methods in PINNs applied to solid mechanics \cite{samaniego2020energy}. Instead of directly minimizing the residuals of the governing equations at discrete points, these functions integrate the physical energy of the system into the loss calculation. This approach ensures that the network's predictions are not only consistent with available data but also adhere to fundamental physical principles. The specific form of the energy-based loss function depends on the problem at hand; for example, in linear elasticity, the loss might be based on the total potential energy, defined as the sum of the strain energy and the external work, Eq. \eqref{totalPotential}. Express this mathematically, yields 
\begin{equation}\label{energyLoss}
\mathcal{L}_{Energy-Based} = \frac{1}{2} \int_{\Omega} \sigma_{ij} \epsilon_{ij}\, dV -\int_{\Omega} f_i u_i \, dV - \int_{\Gamma_t} \bar{t}_{i} u_i \, dS + \int_{\Gamma_u} \lambda_i |u_i - \bar{u_i}| \, dS
\end{equation}
All components and variables in Eq. \eqref{energyLoss} have been previously defined.

In PINN-based computational solid mechanics, effectively enforcing Dirichlet boundary conditions is essential. In the formulation above, these conditions are weakly enforced through the final term in the loss function. However, an alternative approach involves modifying the neural network output directly to satisfy the prescribed boundary conditions \cite{sukumar2022exact}. This can be achieved using methods such as multiplying by distance functions or designing custom neural network architectures, allowing the boundary constraints to be inherently satisfied. Such techniques simplify the physics-informed loss function and significantly reduce training complexity and computational cost. When the neural network is properly constructed to account for these conditions, the Dirichlet term can be excluded from the loss function entirely.

\section{Proposed Method} \label{proposedMethod}
\subsection{eXtended Physics Informed Neural Network (X-PINN)}
While the Universal Approximation Theorem guarantees that neural networks can approximate any continuous function, they often struggle with discontinuous functions. To address this limitation within the PINN framework, an approach analogous to the XFEM idea is proposed. This strategy, termed X-PINN, decomposes the solution field into three distinct components: a continuous base part, a discontinuous enrichment part (specifically designed to capture discontinuities), and a separate singular enrichment part (tailored to represent crack-tip singularities). This decomposition enables the framework to accurately represent and resolve solution features associated with both discontinuities and singularities, improving the overall accuracy and convergence of the PINN. Mathematically, this concept can be expressed as:
\begin{equation}\label{XPINN}
\begin{aligned}
&\left\{ \begin{array}{l}
\mathbf{u}(\mathbf{x})= \mathbf{u}_{\text{C}}(\mathbf{x}) + \mathbf{u}_{\text{D}}(\mathbf{x}, \mathbf{x}_{cb}) + \mathbf{u}_{\text{S}}(\mathbf{x}, \mathbf{x}_{ct}) \\
\mathbf{u}_{\text{C}}(\mathbf{x}) = \mathcal{N}_C(\mathbf{x}; \boldsymbol{\theta}_C) \\
\mathbf{u}_{\text{D}}(\mathbf{x}, \mathbf{x}_{cb}) = \mathbb{D}(\mathbf{x}, \mathbf{x}_{cb})\mathcal{N}_D(\mathbf{x}; \boldsymbol{\theta}_D) \\
\mathbf{u}_{\text{S}}(\mathbf{x}, \mathbf{x}_{ct}) = \mathbb{S}(\mathbf{x}, \mathbf{x}_{ct})\mathcal{N}_S(\mathbf{x}; \boldsymbol{\theta}_S)
\end{array} \right.
\end{aligned}
\end{equation}
where \( \mathcal{N}_C(\mathbf{x}; \boldsymbol{\theta}_C) \), \( \mathcal{N}_D(\mathbf{x}; \boldsymbol{\theta}_D) \), and \( \mathcal{N}_S(\mathbf{x}; \boldsymbol{\theta}_S) \) represent the neural networks associated with the continuous displacement field, displacement discontinuities, and crack-tip singularities, respectively. Furthermore, \( \mathbb{D}(\mathbf{x}, \mathbf{x}_{cb}) \) and \( \mathbb{S}(\mathbf{x}, \mathbf{x}_{ct}) \) denote the enrichment functions for displacement discontinuities and singularities, with \( \mathbf{x}_{cb} \) and \( \mathbf{x}_{ct} \) representing the coordinates of the crack body and crack tip, respectively.
Here's a more coherent description of each component in the X-PINN:

\begin{figure}[htbp]
    \centering
\includegraphics[scale=0.7]{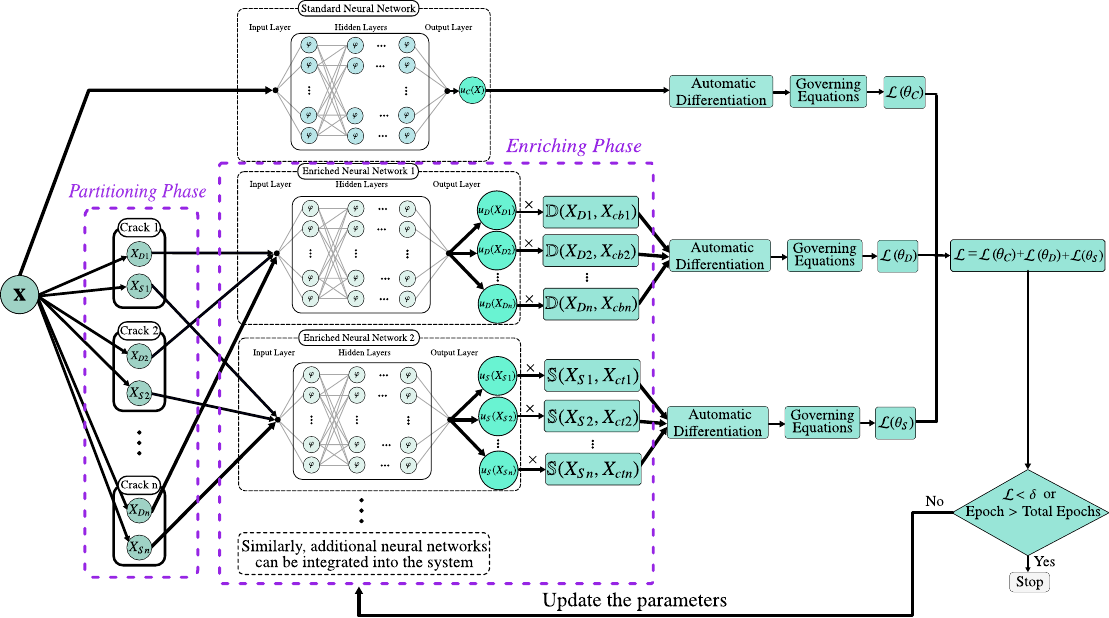}
    \caption{Workflow diagram of the X-PINN solution scheme 1}
    \label{fig:enrichedneuralNetwork}
\end{figure}

\begin{figure}[htbp]
    \centering
\includegraphics[scale=0.6]{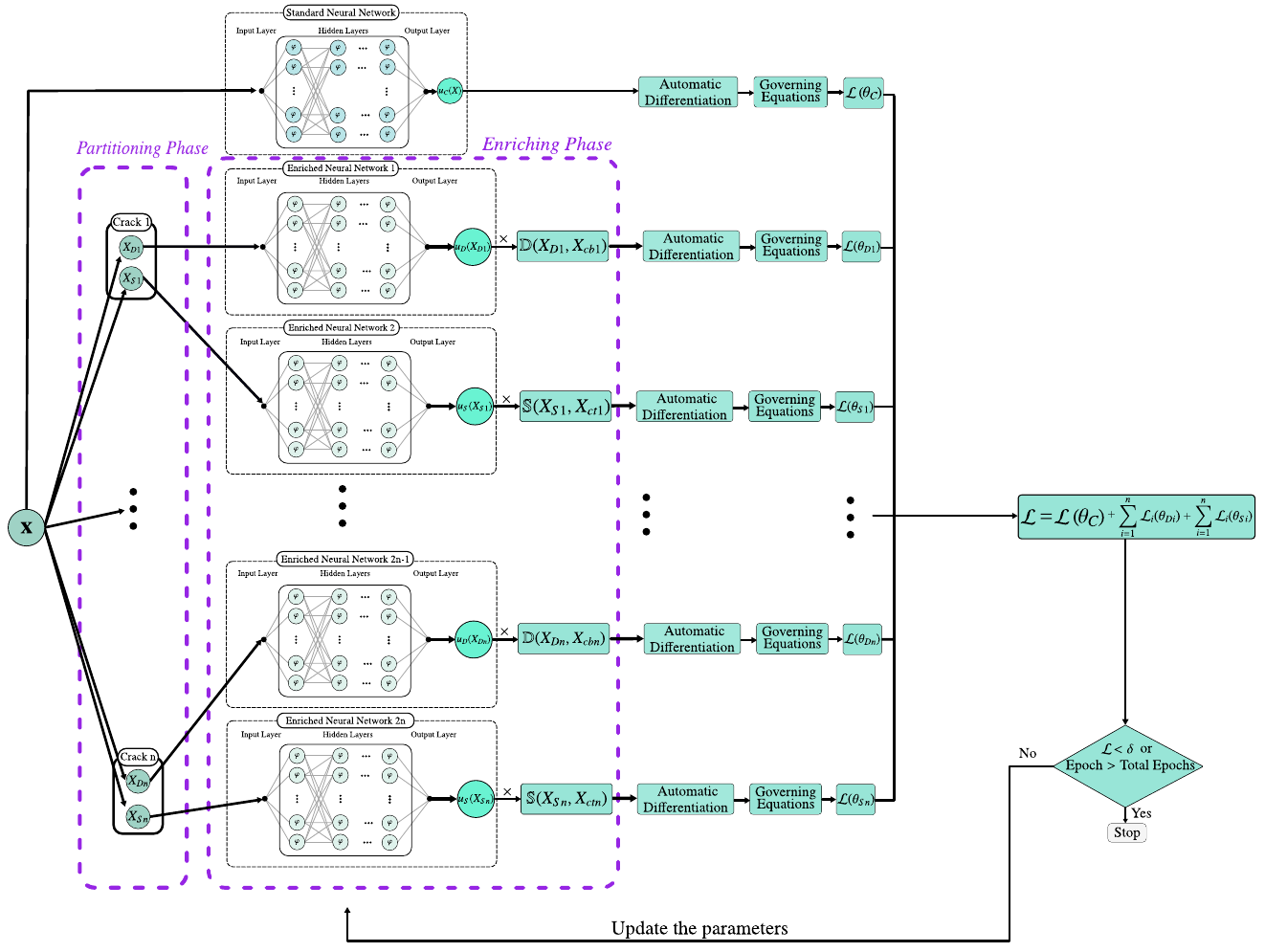}
    \caption{Workflow diagram of the X-PINN solution scheme 2}
    \label{fig:enrichedneuralNetwork_b}
\end{figure}
\noindent \textbf{Continuous Component, }\( \mathbf{u}_{\text{C}}(\mathbf{x}) \): Operating across the entire domain, this component captures the underlying smooth displacement field, effectively representing the material's continuous response away from any discontinuities or singularities. It's modeled using a neural network, \( \mathcal{N}_C \), which learns the material's inherent continuous behavior under normal conditions.

\noindent \textbf{Discontinuous Component,} \( \mathbf{u}_{\text{D}}(\mathbf{x}, \mathbf{x}_{cb}) \): This component operates on the enriched domain around crack bodies, addressing the abrupt jumps in the displacement field that occur at these locations. The enrichment function, \( \mathbb{D}(\mathbf{x}, \mathbf{x}_{cb}) \), plays a critical role by identifying and quantifying the effects of these discontinuities. Specifically, \( \mathbb{D}(\mathbf{x}, \mathbf{x}_{cb}) \) is multiplied by the output of the neural network \( \mathcal{N}_D(\mathbf{x}; \boldsymbol{\theta}_D) \), which models the displacement response associated with the discontinuity. This allows the model to capture the nuanced displacement behavior in the vicinity of the crack body, ensuring a more accurate representation of the transition across the discontinuity. By incorporating this enrichment, the X-PINN framework more reliably reflects the physical reality of stress distributions and displacement changes induced by cracks.

\noindent \textbf{Singular Component, }\( \mathbf{u}_{\text{S}}(\mathbf{x}, \mathbf{x}_{ct}) \): This component operates on the crack tip enrichment domain, addressing the singular stress and displacement fields that arise near crack tips. The enrichment function \( \mathbb{S}(\mathbf{x}, \mathbf{x}_{ct}) \) is specifically designed to enhance the neural network \( \mathcal{N}_S(\mathbf{x}; \boldsymbol{\theta}_S) \) by concentrating on the singular behaviors present at the crack tip. By multiplying \( \mathbb{S}(\mathbf{x}, \mathbf{x}_{ct}) \) with the output of the neural network, the approach enables a more accurate representation of stress concentrations and displacement fields in these critical regions. As a result, the X-PINN framework significantly improves its ability to model the complex behavior of materials under stress near crack tips, capturing important phenomena that influence crack propagation and structural integrity.

The above framework is implemented in two different schemes for multiple crack problems, as illustrated in Fig. \hyperref[fig:enrichedneuralNetwork]{\ref{fig:enrichedneuralNetwork}} and Fig. \hyperref[fig:enrichedneuralNetwork_b]{\ref{fig:enrichedneuralNetwork_b}}. The primary distinction between these schemes lies in their treatment of the enriched displacement components. It will be shown that selection of the scheme has a noticeable effect on the solution accuracy and computational cost. In the first scheme, a single neural network models the discontinuous components, and another network models the singular component of the displacement field. These two networks are shared across all cracks and are trained using input data from all enrichment domains collectively. In contrast, the second scheme allocates a dedicated pair of neural networks, one for the discontinuous component and one for the singular component, to each individual crack. Each pair is trained independently using data specific to its corresponding enrichment domain.

Both schemes integrate a standard neural network with multiple enriched neural networks within a partitioned domain learning strategy. Initially, the input domain is partitioned based on crack locations or geometric indicators to identify regions requiring enrichment. The input is then routed through a standard neural network that captures the global and smooth solution behavior, and through enriched neural networks tailored to model localized phenomena. These enriched networks receive not only the original input features, but also enrichment functions, such as 2D Sawtooth-Like or asymptotic crack-tip functions, to better represent singularities and discontinuities.

To ensure accurate learning, each enrichment neural network is trained exclusively on data points within its designated enrichment region(s). Including external points in the training set can mislead the optimizer, forcing it to minimize outputs to zero outside the region, thereby compromising the network’s ability to capture relevant local behavior effectively. The detailed partitioning and training process will be explained in the following sections.

The outputs of all networks are subjected to automatic differentiation to incorporate governing equations directly into the loss formulation. The total loss is computed as a weighted sum of residuals from the standard and enriched networks. During training, the network parameters are updated iteratively until a predefined convergence criterion such as a loss threshold or maximum number of epochs is met.

\subsection{Implementation}
\subsubsection{Enrichment Functions}  
\begin{figure}[h!]
    \centering
    \includegraphics[scale=0.4]{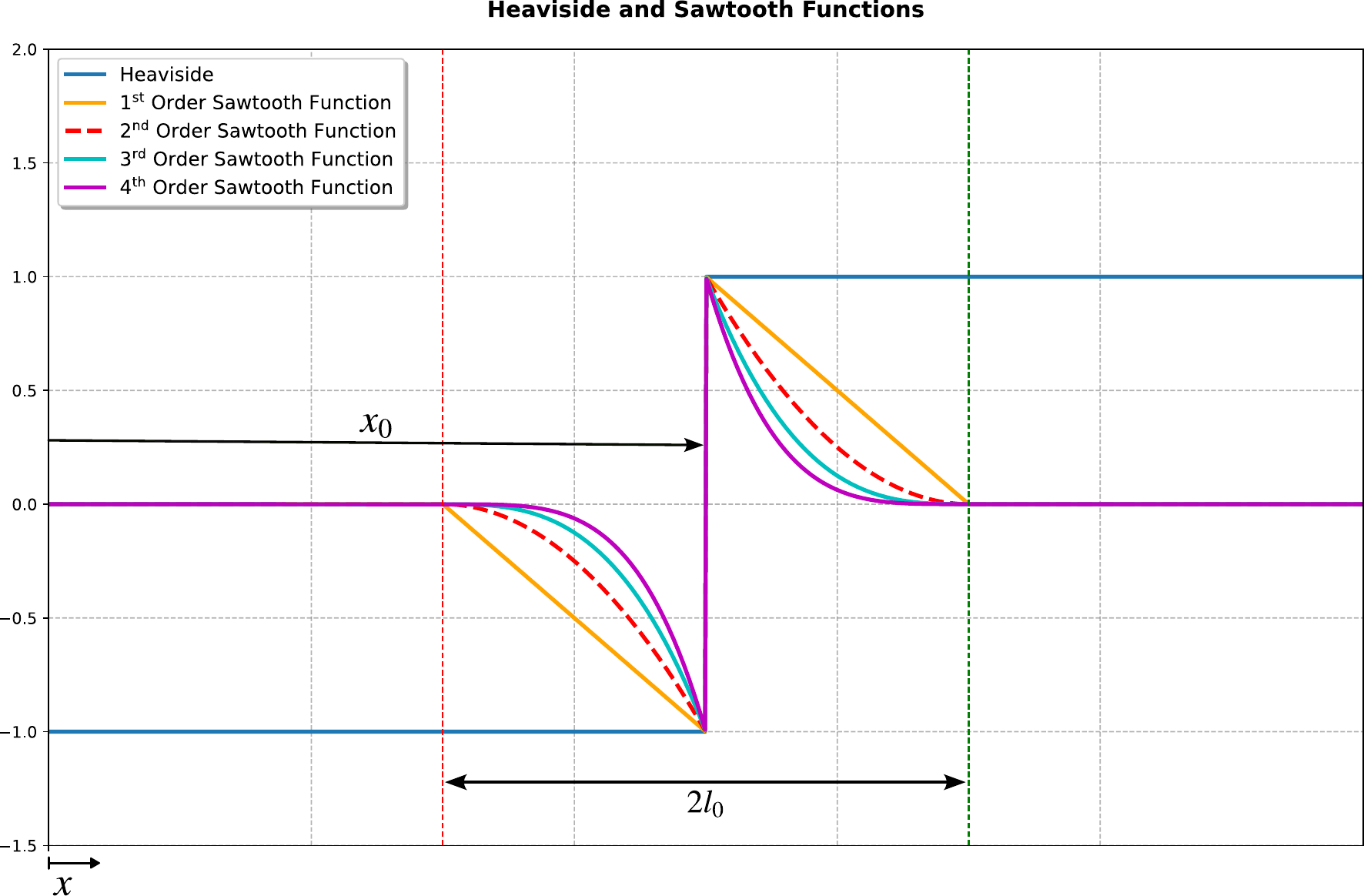}
    \caption{Heaviside and Sawtooth Functions (n=1,2,3,4)}
    \label{fig:ramHeaviside}
\end{figure}

In the proposed methodology, the incorporation of enrichment functions is crucial for capturing localized behaviors and discontinuities. For one-dimensional problems, the Heaviside and Sawtooth functions (Fig.~\hyperref[fig:ramHeaviside]{\ref*{fig:ramHeaviside}}) were employed as prominent enrichment functions to add discontinuity in the solution.

The Heaviside function, \(H(x)\), is a step function that is used to model discontinuities:

\begin{equation}\label{HeavisideFunction}
H(x) =
\begin{cases}
-1 & \text{for } x < x_0 \\
1 & \text{for } x \geq x_0
\end{cases}
\end{equation}

Sudden changes are represented effectively by this function, and it is foundational in many mathematical models.

The Sawtooth function (Eq. \eqref{sawtoothFunction}) enriches the analysis by modeling localized variations and discontinuities. It provides a smooth transition around a central point \(x_0\) with half-width \(l_0\), enabling controlled approximations between states. Specifically, the function smoothly ramps down to -1 as \(x\) approaches \(x_0\) from the left and ramps up to 1 as \(x\) approaches \(x_0\) from the right.

\begin{equation}\label{sawtoothFunction}
S(x) =
\begin{cases}
-\left(\frac{x - (x_0 - l_0)}{l_0}\right)^{n} & \text{if } x_0 - l_0 \leq x < x_0 \\
\left(\frac{x - (x_0 + l_0)}{l_0}\right)^{n} & \text{if } x_0 < x \leq x_0 + l_0 \\
0 & \text{otherwise}
\end{cases}
\end{equation}
where \(n\) is the order of the Sawtooth function that controls its smoothness (Fig. \ref{fig:ramHeaviside}). The Sawtooth function offers the advantage that its value and its (\(n-1\))-th derivative(s) vanish at a distance \(l_0\) from the discontinuity. This property guarantees continuity of the displacement field and its derivative(s), as described in Eq. \eqref{XPINN}, at the interface between the enriched and standard domains. Unlike the Sawtooth function, the Heaviside function may introduce discontinuities in that interface.

\begin{figure}[htbp]  
    \centering  
    \begin{subfigure}[b]{0.49\linewidth}  
        \centering  
        \includegraphics[height=0.64\textwidth]{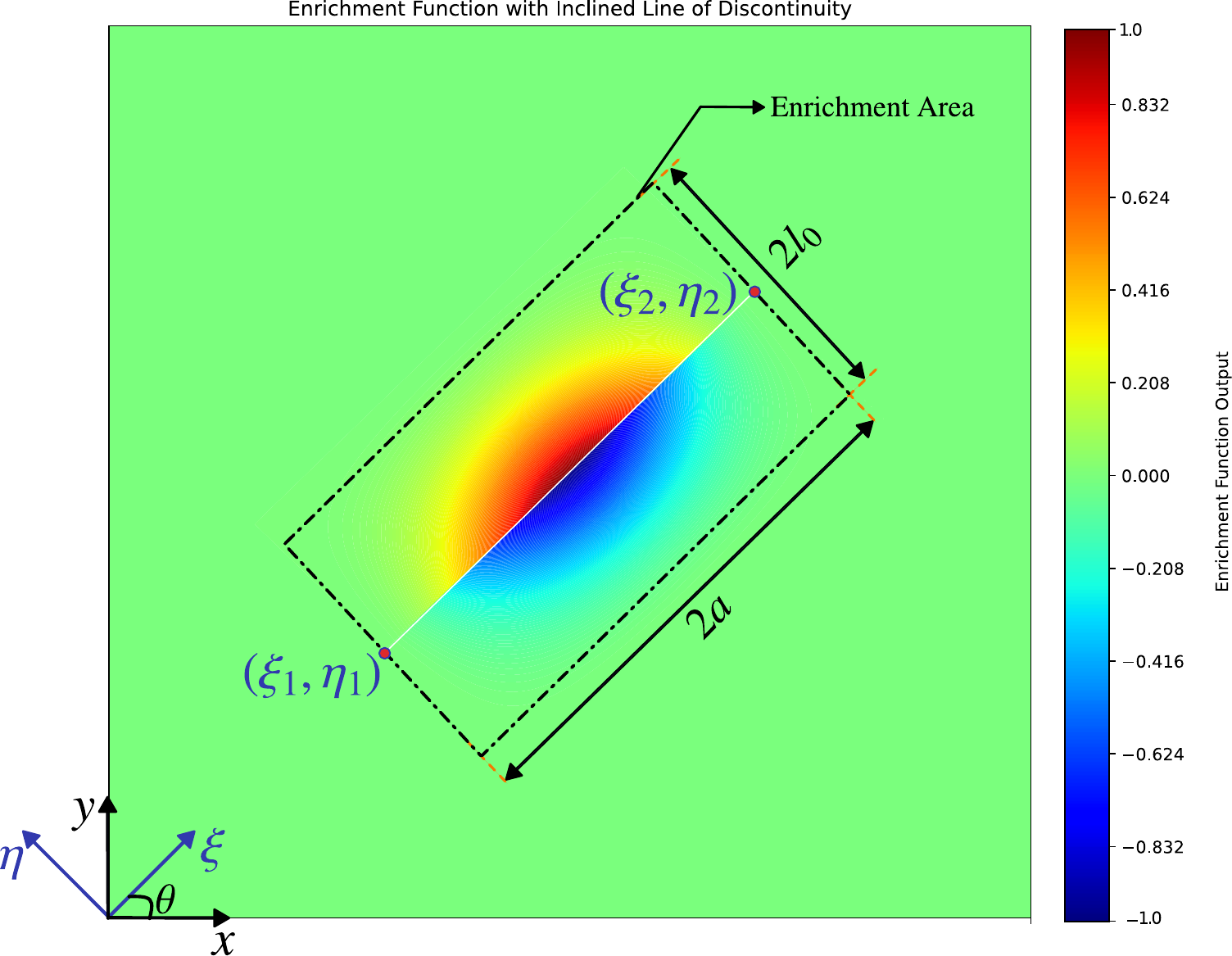}  
        \caption*{\text{(a)}}  
    \end{subfigure}  
    \hfill  
    \begin{subfigure}[b]{0.49\linewidth}  
        \centering  
        \includegraphics[height=0.64\textwidth]{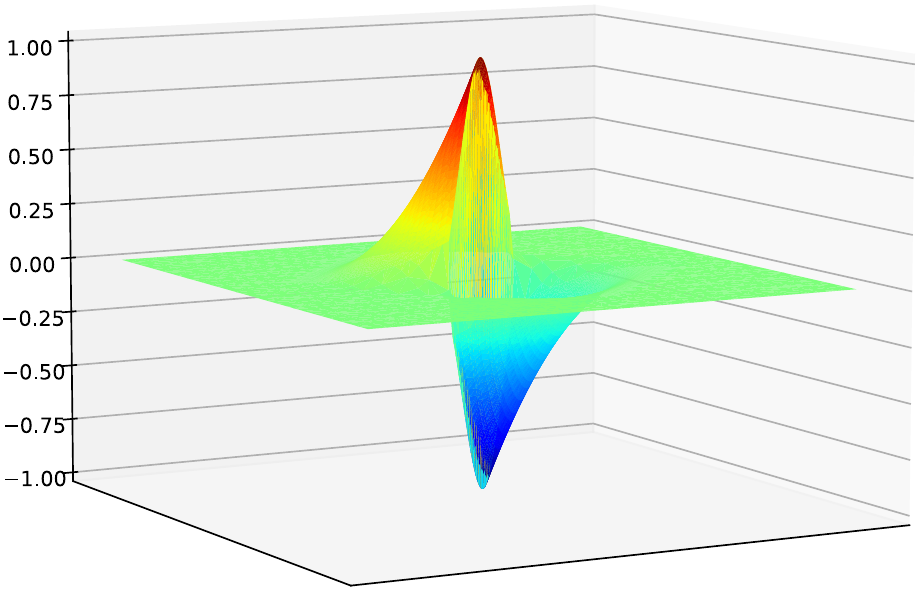} 
        \caption*{\text{(b)}}  
    \end{subfigure}  
    \caption{The 2D enrichment function and corresponding enrichment domain for an inclined center crack: (a) Top view, (b) 3D view}  
    \label{fig:inclinedEnrichment}  
\end{figure}

For two-dimensional fracture modeling, we introduce a novel enrichment strategy employing a sawtooth-like function, \(\mathbb{D}(\xi, \eta)\), defined within a rectangular enrichment domain surrounding the crack body (refer to Fig.~\hyperref[fig:inclinedEnrichment]{\ref*{fig:inclinedEnrichment}}). This enrichment function is mathematically expressed as:

\begin{equation}
\mathbb{D}(\xi, \eta) =
\begin{cases}
\Xi(\xi) \cdot \Lambda(\eta), & \text{if inside the enrichment region} \\
0, & \text{otherwise}
\end{cases}\label{2DEnrichment} 
\end{equation}

where \(\Xi(\xi)\), governing the behavior along the \(\xi\) direction, is a cubic polynomial:
\begin{equation}
\Xi(\xi) =
\begin{cases}
a_0 + a_1 \xi + a_2 \xi^2 + a_3 \xi^3, & \text{if } \xi_1 \leq \xi \leq \xi_2 \\
0, & \text{otherwise}
\end{cases}\label{Xi}
\end{equation}

and \(\Lambda(\eta)\), defining the function's profile in the \(\eta\) direction, is a quadratic polynomial:

\begin{equation}
\Lambda(\eta) =
\begin{cases}
b_0 + b_1 \eta + b_2 \eta^2, & \text{if } \eta_1 - l_0 \leq \eta \leq \eta_1 + l_0 \\
0, & \text{otherwise}
\end{cases}\label{Lambda}
\end{equation}
The coefficients \(a_0\), \(a_1\), \(a_2\), \(a_3\), \(b_0\), \(b_1\), and \(b_2\) are determined to ensure the desired smoothness and continuity properties at the interface between the enriched and standard domains; detailed calculations and derivations of these functions are provided in the Appendix. The parameters \( (\xi_1, \eta_1) \) and \( (\xi_2, \eta_2) \) represent the coordinates of the crack tips in the local \(\xi\)-\(\eta\) coordinate system, which is rotated with respect to the global coordinate system by an angle \(\theta\), defined as:

\begin{subequations}
\begin{equation}
    \xi = x \cos\theta + y \sin\theta
\end{equation}
\begin{equation}
    \eta = -x \sin\theta + y \cos\theta
\end{equation}
\begin{equation}
    \theta = \tan^{-1} \left(\frac{y_2 - y_1}{x_2 - x_1}\right)
\end{equation}
\end{subequations}
where \((x_1, y_1)\) and \((x_2, y_2)\) are the coordinates of the crack tips in the global coordinate system. As illustrated in Figure \ref{fig:inclinedEnrichment}, the rectangular enrichment domain, within which this function operates, is strategically positioned around each crack. The key advantage of this approach lies in the smooth transition it provides for the displacement field and its derivatives across the interface between the enriched and standard domains. This smoothness is crucial for obtaining accurate and stable solutions, particularly in fracture mechanics problems. In contrast, our simulations have demonstrated that employing Heaviside-like functions, which introduce discontinuities at the interface, leads to convergence difficulties and prevents the attainment of feasible solutions.

Although the enrichment function in Eq. \eqref{2DEnrichment} are designed for internal cracks, it can also be applied to edge cracks. The only requirement is that the crack tip coordinates are specified such that the crack center lies on the edge.

Asymptotic enrichment functions are used to represent the local behavior of fields near singularities, such as crack tips, by reflecting the dominant terms of the asymptotic solution to the governing equations. In the context of linear elastic fracture mechanics, the displacement field near a crack tip is known to exhibit square-root singularities. This behavior can be captured using the following set of functions, expressed in polar coordinates \( (r, \theta) \) centered at the singular point:

\begin{equation}
\begin{aligned}
    F_1(r, \theta) &= \sqrt{r} \sin\left(\frac{\theta}{2}\right), \\
    F_2(r, \theta) &= \sqrt{r} \cos\left(\frac{\theta}{2}\right), \\
    F_3(r, \theta) &= \sqrt{r} \sin\left(\frac{\theta}{2}\right) \sin(\theta), \\
    F_4(r, \theta) &= \sqrt{r} \cos\left(\frac{\theta}{2}\right) \sin(\theta).
\end{aligned}
\end{equation}

These functions provide a mathematically accurate description of the singular field and are widely used in analytical studies and enrichment-based numerical methods to improve the accuracy of solutions near discontinuities. It is worth noting that asymptotic enrichment functions can be incorporated in a similar manner to the enrichment functions used for modeling displacement discontinuities. However, in this study, such asymptotic functions are not employed in the numerical examples.

\subsection{Numerical Integration}
The numerical integration of the energy-based loss function (Equation \ref{energyLoss}) plays a crucial role in our proposed methodology, significantly influencing the accuracy, stability, and overall efficiency of the solutions. Our simulations indicate that inadequate integration can lead to solution divergence and, in many cases, an inability to arrive at a feasible solution. Although the proposed loss function is mathematically convex and has a unique solution, the interplay of neural networks, boundary conditions, and numerical integration techniques can inadvertently create local minima that deviate significantly from the true solution. 

Additionally, the discontinuous nature of the displacement field necessitates tailored integration strategies. In traditional finite element methods, Gauss quadrature is usually employed for integration on each element, requiring a mesh. In the context of XFEM, where discontinuities may traverse an element, additional partitioning techniques such as triangulation and rectangulation become essential \cite{khoei2015extended}. However, these meshing processes can be resource-intensive and may compromise the meshless benefits of the X-PINN approach. To tackle these challenges, this section introduces two effective integration strategies specifically designed for crack problems, focusing on improving accuracy and robustness while alleviating integration difficulties.

\subsubsection{Integration by Uniformly Distributed Integration Points Method}
In PINNs, integration by Uniformly Distributed Integration Points Method (UDIPM) is a common approach for evaluating the loss function, which includes terms from partial differential equations and boundary conditions. This method is a special case of Newton-Cotes quadrature, where the integration weights are uniform.

In one dimension, the integral of a function \( f(x) \) over a domain \( [a, b] \) is approximated using \( n \) uniformly distributed points as:

\begin{equation}
I \approx \int_a^b f(x) \, dx \approx \frac{b-a}{n} \sum_{i=1}^{n} f(x_i)
\end{equation}

In two dimensions, for a function \( f(x, y) \) over a domain \( \Omega \), the integral is approximated using uniformly distributed points \( (x_i, y_i) \) as:

\begin{equation}
I \approx \iint_{\Omega} f(x, y) \, dA \approx \frac{A}{n} \sum_{i=1}^{n} f(x_i, y_i)
\end{equation}

Here, \( n \) is the number of integration points, and \( A \) is the area of the domain in 2D. This approach provides a straightforward way to approximate integrals across both 1D and 2D domains in PINNs. This approach offers simplicity and broad applicability; however, it may not be as effective in scenarios where there are hollow regions or regions with significant variations in the function being integrated, such as in crack problems. In such cases, uniformly distributed points can lead to inaccuracies because they may not adequately capture the behavior of the function in critical areas. For instance, if there are voids or discontinuities in the domain, or if the function varies rapidly in some regions, the uniform distribution of points might miss sampling these regions effectively, resulting in a biased approximation of the integral. Also, if \( \Omega \) has a complex shape, ensuring truly uniform distribution can be tricky. As a remedy, in such situations, the domain can be partitioned into regular shapes, and the integration can be applied to each partition separately. For example, in the case of a crack, to handle the discontinuity in the displacement field, the enrichment domain is often partitioned into positive and negative domains based on the sign of the enrichment function. Integration is then performed on each domain separately, and the results are summed. This technique is very similar to triangulation or rectangulation techniques employed in the integration of enriched elements in XFEM \cite{khoei2015extended}. Alternatively, more advanced integration techniques might also be employed.

\subsubsection{Cartesian Transformation Method}
Cartesian Transformation Method (CTM) was specifically developed for integration in meshless methods \cite{khosravifard2010new}. Since X-PINN functions independently of conventional mesh requirements, this study utilizes CTM as the primary integration process. This method is not only accurate and easy to implement but also aligns seamlessly with the proposed domain decompositions in X-PINN.

In this section, CTM is illustrated using the two-dimensional integration as an example. A similar methodology can be employed for three-dimensional integration. For additional details, refer to \cite{khosravifard2010new}.

Consider the computation of the following integral:
\begin{equation}\label{}
I = \int_\Omega f(x, y) \, d\Omega
\end{equation}
where \( f \) is an arbitrary regular function and \( \Omega \) represents the integration domain. A simple auxiliary domain $\bar{\Omega}$, such as a rectangle, is constructed such that it contains the original domain as shown in Fig.~\hyperref[fig:CTM]{\ref*{fig:CTM}}. A new function \( g \) is defined over the auxiliary domain $\bar{\Omega}$ such that:
\begin{figure}[h!]
    \centering
\includegraphics[scale=0.3]{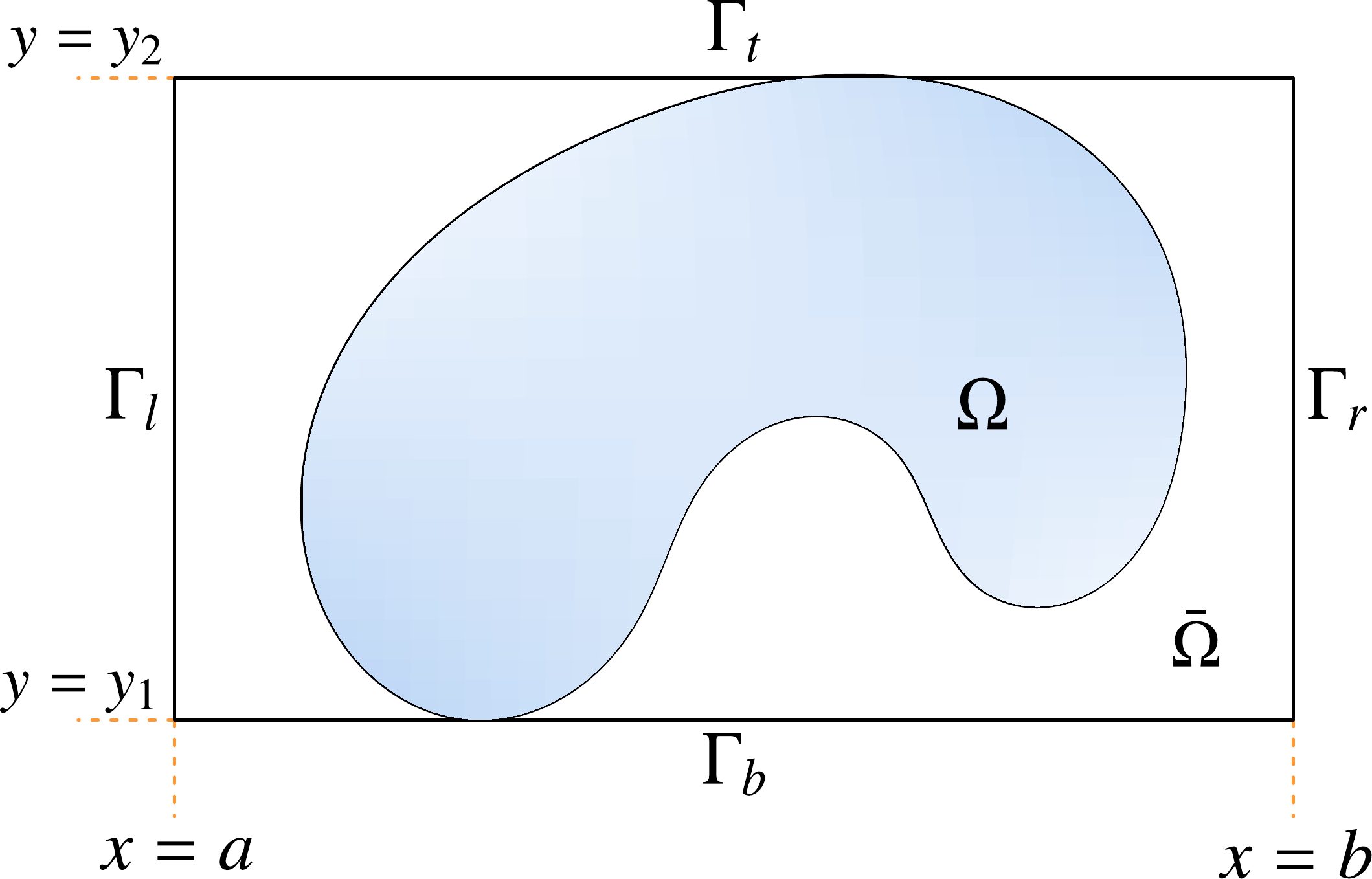}
    \caption{Integration domain \(\Omega\), contained in a rectangular auxiliary domain \(\bar \Omega\).}
    \label{fig:CTM}
\end{figure}
\begin{equation}\label{integraion2D}
g(x, y) = 
\begin{cases} 
f(x, y) & \text{if } (x, y) \in \Omega \\
0 & \text{otherwise} 
\end{cases} 
\end{equation}
the integral in Eq. \eqref{integraion2D} can be rewritten as:
\begin{equation}\label{2DIntegral}
I = \int_\Omega f(x, y) \, d\Omega = \int_{\bar{\Omega}} g(x, y) \, d\Omega.
\end{equation}
Applying Green’s theorem, the domain integral Eq. \eqref{2DIntegral} can be transformed into a double integral:
\begin{equation}\label{transformed2DIntegral}  
\int_{\bar{\Omega}} g(x, y) \, d\Omega = \int_{\bar{\Gamma}} \left( \int_{x}^{c} g(\xi, y) \, d\xi \right) \, dy.  
\end{equation}
where, \( c \) is an arbitrary constant. In this equation, the outer integral represents a boundary integral over the simple boundary \( \bar{\Gamma} \), while the inner integral is a one-dimensional integral that is independent of the domain geometry. Since \( \bar{\Omega} \) is rectangular, the boundary \( \bar{\Gamma} \) consists of four portions:

\begin{equation}  
    \bar{\Gamma}_b: \, y = y_1, \quad \bar{\Gamma}_r: \, x = b, \quad \bar{\Gamma}_t: \, y = y_2, \quad \bar{\Gamma}_l: \, x = a.  
\end{equation}
Assuming \( c = a \), the integral vanishes on \( \Gamma_b, \Gamma_t, \Gamma_l \) and remains only on \( \Gamma_r \):
\begin{equation}
I = \int_{y_1}^{y_2} \int_a^b g(x, y) \, dx \, dy. 
\end{equation}
This can be written as:
\begin{equation}
I = \int_{y_1}^{y_2} h(y) \, dy, 
\end{equation}
where
\begin{equation}
h(y) = \int_a^b g(x, y) \, dx. \tag{10}
\end{equation}
Using the composite Gaussian quadrature method, divide \( \Gamma_r \) into \( n \) intervals and apply an \( m \)-point Gaussian quadrature to each interval (Fig.~\hyperref[fig:CTM2_3]{\ref*{fig:CTM2_3}}):

\begin{figure}[htbp]  
    \centering  
    \begin{minipage}{0.49\linewidth}  
        \centering  
        \includegraphics[scale=0.2]{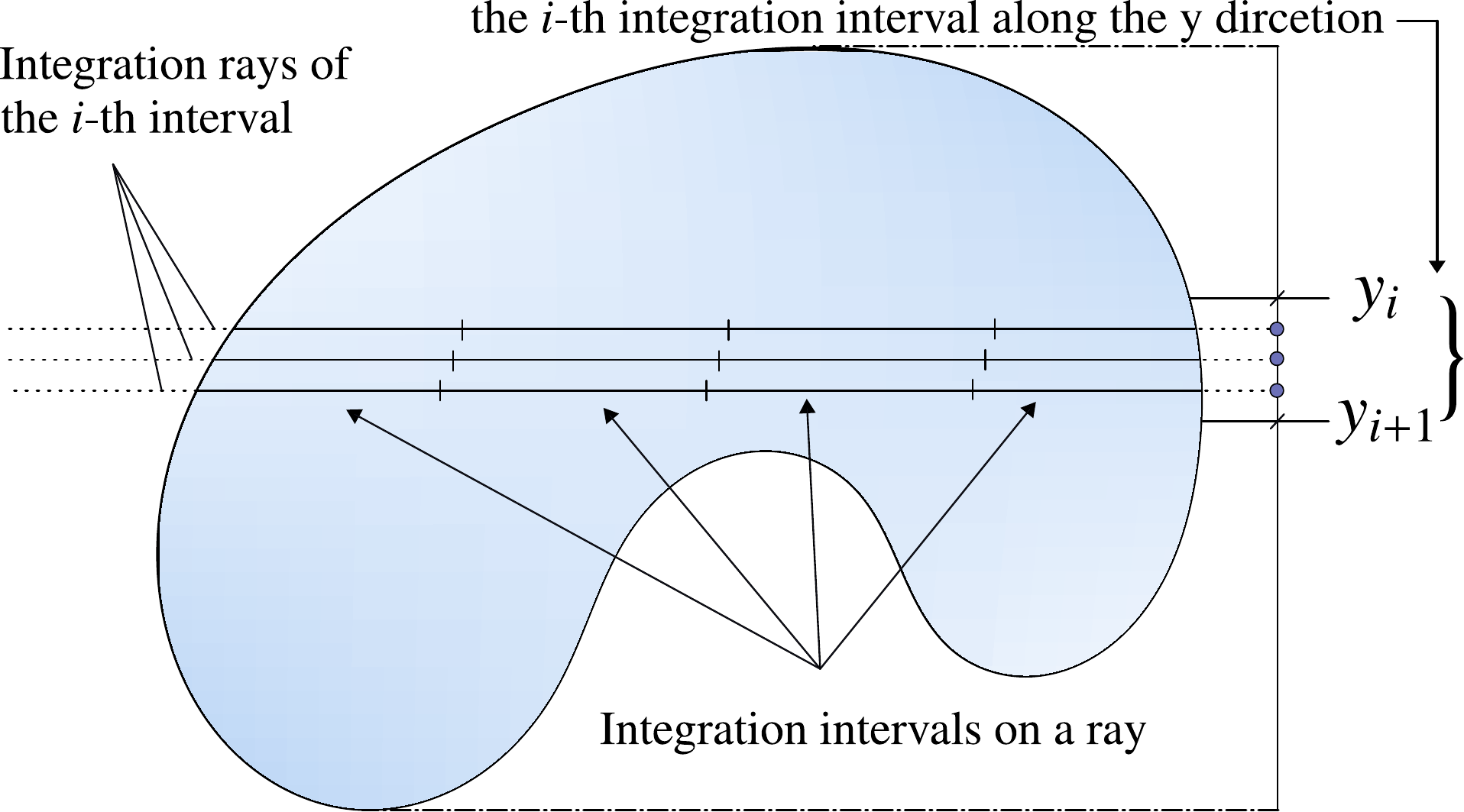}  
        \caption*{(a)}  
        \label{fig:CTM_2}  
    \end{minipage}  
    \hfill  
    \begin{minipage}{0.49\linewidth}  
        \centering  
        \includegraphics[scale=0.17]{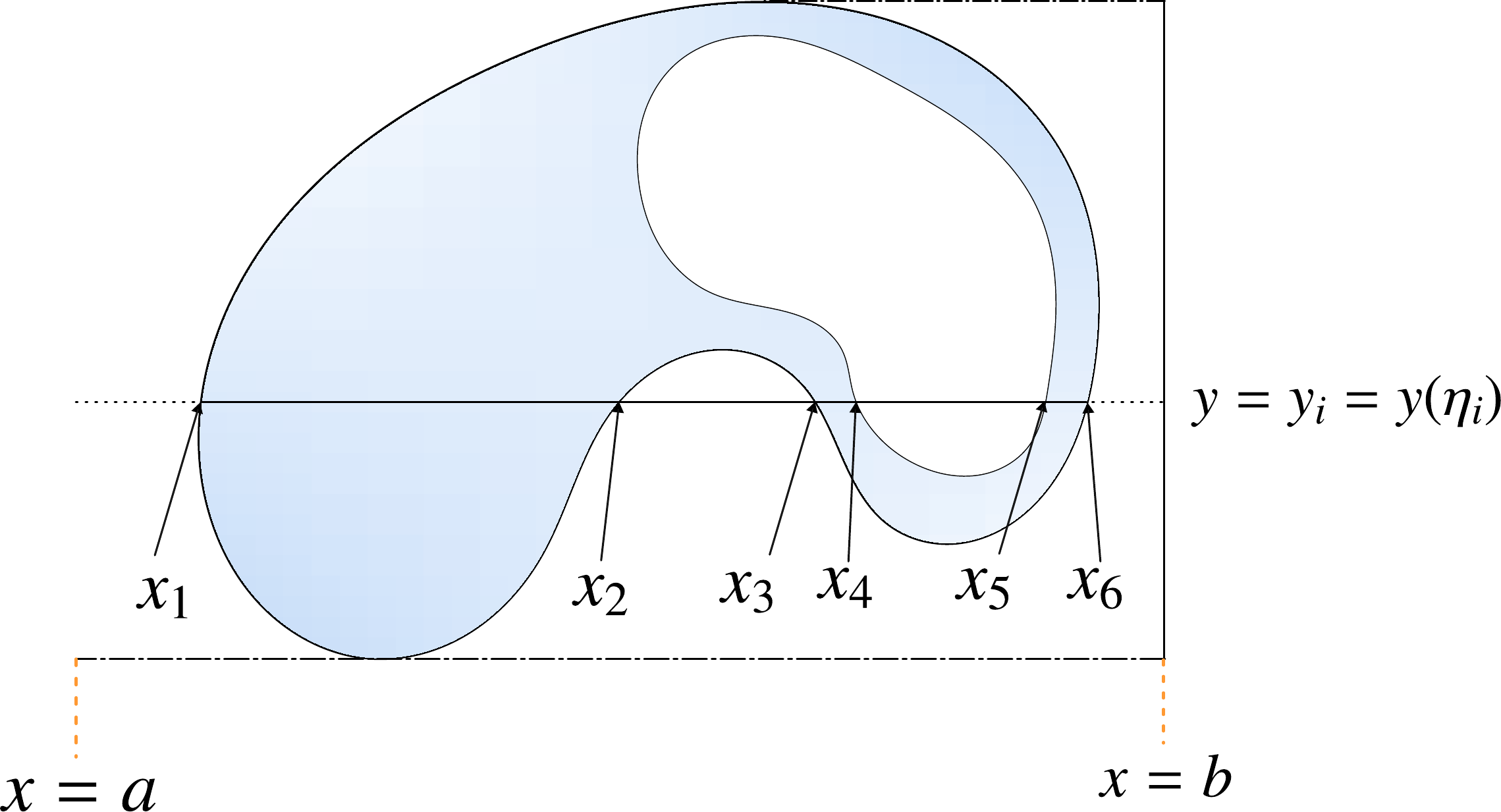}  
        \caption*{(b)}  
        \label{fig:CTM_3}  
    \end{minipage}  
    \caption{CTM: (a) Integration rays and intervals, (b) Intersection of a ray with boundary at even number of points (The interval $[x_2, x_3]$ is outside the domain and is excluded from the calculation. Additionally, the interval $[x_3, x_4]$ is excluded to account for the presence of the hole).}  
    \label{fig:CTM2_3}  
\end{figure}
 
\begin{equation}
I = \sum_{i=1}^{n} \int_{y_i}^{y_{i+1}} h(y) \, dy = \sum_{i=1}^{n} \int_{-1}^{1} H(\eta) J \, d\eta, 
\end{equation}
or
\begin{equation}
I = \sum_{i=1}^{n} \sum_{j=1}^{m} J w_j H(\eta_j), 
\end{equation}
where
\begin{equation}
H(\eta) = h(y(\eta)),
\end{equation}
and $\eta_j$ and $w_j$ are the Gaussian points and weights, respectively. \( J \) is the transformation Jacobian:
\begin{equation}
J = \frac{dy}{d\eta} = \frac{y_{i+1} - y_i}{2}. 
\end{equation}

The expression \( y = y(\eta_j) \) denotes a horizontal line, referred to as the integration ray (Fig.~\hyperref[fig:CTM2_3]{\ref*{fig:CTM2_3}}). For \( h(y) \) evaluation, suppose a ray intersects the domain boundary \( 2l \) times (Fig.~\hyperref[fig:CTM2_3]{\ref*{fig:CTM2_3}}). The ray is divided alternately inside and outside \(\Omega \):
\begin{equation}\label{integrationAlongX}
h(y_i) = \int_a^b g(x, y_i) \, dx = \sum_{j=1}^{l} \int_{x_{2j-1}}^{x_{2j}} f(x, y_i) \, dx. 
\end{equation}

Apply the composite Gaussian method to each integral:
\begin{equation}
\int_{x_{2j-1}}^{x_{2j}} f(x, y_i) \, dx = \sum_{r=1}^{n} \int_{-1}^{1} F(x) J_0 \, dx = \sum_{r=1}^{n} \sum_{s=1}^{m} \bar{J} w_s F(\xi_s), 
\end{equation}
where
\begin{equation}
F(\xi) = f(x(\xi), y_i), 
\end{equation}
and
\begin{equation}
\bar{J} = \frac{x_{2j} - x_{2j-1}}{2n}. 
\end{equation}

CTM effectively evaluates domain integrals by ensuring that integration points remain within the domain, simply omitting the corresponding intervals in Eq. \eqref{integrationAlongX}.
\subsubsection{Domain Partitioning}
\begin{figure}[h!]
    \centering
    \includegraphics[scale= 0.85]{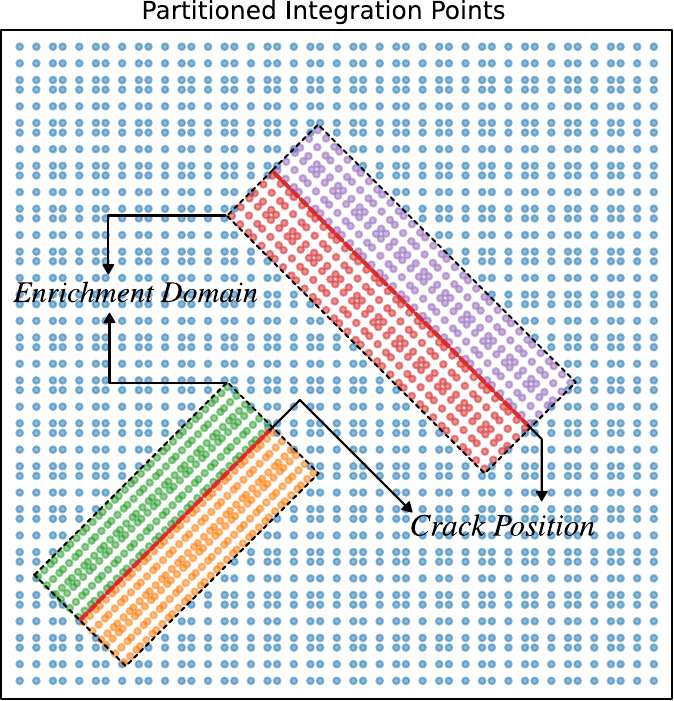}
    \caption{Partitioning of the domain for a two crack problem. A rectangular aligned partition is devised for each crack with different sign values at each side. CTM integration points are placed in each partition. }
    \label{fig:partitionGP}
\end{figure}

To achieve sufficiently accurate results with the X-PINN method, it is essential to avoid using integration points located outside the enrichment domains when training the enriched network(s). Instead, integration points should be generated separately, using the CTM for both the enrichment domains and the regions outside them. This requires a clear and precise partitioning of the computational domain to ensure the effectiveness of the training process. As shown in Fig.~\hyperref[fig:partitionGP]{\ref*{fig:partitionGP}}, a rectangular enriched partition is defined around each crack, aligned with its orientation. Within each partition, CTM-based integration points are placed such that they carry different signs on either side of the crack, capturing the discontinuity. Integration points outside the enriched partitions are also generated uniformly using the CTM procedure. This partitioning strategy is integrated into the X-PINN framework as shown in Figs. \ref{fig:enrichedneuralNetwork} and \ref{fig:enrichedneuralNetwork_b}.

The rationale behind this partitioning is that, when integration points from both inside and outside the enrichment domain are used to train the enriched network, the network is expected to output zero for points outside the enrichment region. In such cases, the optimizer is burdened with the difficult task of tuning the network parameters to produce accurate outputs within the enrichment domain, while simultaneously forcing the output to be zero outside it. This dual objective can significantly compromise training efficiency and accuracy.

However, if the training is restricted solely to points within the enrichment domain, there is no need to enforce a zero-output condition outside the domain. This allows the optimizer to focus exclusively on learning the correct behavior within the enrichment region, thereby eliminating a major source of error. It is important to note that all integration points—both within and outside the enrichment domains—are still utilized in the training process of the standard (non-enriched) network.

\section{Numerical Examples} \label{numericalExamples}
This section presents several numerical examples to evaluate the accuracy, robustness, and efficiency of the proposed X-PINN method. A total of four examples are considered: one one-dimensional and three two-dimensional problems. The results are compared with analytical solutions or extended finite element solutions obtained from Abaqus. In the two-dimensional cases, different crack configurations and loading conditions are examined to test the ability of the method to capture complex mechanical behavior.
\subsection{1D Cracked Bar}
\begin{figure}[h!]
    \centering
    \includegraphics[scale=0.4]{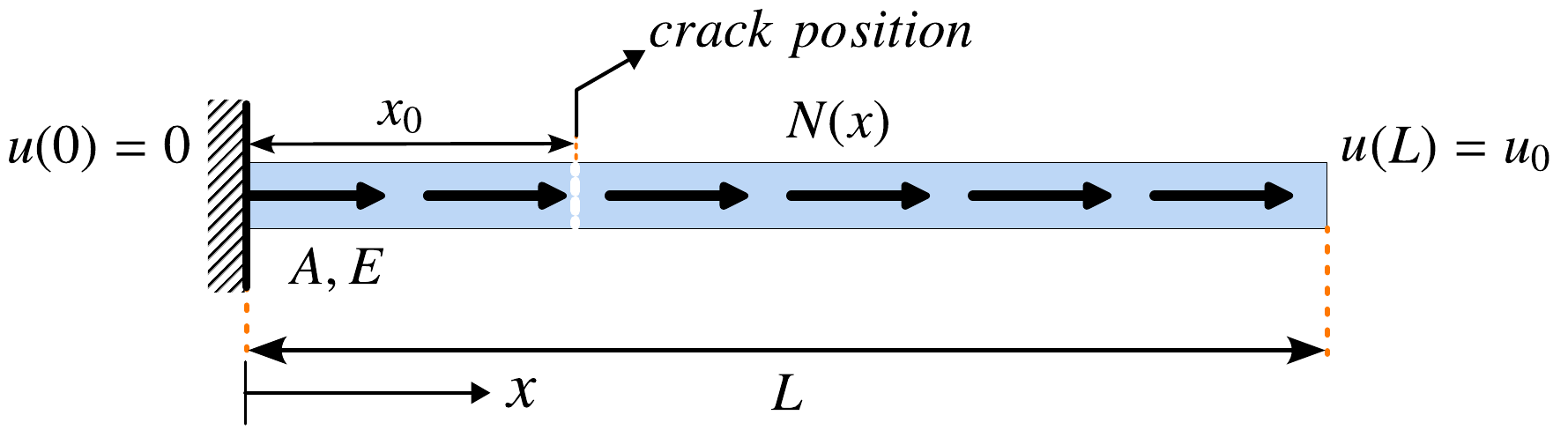}
    \caption{Geometry and boundary conditions of the 1D Cracked Bar problem}
    \label{fig:barProblem}
\end{figure}

Since many key properties of the proposed method are best demonstrated in one dimension, the first example analyzes a cracked one-dimensional elastic bar, with its geometry and corresponding boundary conditions illustrated in
Fig.~\hyperref[fig:barProblem]{\ref*{fig:barProblem}}.

\subsubsection{Analytical Solution}
The analytical solution for the one-dimensional cracked elastic bar problem can be derived by transforming it into two separate differential equations, each with its own set of boundary conditions. This approach is illustrated in Eqs. \eqref{differntialEqu1} and \eqref{differntialEqu2}.

\begin{equation}\label{differntialEqu1}
    \begin{aligned}
        &\frac{d^2 u(x)}{dx^2} + \frac{N(x)}{AE} = 0, \quad \quad 0 \leq x \leq x_0, \\ 
        &u(0) = 0, \quad \frac{du}{dx}(x_0) = 0.
    \end{aligned}
\end{equation}

\begin{equation}\label{differntialEqu2}
    \begin{aligned}
        &\frac{d^2 u(x)}{dx^2} + \frac{N(x)}{AE} = 0, \quad \quad x_0 \leq x \leq L, \\
        &\frac{du}{dx}(x_0) = 0, \quad u(L) = u_0.
    \end{aligned}
\end{equation}
Under the assumption that \( A \), \( E \), and \( N(x) \) are constant throughout the length of the bar, the analytical solution is:
\begin{equation} \label{analyticalSol}
    u(x) =
    \begin{cases}
        -\frac{1}{2AE}(N(x)x^{2} - 2x_0N(x)x) & 0 \leq x \leq x_0, \\
        -\frac{1}{2AE}N(x)(x-x_0)^{2} + \frac{1}{2AE}N(x)(L-x_0)^{2} + u_0 & x_0 \leq x \leq L.
    \end{cases}
\end{equation}

For simplicity, it is assumed that Young’s modulus \( E \), the cross-sectional area \( A \), and the length of the bar \( L \) are all set to one. The crack position \( x_0 \) is defined as 0.3, and the applied body force \( N(x) \) is set to 10.

\subsubsection{Energy-based X-PINN Solution}
In this section, the one-dimensional cracked elastic bar is analyzed using the  X-PINN. Various factors affecting the response are explored, including the selected integration method, the enrichment function used, and the architecture of the neural network, specifically, the activation function, the number of layers, and the number of neurons. 

For the standard response calculation, a neural network with two layers, each containing 20 neurons, is utilized. To compute the response in the enriched region, a second network with two layers, each containing 10 neurons, is employed alongside the standard network. A second-order Sawtooth function is used as the enrichment function. Given the inherently stochastic nature of PINN training, a probabilistic approach is adopted: all simulations are repeated 50 times, and the results are reported as averages accompanied by one standard deviation bands.

\begin{figure}[htbp]  
    \centering  
    \begin{minipage}{0.95\textwidth}  
        \centering  
        \begin{minipage}{0.49\linewidth}  
            \begin{minipage}{\linewidth}  
                \includegraphics[width=\linewidth]{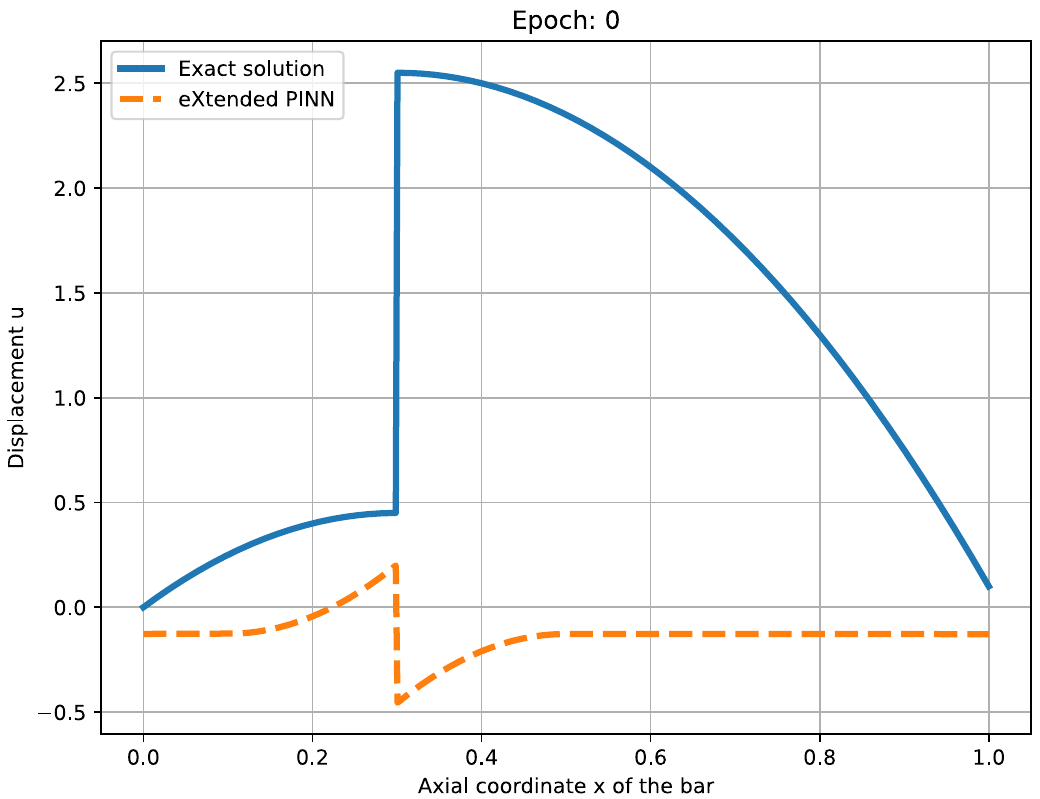}  
                \caption*{(a)}  
            \end{minipage}  
            \begin{minipage}{\linewidth}  
                \includegraphics[width=\linewidth]{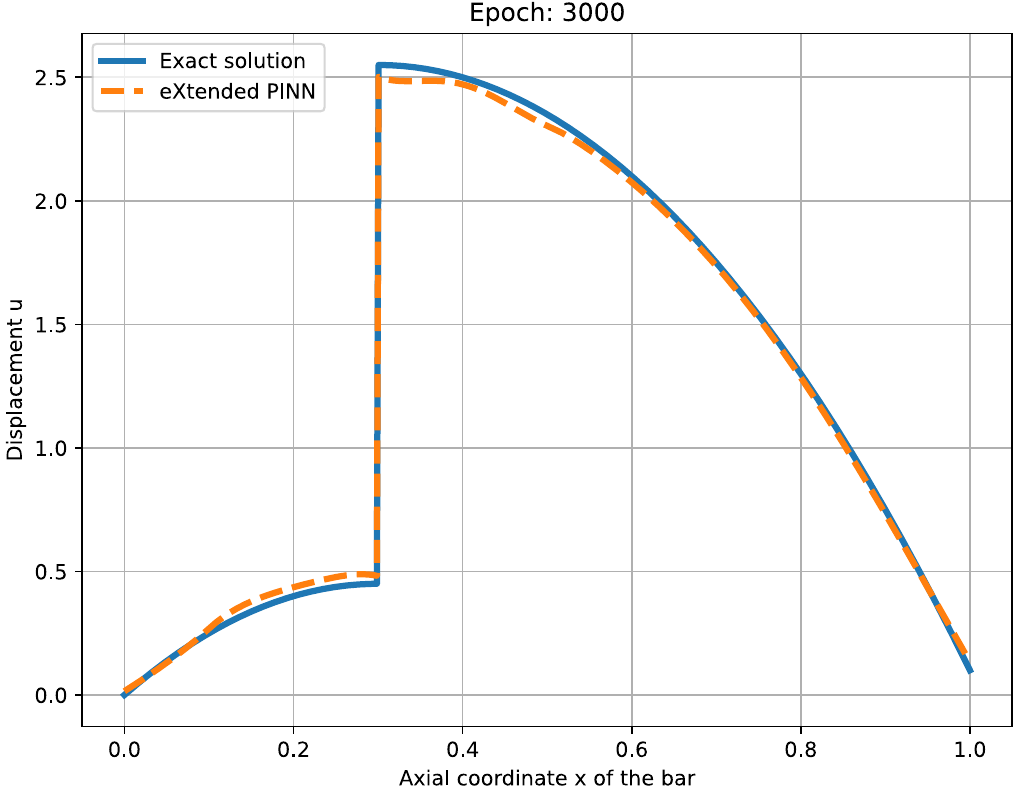}  
                \caption*{(c)}  
            \end{minipage}  
        \end{minipage}  
        \begin{minipage}{0.49\linewidth}  
            \begin{minipage}{\linewidth}  
                \includegraphics[width=\linewidth]{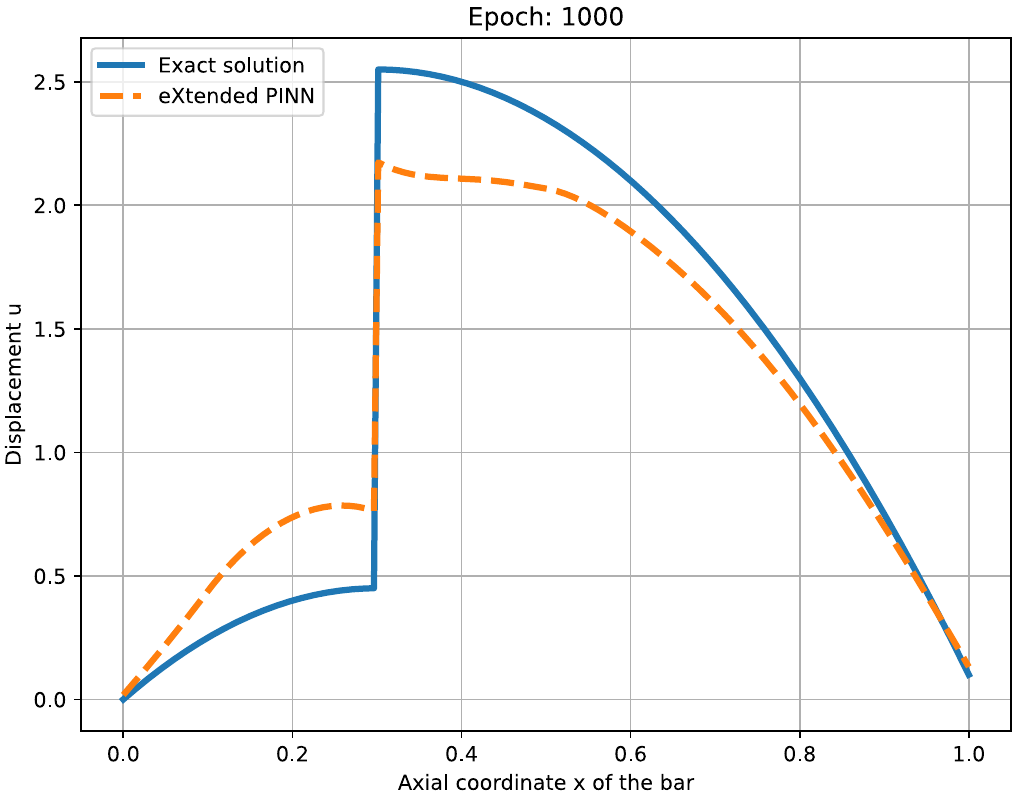}  
                \caption*{(b)}  
            \end{minipage}  
            \begin{minipage}{\linewidth}  
                \includegraphics[width=\linewidth]{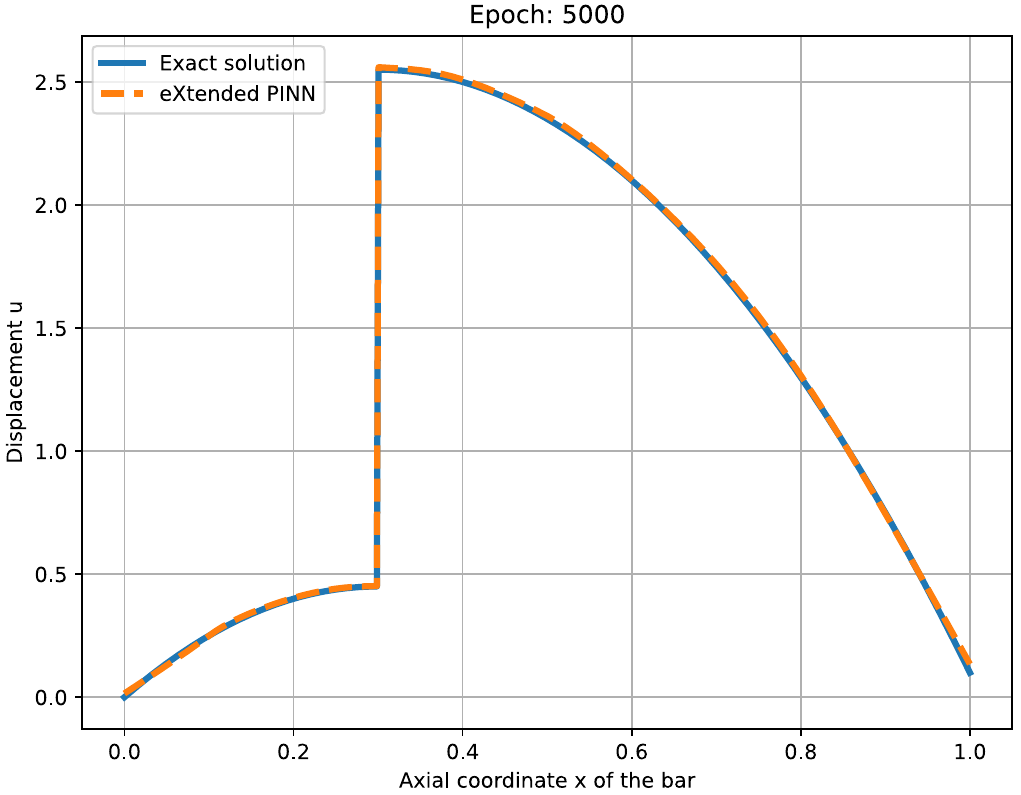}  
                \caption*{(d)}  
            \end{minipage}  
        \end{minipage}  
        \caption{Solutions for the one-dimensional Cracked Bar using UDIPM: (a) Epoch 0, (b) Epoch 1000, (c) Epoch 3000, and (d) Epoch 5000}  
        \label{fig:UDIPM_1D}  
    \end{minipage}  
\end{figure}

\begin{figure}[htbp]  
    \centering  
    \begin{minipage}{0.95\textwidth}  
        \centering  
        \begin{minipage}{0.49\linewidth}  
            \begin{minipage}{\linewidth}  
                \includegraphics[width=\linewidth]{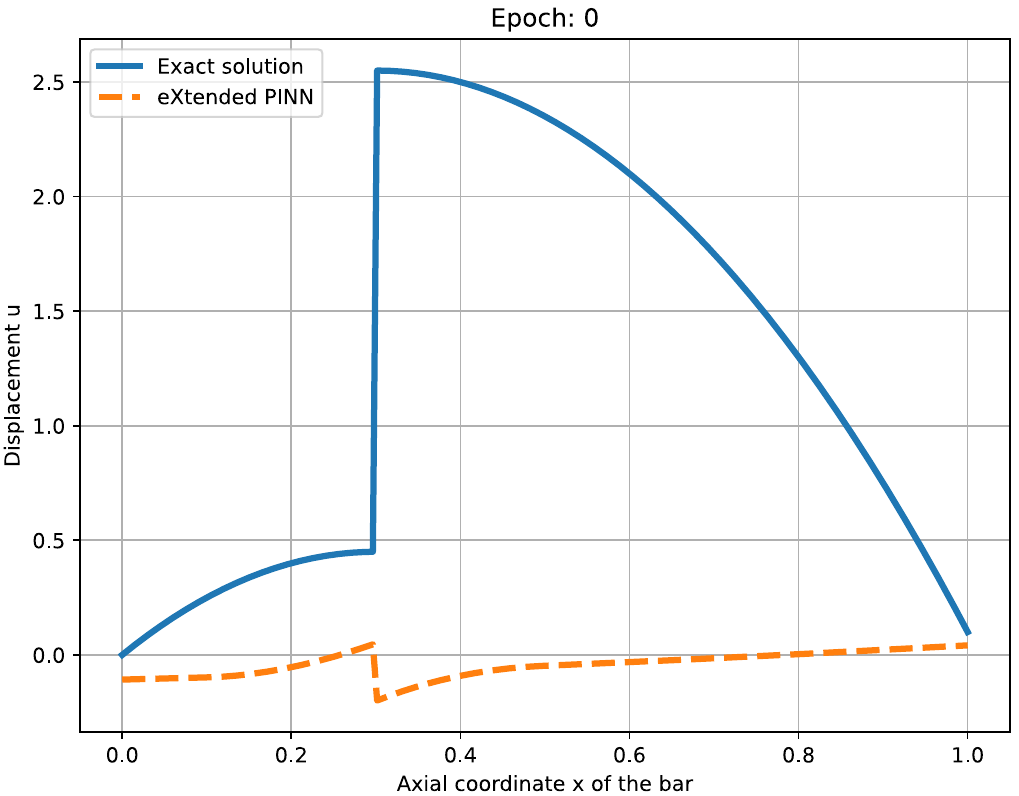}  
                \caption*{(a)}  
            \end{minipage}  
            \begin{minipage}{\linewidth}  
                \includegraphics[width=\linewidth]{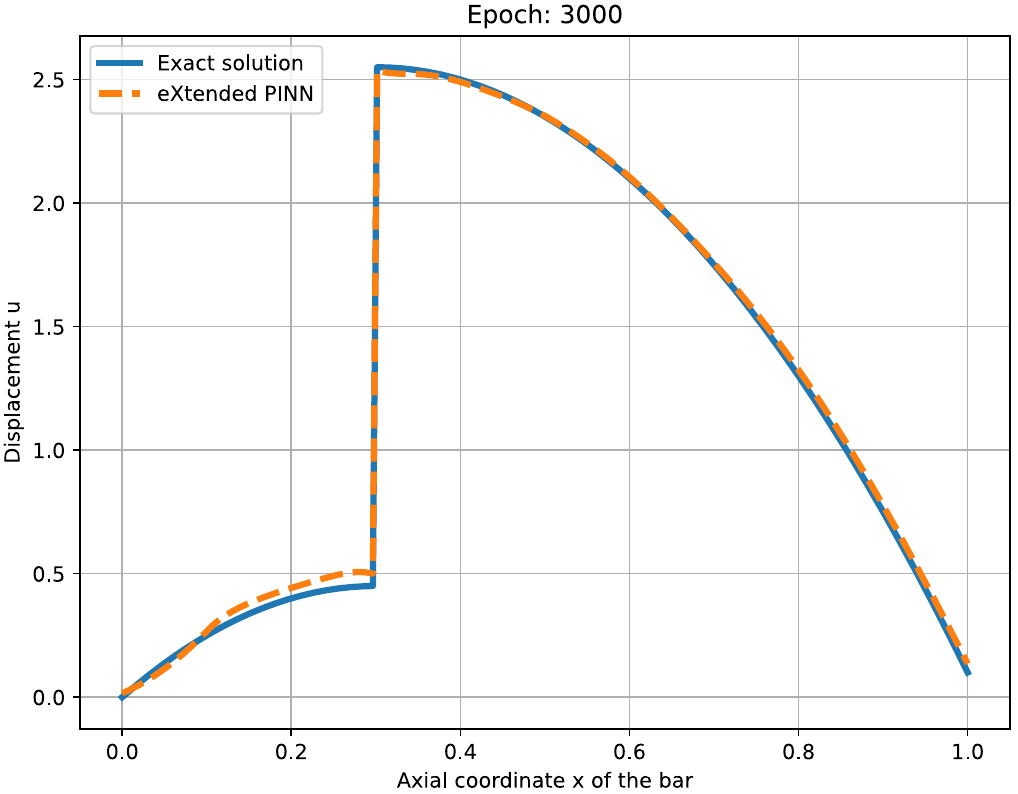}  
                \caption*{(c)}  
            \end{minipage}  
        \end{minipage}  
        \begin{minipage}{0.49\linewidth}  
            \begin{minipage}{\linewidth}  
                \includegraphics[width=\linewidth]{Figures/plot_epoch_1000_CTM.pdf}  
                \caption*{(b)}  
            \end{minipage}  
            \begin{minipage}{\linewidth}  
                \includegraphics[width=\linewidth]{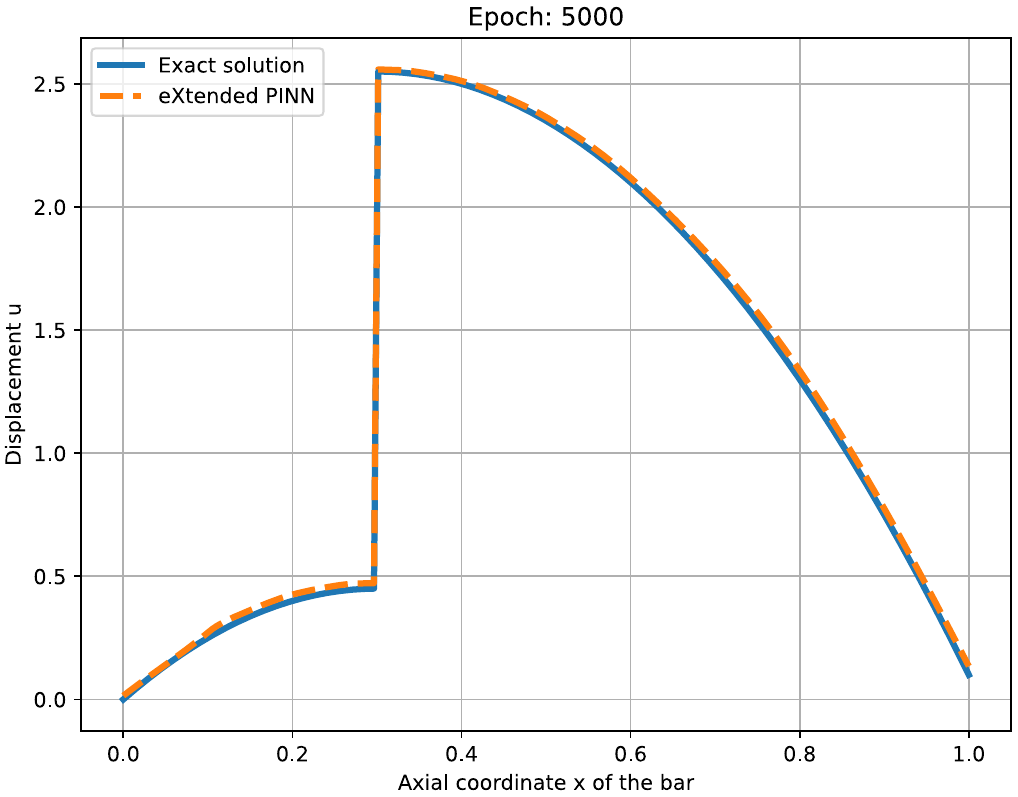}  
                \caption*{(d)}  
            \end{minipage}  
        \end{minipage}  
        \caption{Solutions for the one-dimensional Cracked Bar using CTM: (a) Epoch 0, (b) Epoch 1000, (c) Epoch 3000, and (d) Epoch 5000}  
        \label{fig:CTM_1D}  
    \end{minipage}  
\end{figure}

In the first part, the total potential energy is calculated using UDIPM and CTM integration schemes. The results obtained from these methods at different epochs are presented in Figs.~\hyperref[fig:UDIPM_1D]{\ref*{fig:UDIPM_1D}} and \hyperref[fig:CTM_1D]{\ref*{fig:CTM_1D}}.  

\begin{figure}[htbp]  
    \centering  
    \includegraphics[scale=0.5]{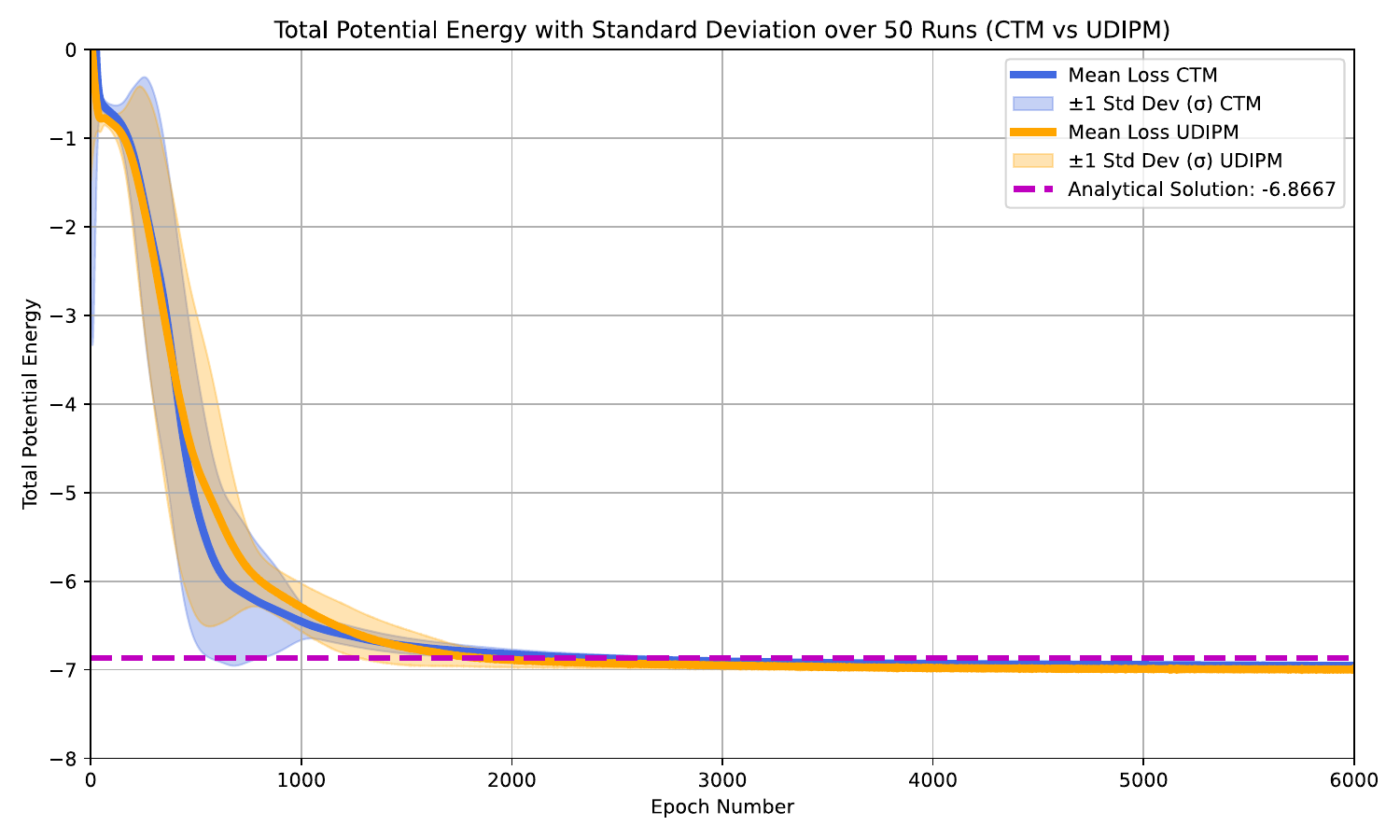}  
    \caption{Total potential energy obtained for UDIPM and CTM integration schemes}  
    \label{fig:lossHistory}  
\end{figure}

As observed, the solutions obtained from both integration methods align closely with the analytical solution. For a comprehensive comparison, Fig.~\hyperref[fig:lossHistory]{\ref*{fig:lossHistory}} presents the convergence of total potential energy over 50 runs for CTM and UDIPM, alongside the analytical value. The results, show that CTM converges more rapidly and consistently, with a narrower deviation indicating higher stability across runs. In contrast, UDIPM converges more slowly and exhibits greater variability, particularly in the early epochs. Despite these differences, both methods ultimately approach the analytical solution of $-6.8667$ with similar final accuracy.

\begin{figure}[h!]  
    \centering  
    \includegraphics[scale=0.5]{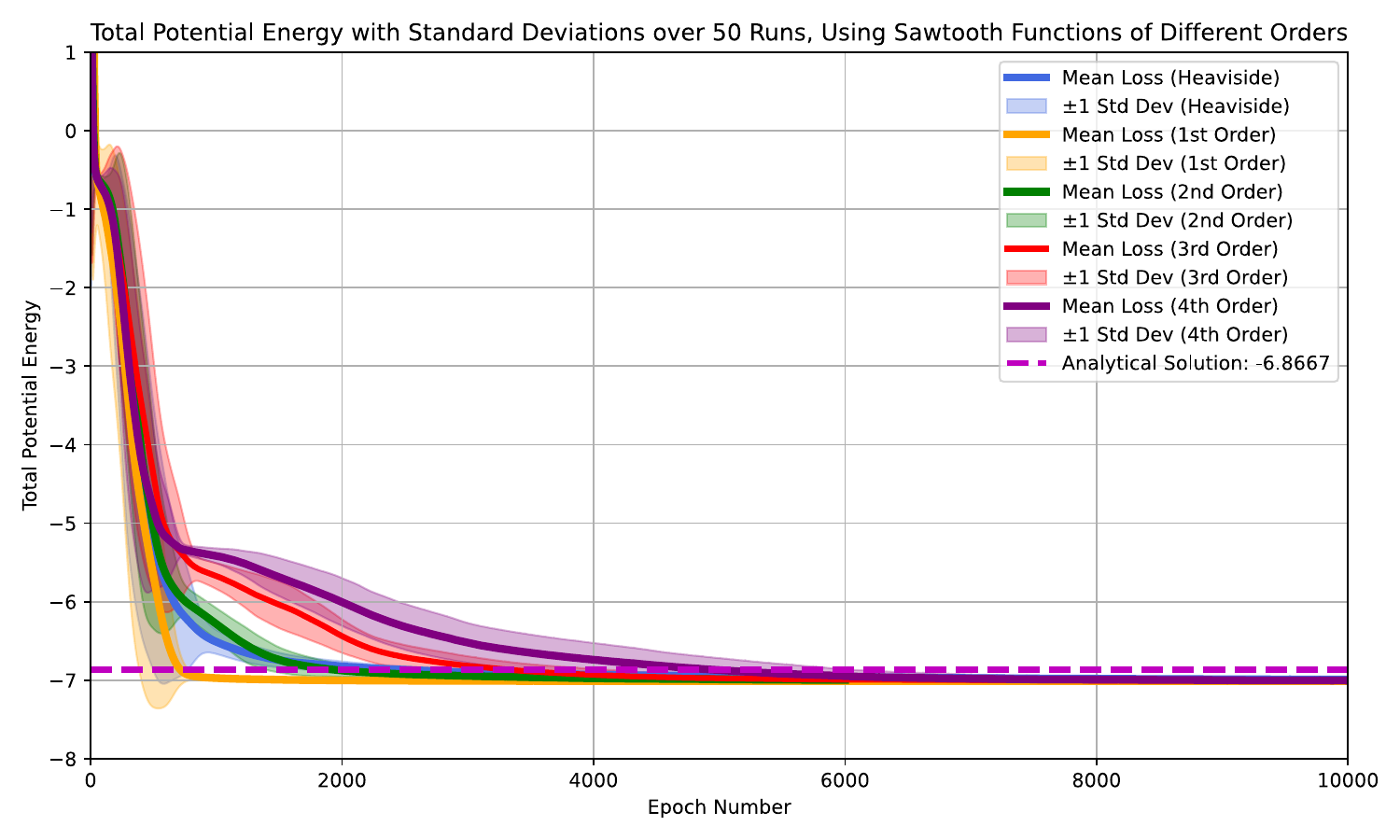}  
    \caption{Total potential energy obtained by different sawtooth enrichment functions}  
    \label{fig:TotalPotential1D_DifferentEnrichments}  
\end{figure}

To further investigate the effect of enrichment functions, the total potential energy was calculated using Sawtooth enrichment functions of various orders. The convergence behaviors of these functions are illustrated in Fig.~\hyperref[fig:TotalPotential1D_DifferentEnrichments]{\ref*{fig:TotalPotential1D_DifferentEnrichments}}. It is observed that the 1st and 2nd order functions exhibit higher convergence rates compared to higher-order functions. However, after approximately 3500 epochs, it becomes evident that the results obtained from all functions converge closely, indicating that the differences between the methods diminish significantly beyond this point.

\begin{figure}[htbp]  
    \centering  
    \includegraphics[scale=0.5]{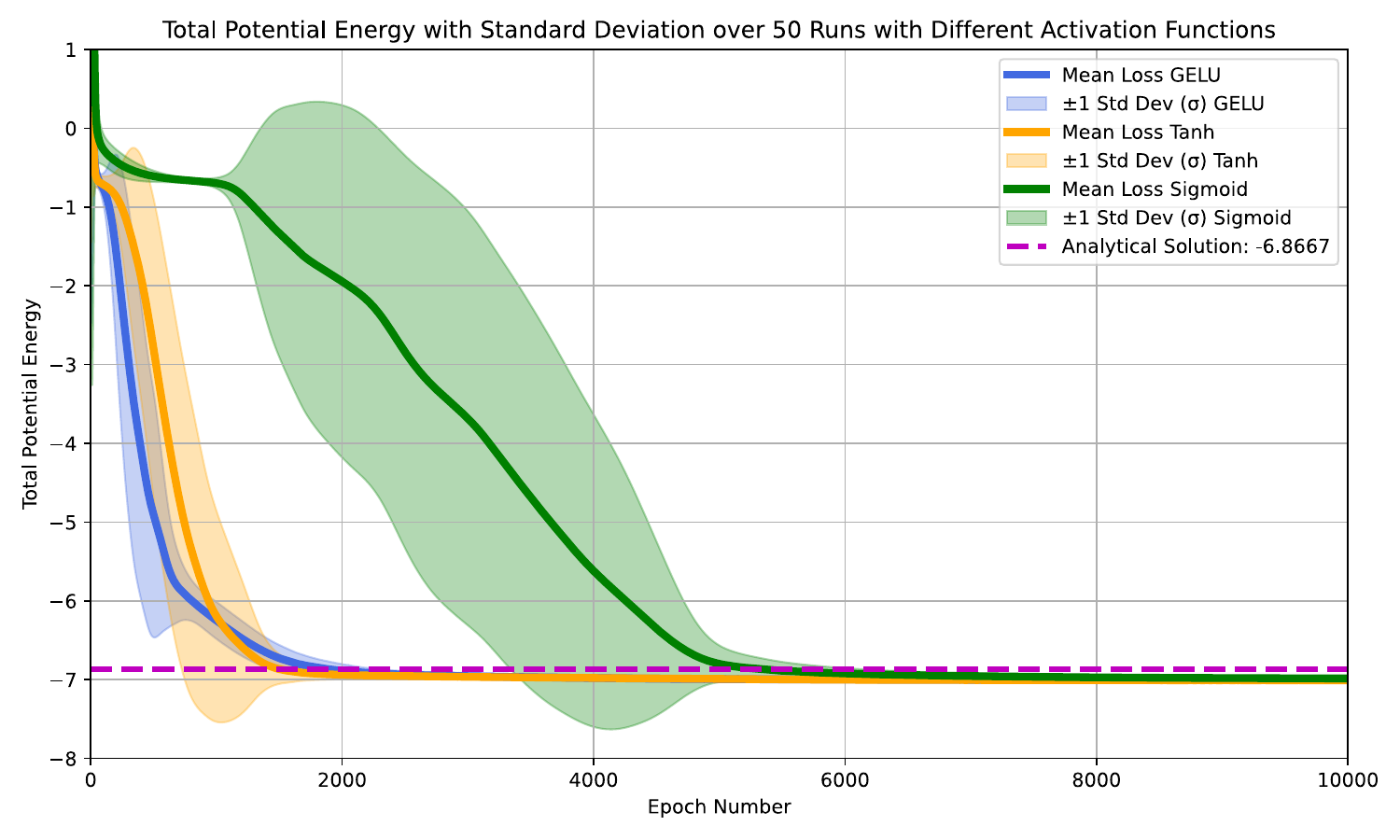}  
    \caption{Total potential energy Obtained by different activation functions}  
    \label{fig:ActicationFunctionComparison}  
\end{figure}

The choice of activation function is widely acknowledged as a critical factor influencing the performance of neural networks. In the numerical experiments conducted, it was found that activation functions such as Tanh, GELU, and Sigmoid produced accurate results, whereas ReLU (Rectified Linear Unit) and HardSwish did not consistently achieve the desired outputs and thus are not employed. Fig.~\hyperref[fig:ActicationFunctionComparison]{\ref*{fig:ActicationFunctionComparison}} illustrates the comparative performance of the activation functions, highlighting their effectiveness in the experiments. Upon comparison, both Tanh and GELU demonstrated superior performance, while Sigmoid converged with less efficacy than Tanh and GELU functions.

\begin{figure}[htbp]  
    \centering  
    \includegraphics[scale=0.5]{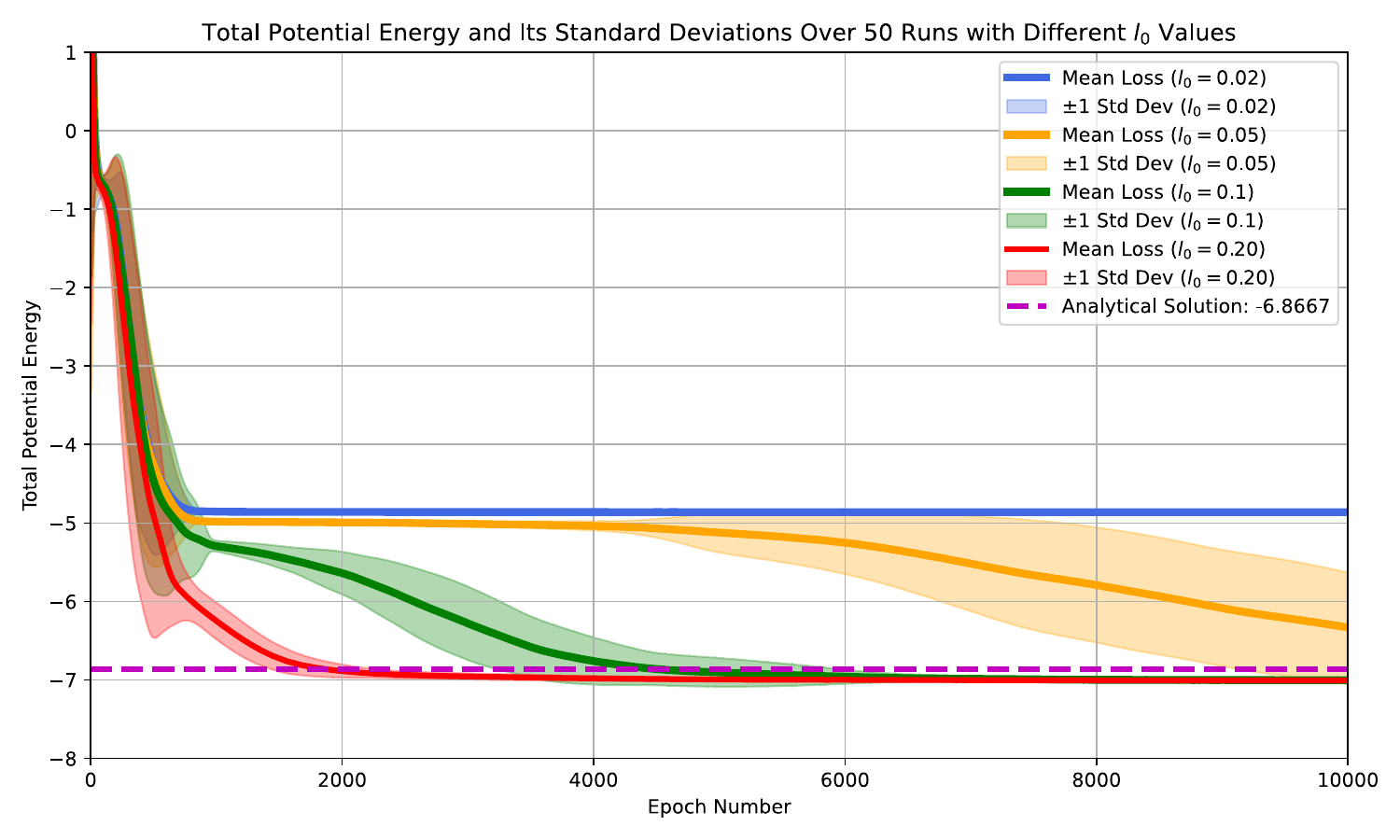}  
    \caption{Total potential energy obtained for different $l_0$ values}  
    \label{fig:l0Comparison}  
\end{figure}

Fig.~\hyperref[fig:l0Comparison]{\ref*{fig:l0Comparison}} shows the evolution of the total potential energy over training epochs for different values of the parameter \( l_0 \) (i.e. width of the enriched domain). Larger values of \( l_0 \), such as 0.1 and 0.2, lead to accurate convergence toward the analytical solution of approximately \(-6.8667\), both showing good performance across runs. Among these, \( l_0 = 0.1 \) provides a strong balance of precision and stability while also defining a smaller domain of influence, helping to reduce computational cost. 

To obtain an acceptable solution, it is crucial to satisfy certain minimum requirements related to the neural network architecture, including the number of training points, hidden layers, and neurons. Although there is no universal guideline, our studies have provided some insight into these parameters for the problem of the cracked one-dimensional elastic bar, as shown in Table \ref{tab:min_requirements}.

\begin{table}[htbp]
    \centering
    \caption{Minimum neural network requirements for a one-dimensional crack problem}
    \renewcommand{\arraystretch}{1.5} 
    \begin{tikzpicture}
        \node[rounded corners=5pt, fill=blue!10, inner sep=10pt] (table) {
            \begin{tabular}{m{8cm} >{\centering\arraybackslash}m{4cm}}
                \toprule
                \rowcolor{blue!25}
                \textbf{} & \textbf{Minimum Requirement} \\
                \midrule
                \rowcolor{blue!10} \text{Minimum Number of Training Points}   & \textbf{50} \\
                \rowcolor{blue!5} \text{Minimum Number of Standard Layers}   & \textbf{2} \\
                \rowcolor{blue!10} \text{Minimum Number of Standard Neurons}  & \textbf{10} \\
                \rowcolor{blue!5} \text{Minimum Number of Enriched Layers}   & \textbf{2} \\
                \rowcolor{blue!10} \text{Minimum Number of Enriched Neurons}  & \textbf{5} \\
                \bottomrule
            \end{tabular}
        };
    \end{tikzpicture}
    \label{tab:min_requirements}
\end{table}

\begin{figure}[htbp]  
    \centering  
    \begin{minipage}{0.95\textwidth}  
        \centering  
        \begin{minipage}{0.49\linewidth}  
            \begin{minipage}{\linewidth}  
                \includegraphics[width=\linewidth,height=4cm]{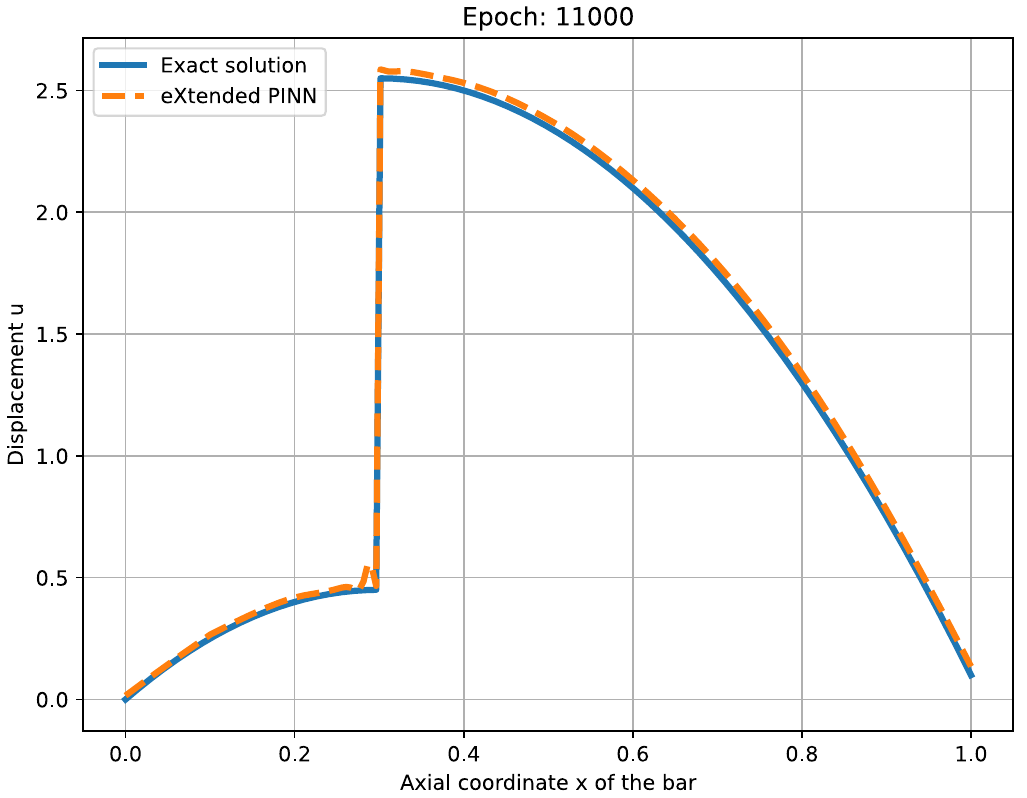}  
                \caption*{(a)}  
            \end{minipage}  
            \begin{minipage}{\linewidth}  
                \includegraphics[width=\linewidth,height=4cm]{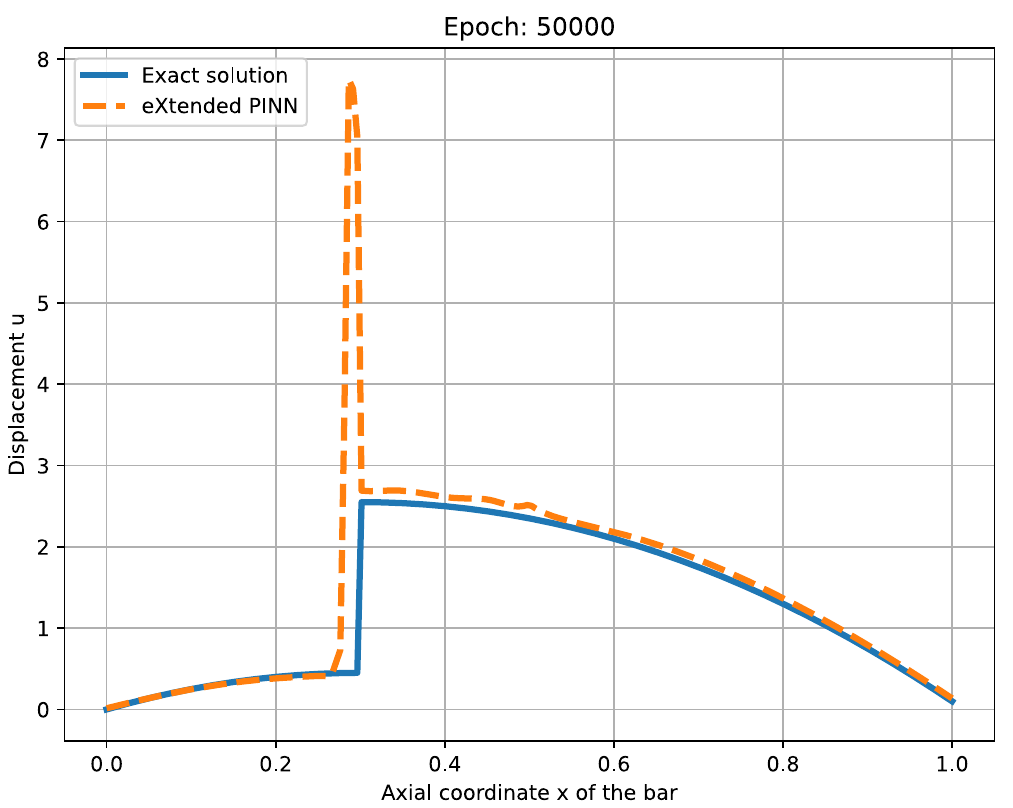}  
                \caption*{(c)}  
            \end{minipage}  
        \end{minipage}  
        \begin{minipage}{0.49\linewidth}  
            \begin{minipage}{\linewidth}  
                \includegraphics[width=\linewidth,height=4cm]{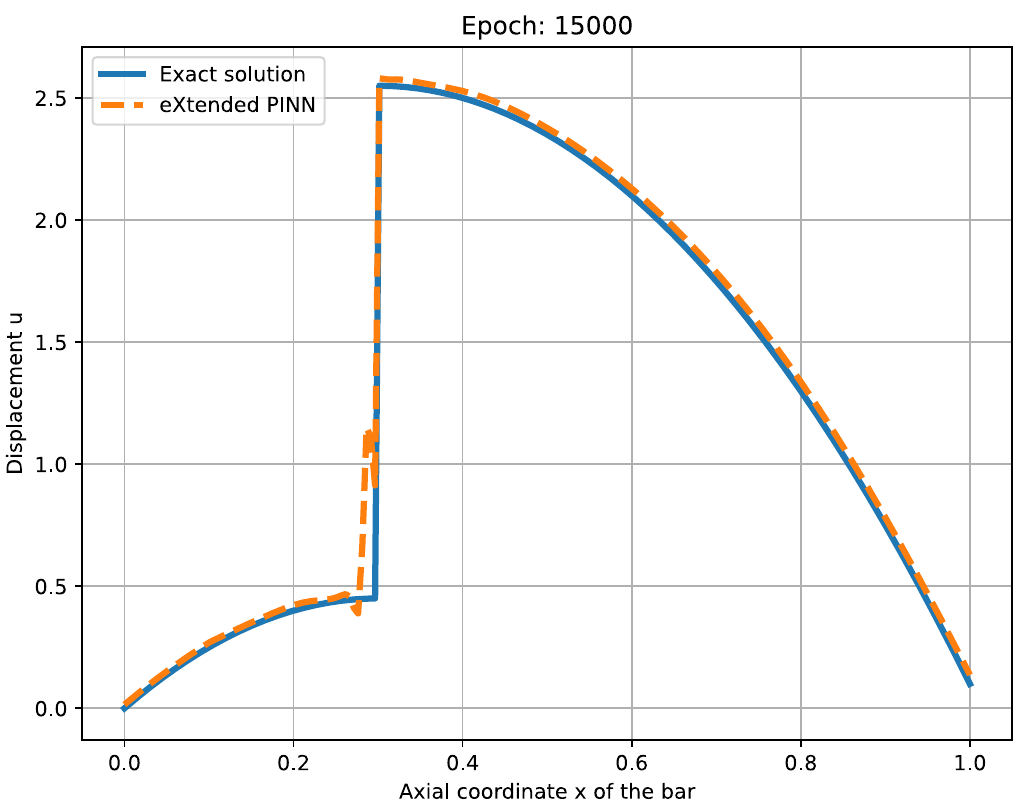}  
                \caption*{(b)}  
            \end{minipage}  
            \begin{minipage}{\linewidth}  
                \includegraphics[width=\linewidth,height=4cm]{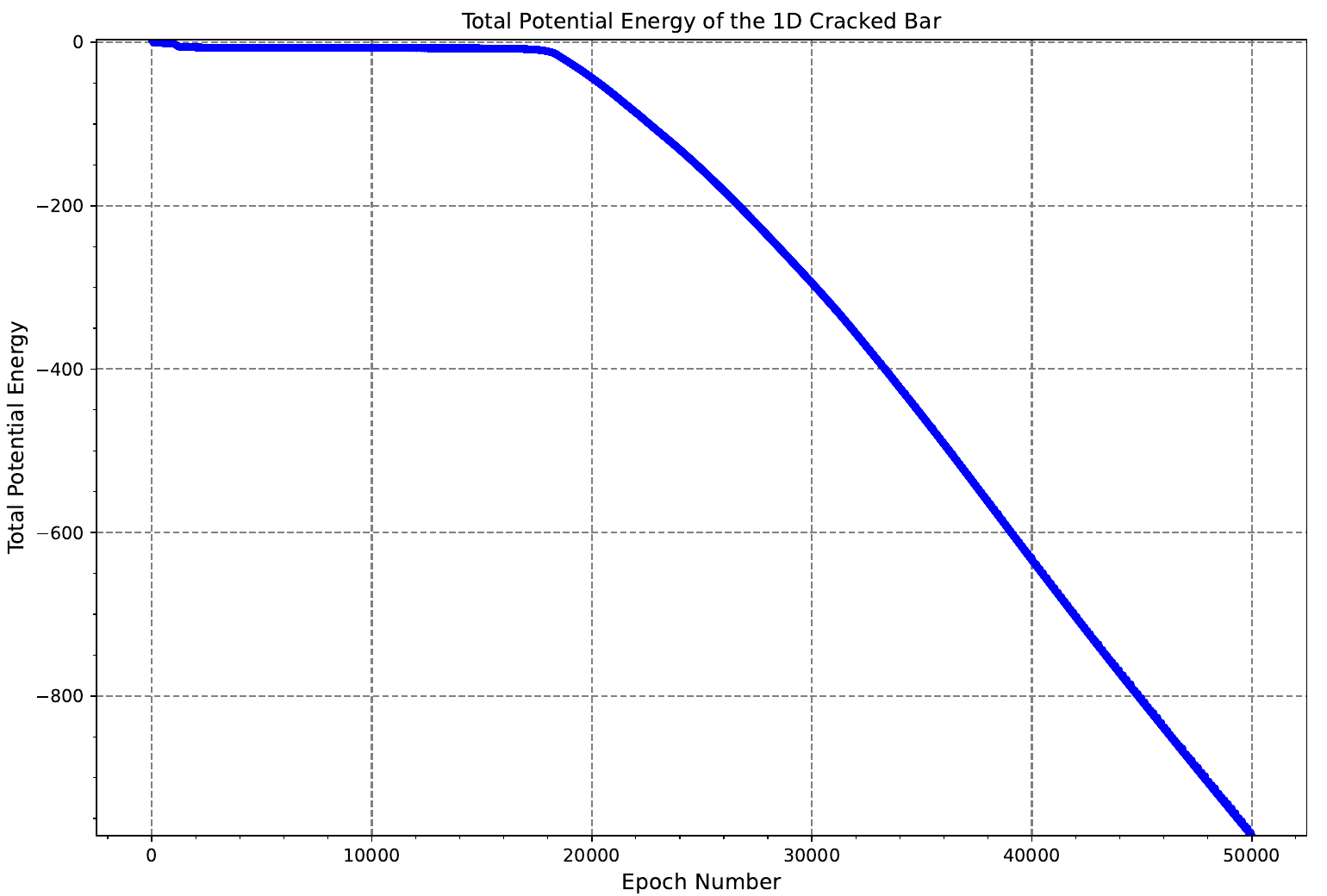}  
                \caption*{(d)}  
            \end{minipage}  
        \end{minipage}  
        \caption{Solutions for the cracked one-dimensional elastic bar: (a) Epoch 11000, (b) Epoch 15000, (c) Epoch 50000, and (d) Total potential energy}  
        \label{fig:instability}  
    \end{minipage}  
\end{figure}

As a final observation, in certain simulations, the X-PINN model for the displacement field in a cracked bar initially converges to the exact solution; however, it begins to exhibit instability around epoch 11,000. As illustrated in Fig.~\hyperref[fig:instability]{\ref*{fig:instability}}, the predicted solution starts to diverge near the crack. By epoch 15,000, errors develop in the displacement field, particularly around the discontinuity, and these errors become increasingly pronounced by epoch 50,000. As shown in the total potential energy plot, the energy remains nearly constant at first, indicating stable and consistent network predictions. However, after approximately 15,000 epochs, a gradual and sustained decrease begins, signaling the onset of instability in the training process. These observations highlight the importance of incorporating training control strategies, such as early stopping or regularization, to ensure the accuracy and stability of the model.

\subsection{2D Center Cracked plate}
\begin{figure}[htbp]
    \centering
\includegraphics[scale=0.3]{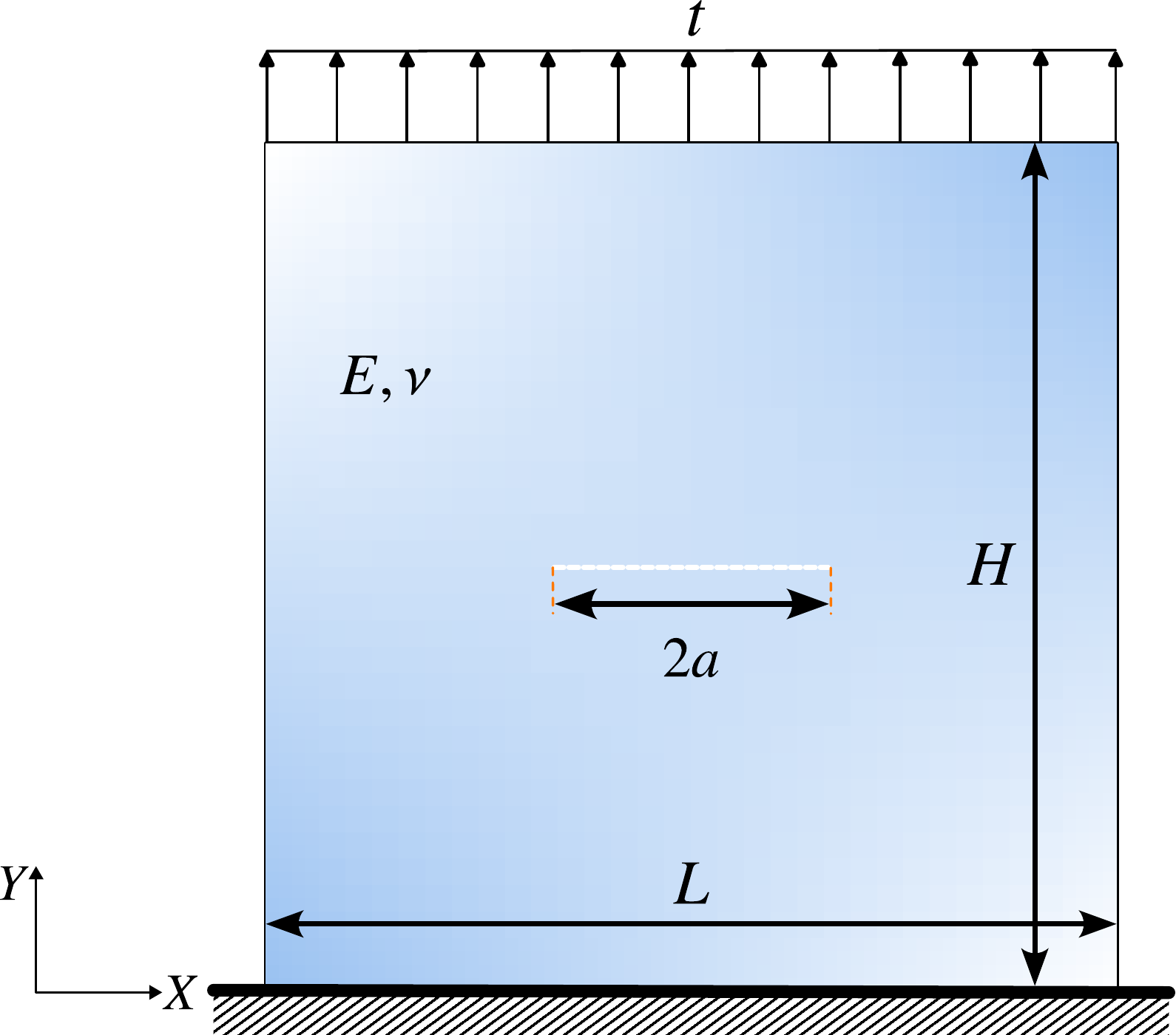}
    \caption{Geometry and boundary conditions for the 2D plate with center crack problem}
    \label{fig:centerCrack}
\end{figure}

In this section, the analysis of a 2D center-cracked plate is presented using X-PINN. The geometry and boundary conditions are shown in Fig.~\hyperref[fig:centerCrack]{\ref{fig:centerCrack}}. For simplicity, the plate's length \(L\) and height \(H\), as well as the modulus of elasticity \(E\), are all considered equal to one. The crack width, \(2a\), is set to 0.3. Additionally, Poisson's ratio \(\nu\) is set to 0.3, and the applied traction \(t\) is set to 0.1. A neural network consisting of 10 layers, each with 20 neurons, is employed to compute the continuous component of the response in each direction. An enriched network is used to calculate the discontinuous component of the response, with 10 layers, each containing 10 neurons. Furthermore, the width of the enrichment region, \(l_0\), is set to 0.1. For integration using CTM, a total of 26,520 integration points were generated, of which 5,400 were located in the enrichment domain.

\begin{figure}[htbp]  
    \centering  
    \begin{minipage}{0.95\textwidth}  
        \centering  
        \begin{minipage}{0.49\linewidth}  
            \begin{minipage}{\linewidth}  
                \includegraphics[height=0.85\textwidth]{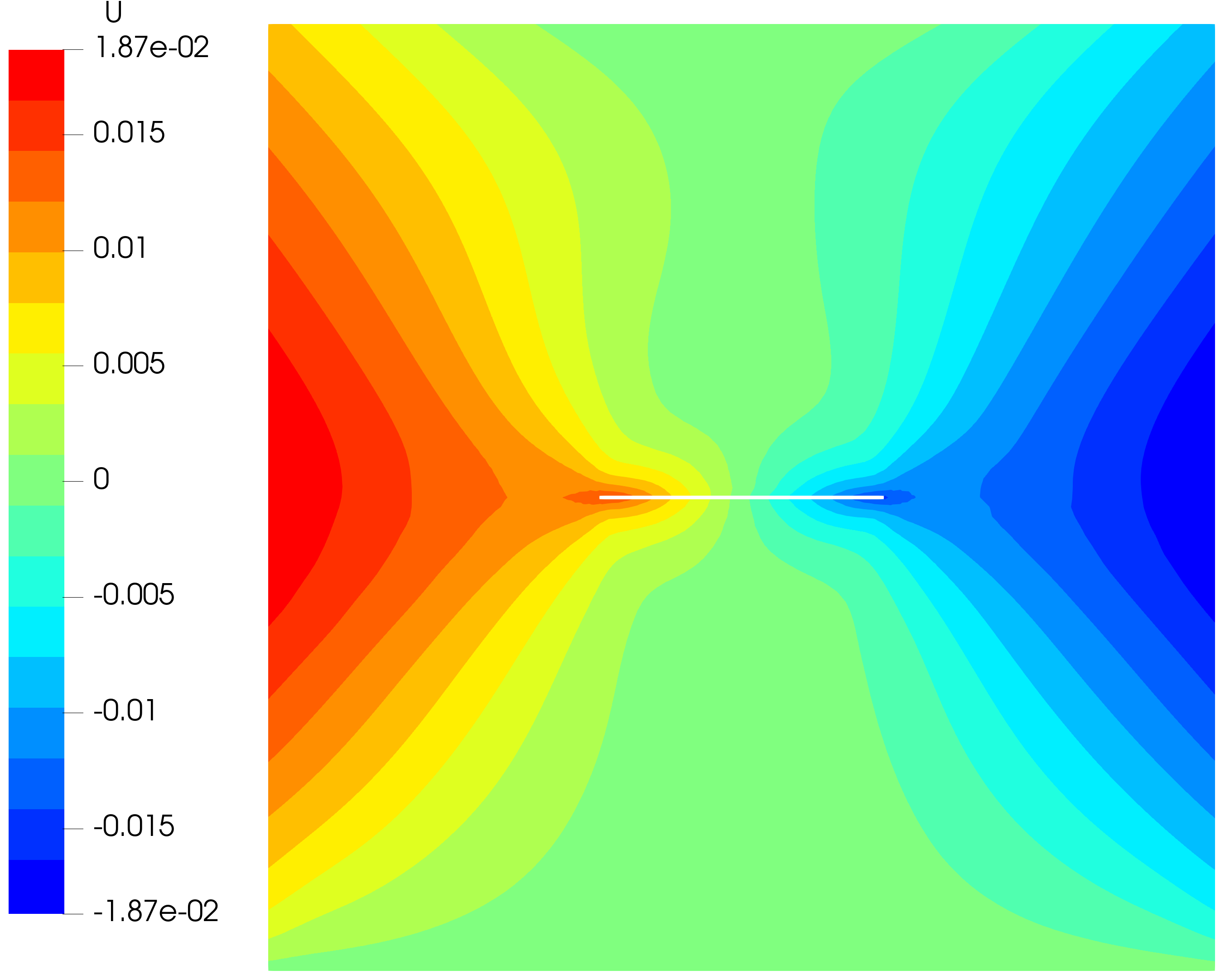}  
                \caption*{(a)}  
            \end{minipage}  
            \begin{minipage}{\linewidth}  
                \includegraphics[height=0.85\textwidth]{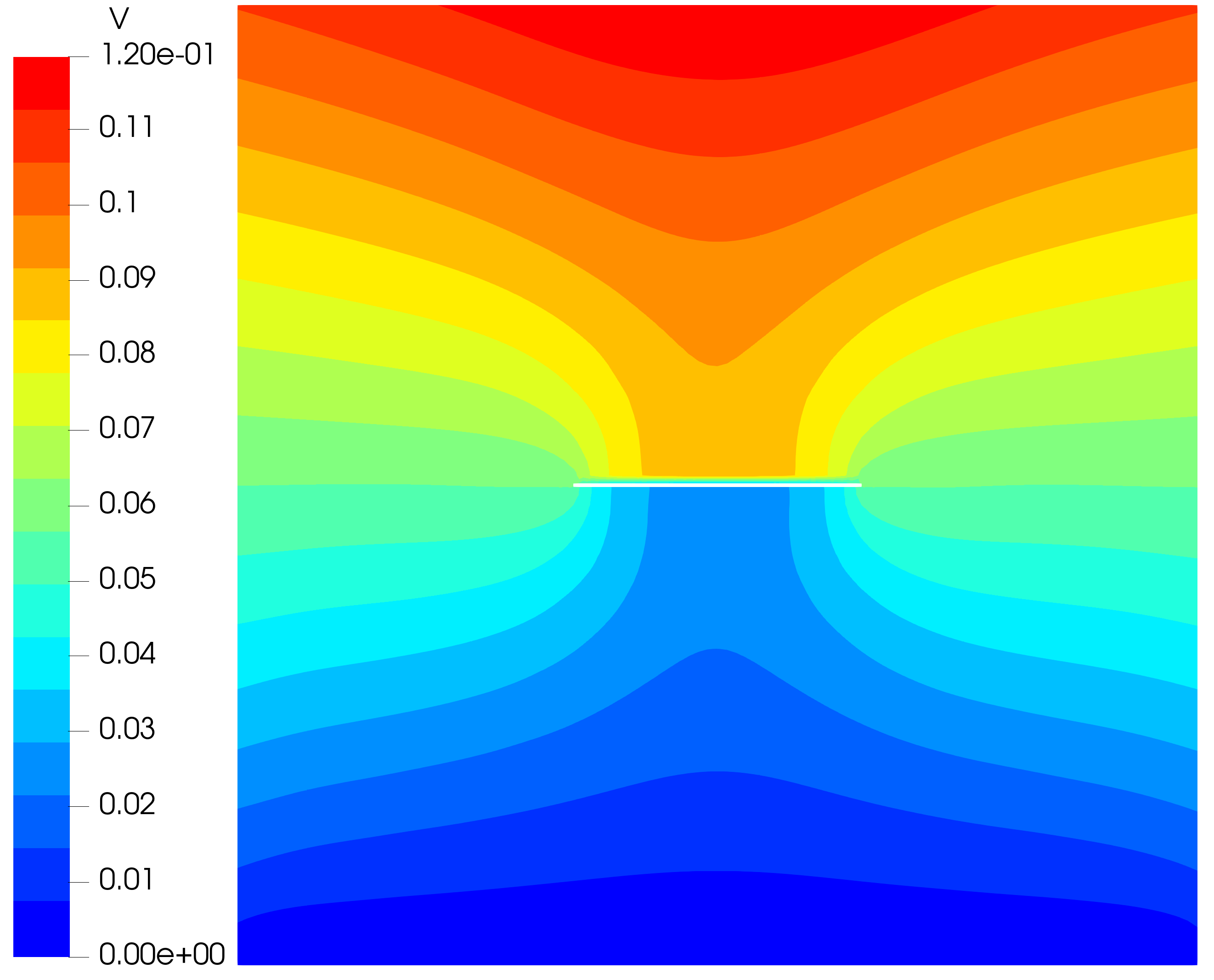}  
                \caption*{(c)}  
            \end{minipage}
        \end{minipage}  
        \begin{minipage}{0.49\linewidth}  
            \begin{minipage}{\linewidth}  
                \includegraphics[height=0.85\textwidth]{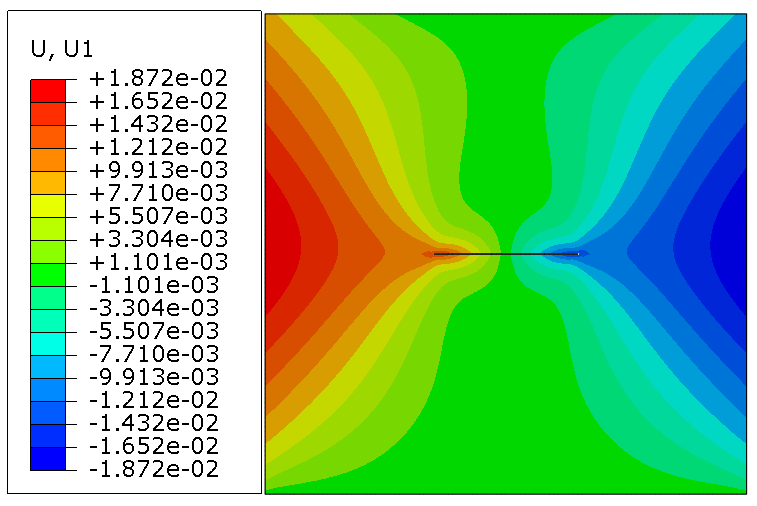}  
                \caption*{(b)}  
            \end{minipage}  
            \begin{minipage}{\linewidth}  
                \includegraphics[height=0.85\textwidth]{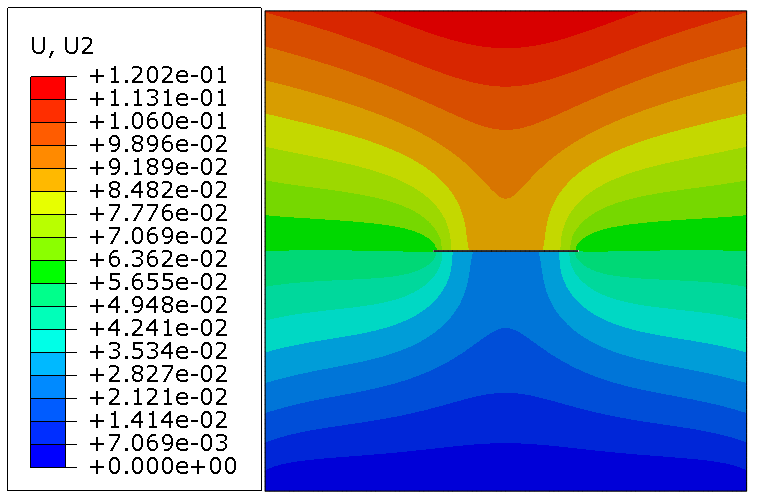}  
                \caption*{(d)}  
            \end{minipage}  
        \end{minipage}  
        \caption{Displacement contours of the Center Crack problem: (a) U using X-PINN, (b) U using Abaqus, (c) V using X-PINN, and (d) V using Abaqus}  
        \label{fig:centerCrackDisplacements}  
    \end{minipage}  
\end{figure}

\begin{figure}[htbp]  
    \centering  
    \begin{minipage}{0.95\textwidth}  
        \centering  
        \begin{minipage}{0.49\linewidth}  
            \begin{minipage}{\linewidth}  
                \includegraphics[height=0.83\textwidth]{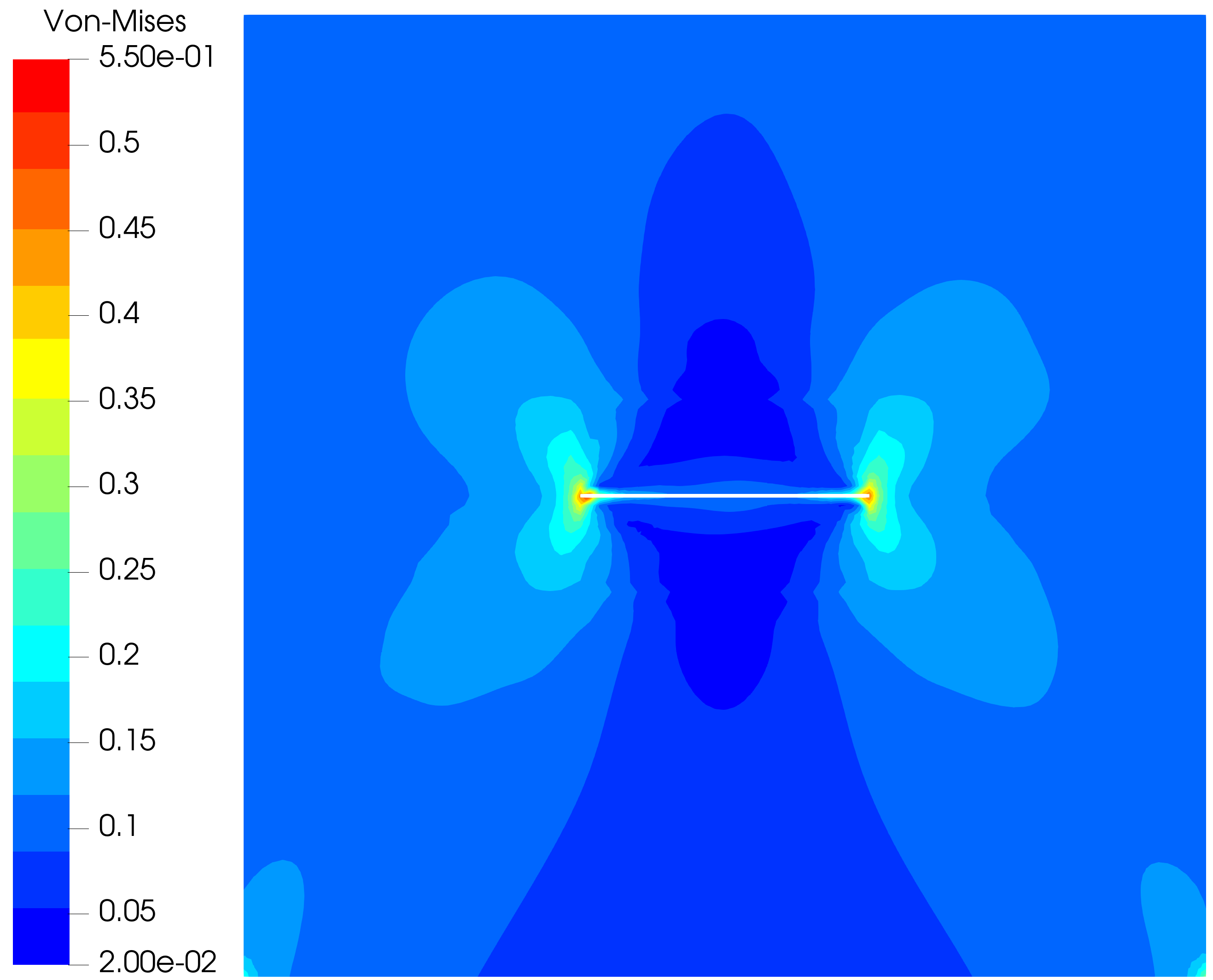}  
                \caption*{(a)}  
            \end{minipage}  
        \end{minipage}  
        \begin{minipage}{0.49\linewidth}  
            \begin{minipage}{\linewidth}  
                \includegraphics[height=0.83\textwidth]{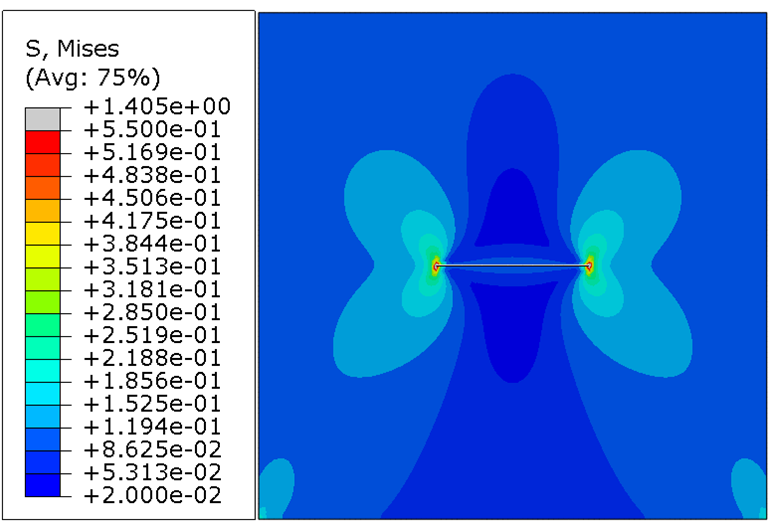} 
                \caption*{(b)}  
            \end{minipage}   
        \end{minipage}  
        \caption{von-Mises stress of Center the Crack problem: (a) Using X-PINN, (b) Using Abaqus}  
        \label{fig:centerCrackVon-Mises}  
    \end{minipage}  
\end{figure}

The displacement contours for each direction, along with the von-Mises stress contours obtained from the X-PINN method, are presented alongside the results from the Abaqus software in Fig.~\hyperref[fig:centerCrackDisplacements]{\ref{fig:centerCrackDisplacements}} and Fig.~\hyperref[fig:centerCrackVon-Mises]{\ref{fig:centerCrackVon-Mises}}. The results obtained from the X-PINN method demonstrate a remarkable consistency with those produced by Abaqus software, highlighting the effectiveness of the X-PINN approach.

Various analyses were conducted to determine the appropriate width of the enrichment region, \( l_0 \), for different lengths of 0.01, 0.05, 0.1, and 0.2. The findings indicate that as the width of the enrichment region decreases, the number of epochs required for the enrichment neural network to effectively learn the discontinuous parts increases, leading to longer training times for the neural network. Based on these analyses, a width of \( l_0 = 0.1 \) is deemed appropriate. Therefore, in this example and subsequent examples, the width of the enrichment region will be set to \( l_0 = 0.1 \).

\begin{figure}[htbp]  
    \centering  
    \includegraphics[scale=0.4]{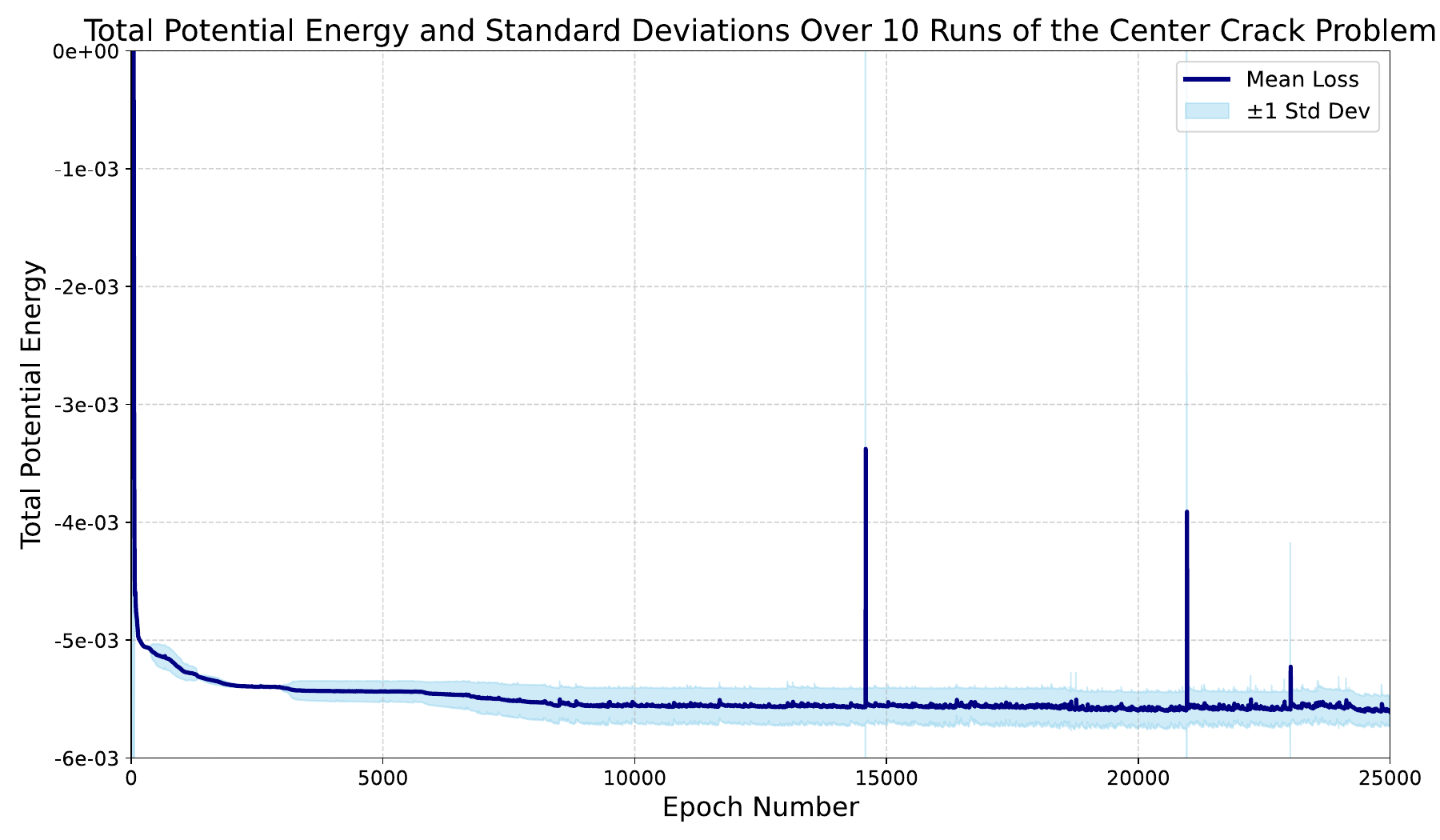}  
    \caption{Total potential energy across 10 runs of the Center Crack problem}  
    \label{fig:lossHistoryCenterCrack}  
\end{figure}

Fig.~\hyperref[fig:lossHistoryCenterCrack]{\ref{fig:lossHistoryCenterCrack}} illustrates the convergence behavior of the total potential energy over 10 independent runs of the center crack problem, tracked over 25,000 epochs. The plot shows the mean total potential energy and its variability, represented by ±1 standard deviation. The sharp initial decrease followed by a plateau indicates effective early optimization, with subsequent epochs maintaining a relatively stable energy level. Occasional spikes in the standard deviation highlight isolated runs with divergence or instability, although the overall trend suggests consistent convergence. These results confirm the robustness of the training procedure despite some stochastic fluctuations.

\subsection{Edge Crack Analysis}
\begin{figure}[htbp]
    \centering
\includegraphics[scale=0.3]{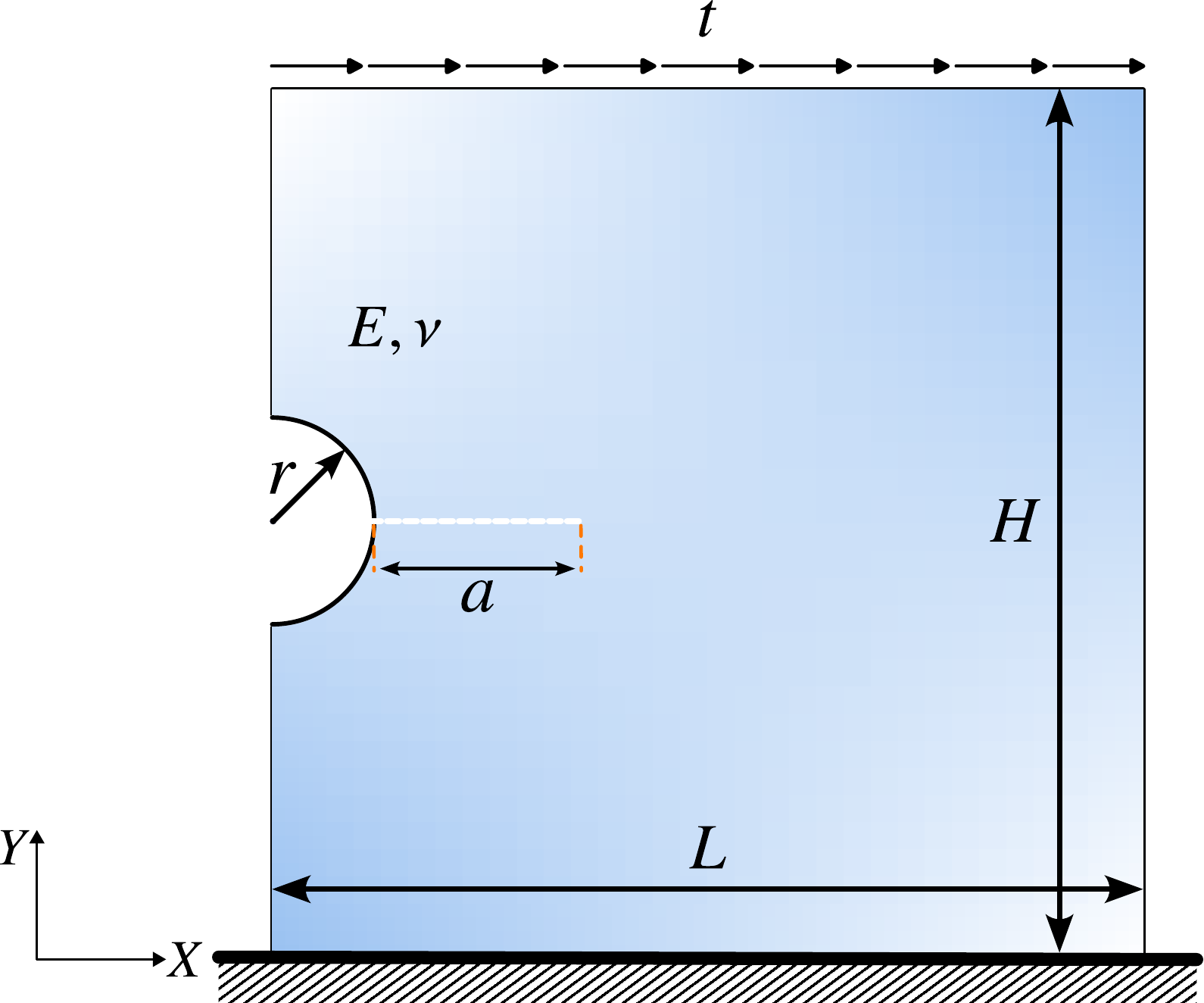}
    \caption{Geometry and boundary conditions for the 2D plate with edge crack problem}
    \label{fig:edgeCrack}
\end{figure}

As another benchmark example, this section is focused on the analysis of a 2D edge cracked, emanating from a semicircular hole in a rectangular plate using the X-PINN methodology. The configuration and boundary conditions are illustrated in Fig.~\hyperref[fig:edgeCrack]{\ref{fig:edgeCrack}}. The dimensions of the plate, \( L \) and \( H \), along with the modulus of elasticity, \( E \), are all assigned a value of one. Additionally, the Poisson's ratio, \( \nu \), is set to 0.3, and a shear traction , \( t \), of magnitude 0.1 is applied to the top edge. The crack length, \( a \), is established as 0.2, and hole radius is (\r=0.1). For this problem a standard neural network architecture is utilized, comprising 15 layers with 20 neurons in each layer, to determine the continuous response component of each direction. In addition, an enriched network is employed to assess the discontinuous response component, featuring 10 layers, each containing 20 neurons. In the integration process utilizing CTM, a total of 24846 integration points were established, with 3600 of these points situated within the enrichment domain.

\begin{figure}[htbp]  
    \centering  
    \begin{minipage}{0.95\textwidth}  
        \centering  
        \begin{minipage}{0.49\linewidth}  
            \begin{minipage}{\linewidth}  
                \includegraphics[height=0.8\textwidth]{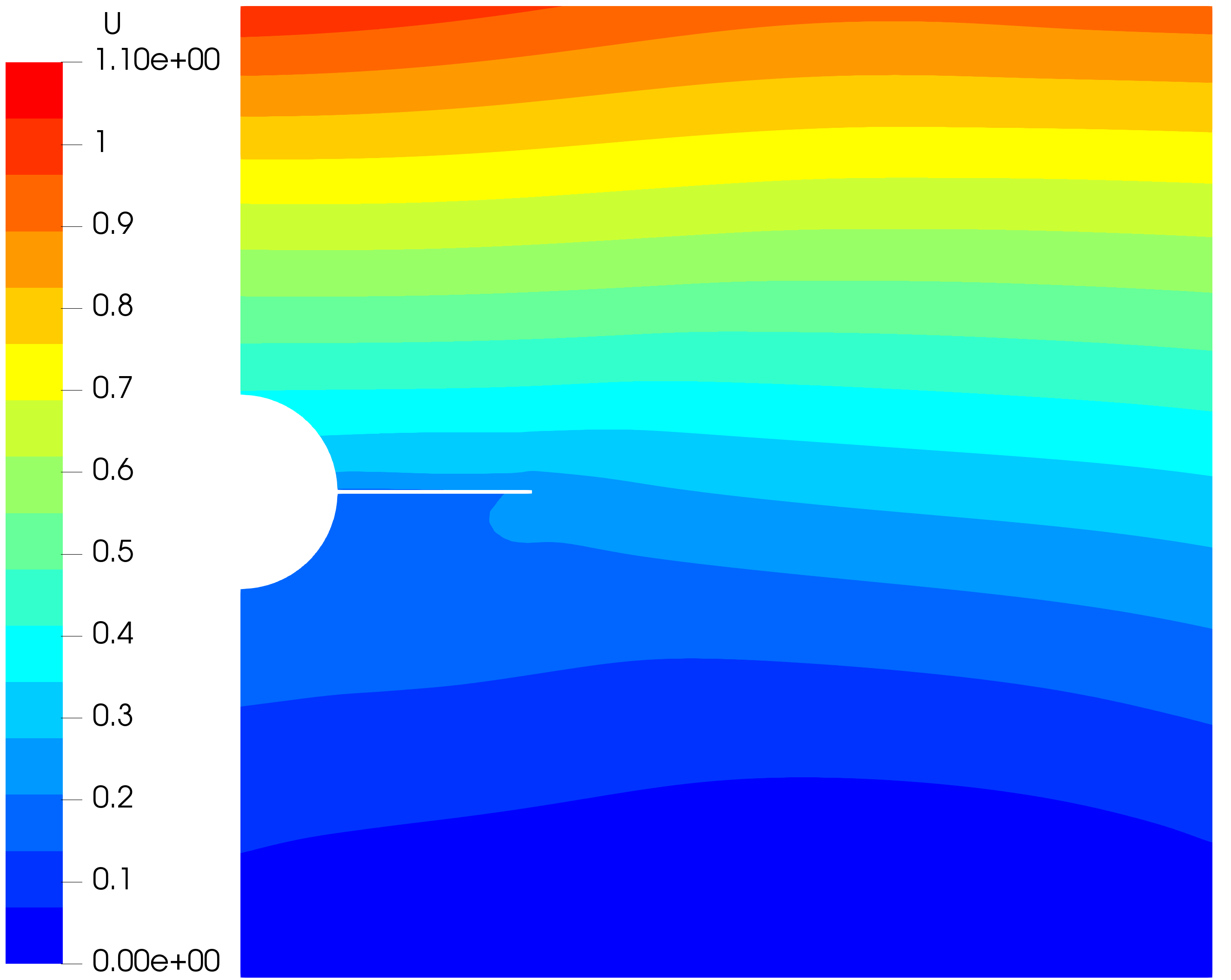}  
                \caption*{(a)}  
            \end{minipage}  
            \begin{minipage}{\linewidth}  
                \includegraphics[height=0.8\textwidth]{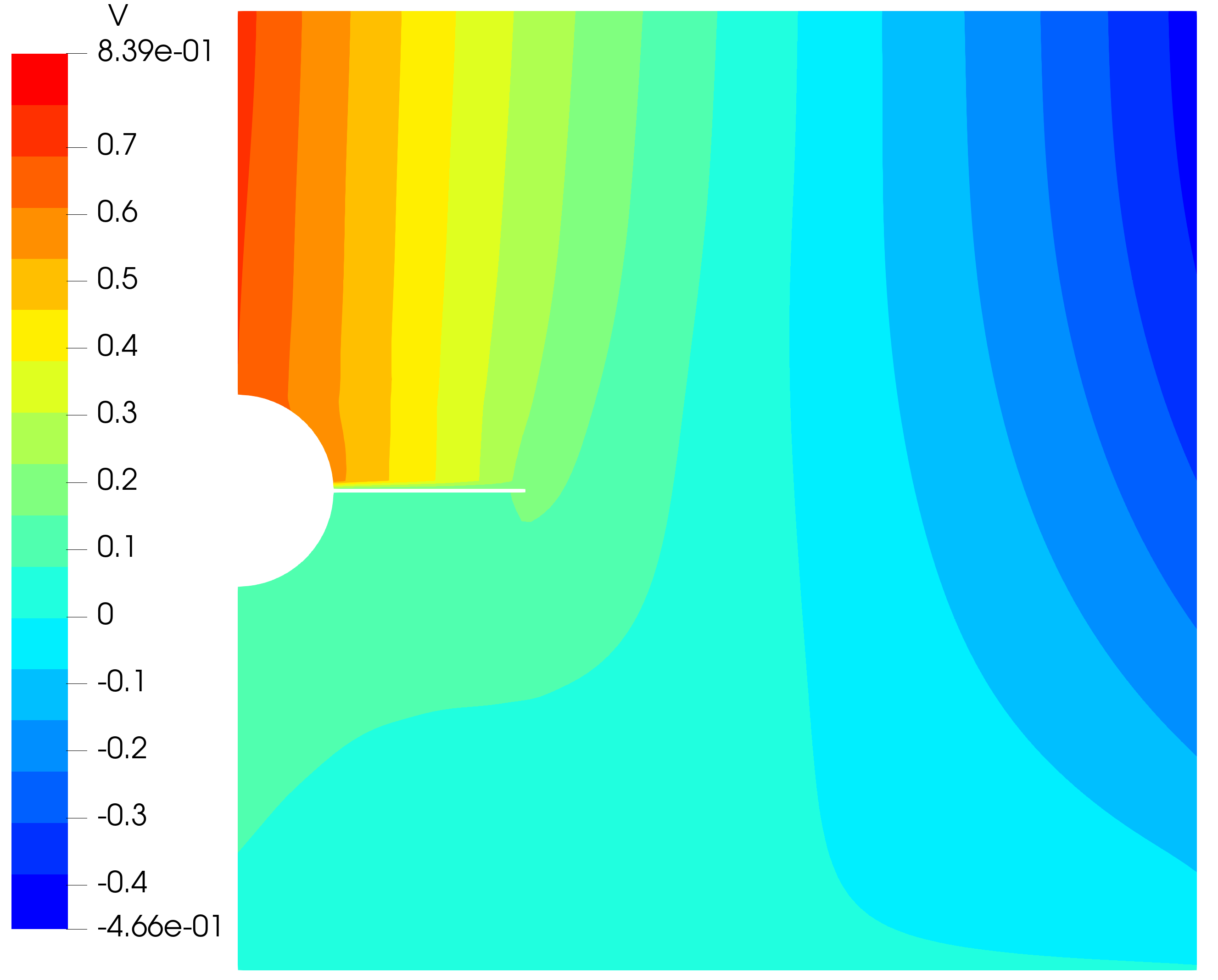}  
                \caption*{(c)}  
            \end{minipage}
        \end{minipage}  
        \begin{minipage}{0.49\linewidth}  
            \begin{minipage}{\linewidth}  
                \includegraphics[height=0.8\textwidth]{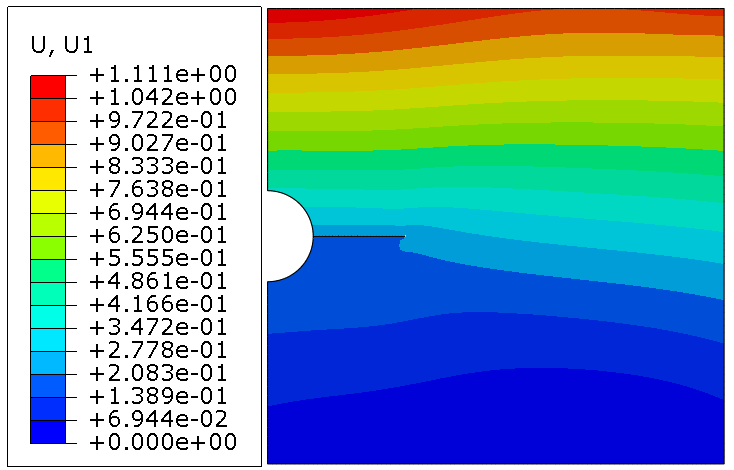}  
                \caption*{(b)}  
            \end{minipage}  
            \begin{minipage}{\linewidth}  
                \includegraphics[height=0.8\textwidth]{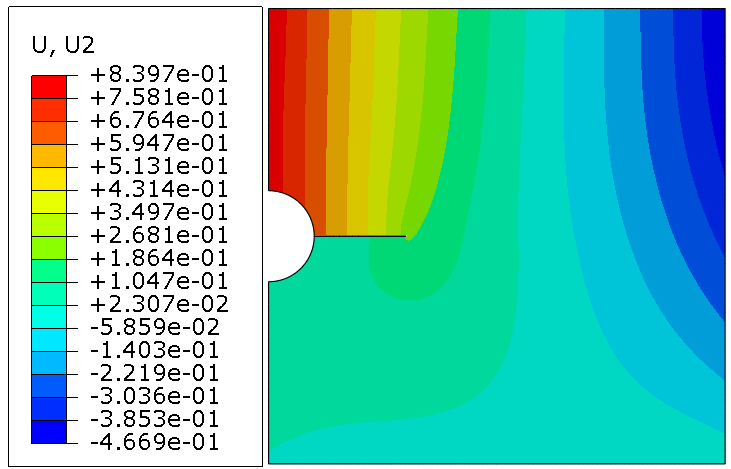}  
                \caption*{(d)}  
            \end{minipage}  
        \end{minipage}  
        \caption{Displacement contours of the Edge Crack problem: (a) U using X-PINN, (b) U using Abaqus, (c) V using X-PINN, and (d) V using Abaqus}  
        \label{fig:edgeCrackDisp}  
    \end{minipage}  
\end{figure}

\begin{figure}[htbp]  
    \centering  
    \begin{minipage}{0.95\textwidth}  
        \centering  
        \begin{minipage}{0.49\linewidth}  
            \begin{minipage}{\linewidth}  
                \includegraphics[height=0.85\textwidth]{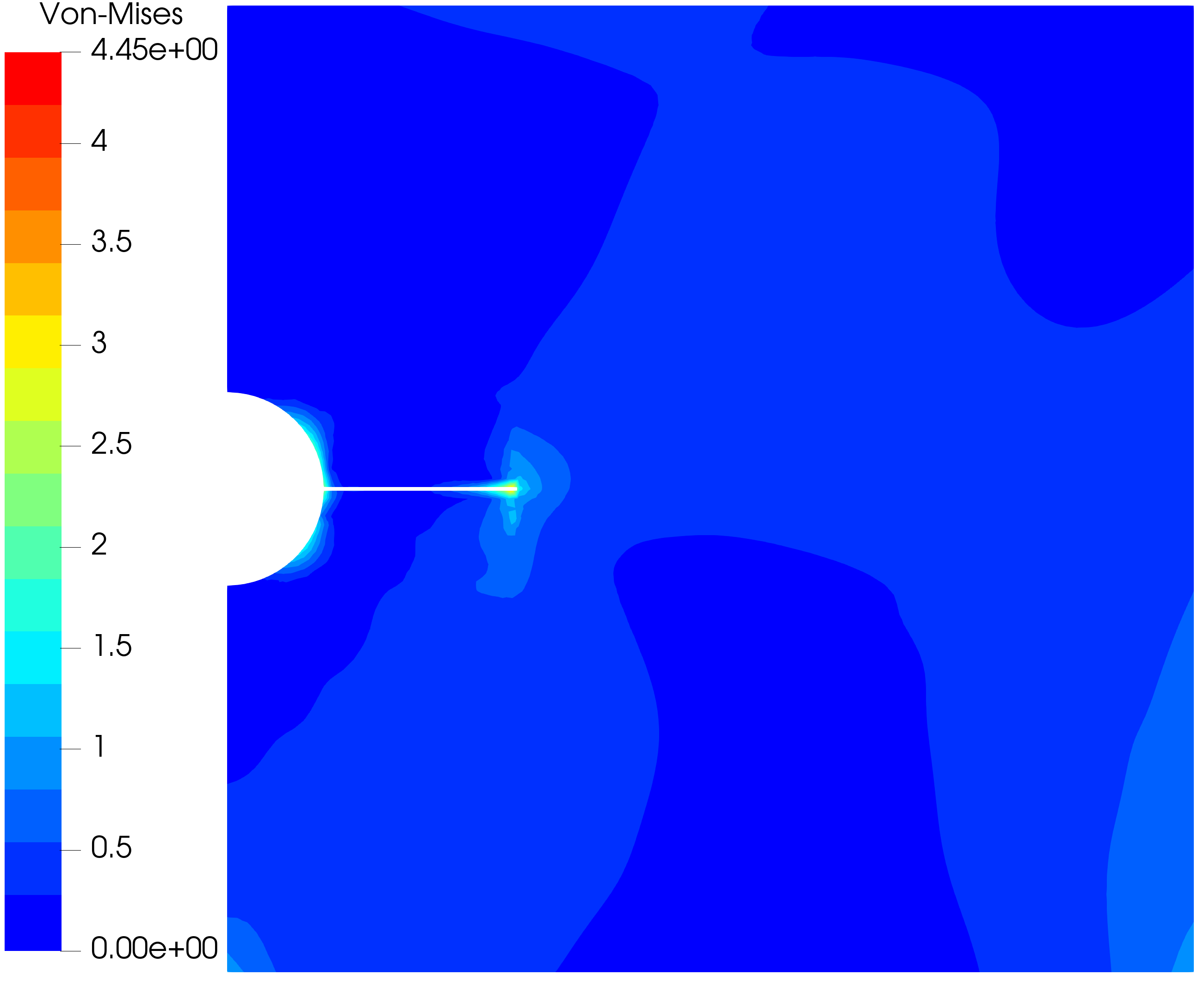}  
                \caption*{(a)}  
            \end{minipage}  
        \end{minipage}  
        \begin{minipage}{0.49\linewidth}  
            \begin{minipage}{\linewidth}  
                \includegraphics[height=0.85\textwidth]{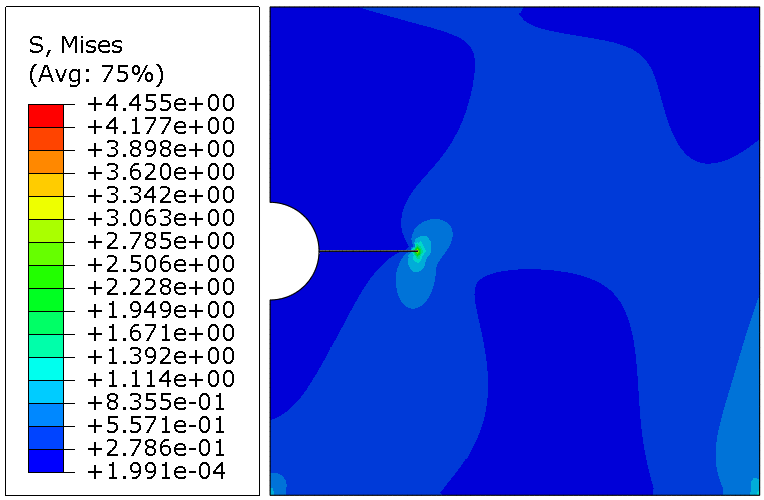} 
                \caption*{(b)}  
            \end{minipage}   
        \end{minipage}  
        \caption{von-Mises stress of the Edge Crack problem: (a) Using X-PINN, (b) Using Abaqus}  
        \label{fig:edgeCrackVon-Mises}  
    \end{minipage}  
\end{figure}

Figs. \ref{fig:edgeCrackDisp} and \ref{fig:edgeCrackVon-Mises} illustrate the displacement and von-Mises stress contours derived from the X-PINN method, compared with the results from Abaqus software. The findings from the X-PINN method align closely with those from Abaqus, underscoring the robustness and reliability of the X-PINN technique.

\begin{figure}[htbp]  
    \centering  
    \includegraphics[scale=0.4]{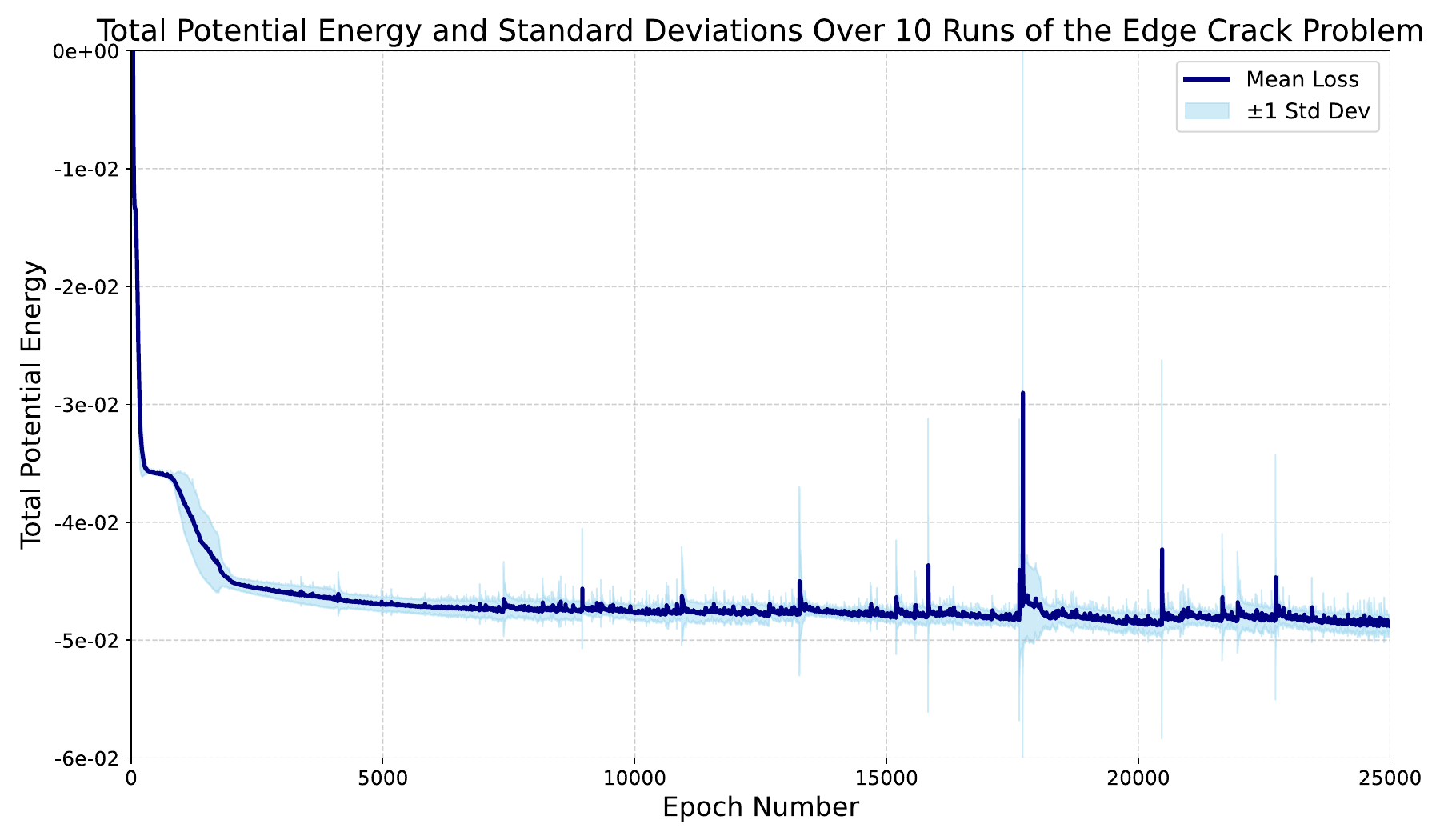}  
    \caption{Total potential energy over 10 runs of the Edge Crack problem}  
    \label{fig:lossHistoryEdgeCrack}  
\end{figure}

Fig.~\hyperref[fig:lossHistoryEdgeCrack]{\ref{fig:lossHistoryEdgeCrack}} shows the total potential energy over 10 runs of the Edge Crack problem across 25,000 training epochs. The average energy drops sharply at first, then levels off as training progresses, showing good convergence. The shaded region shows the variation between runs, with occasional spikes but mostly stable behavior. This suggests the model learns consistently even with some random variation across runs.

\subsection{2D Multiple Cracked Plate}
\begin{figure}[htbp]
    \centering
\includegraphics[scale=0.25]{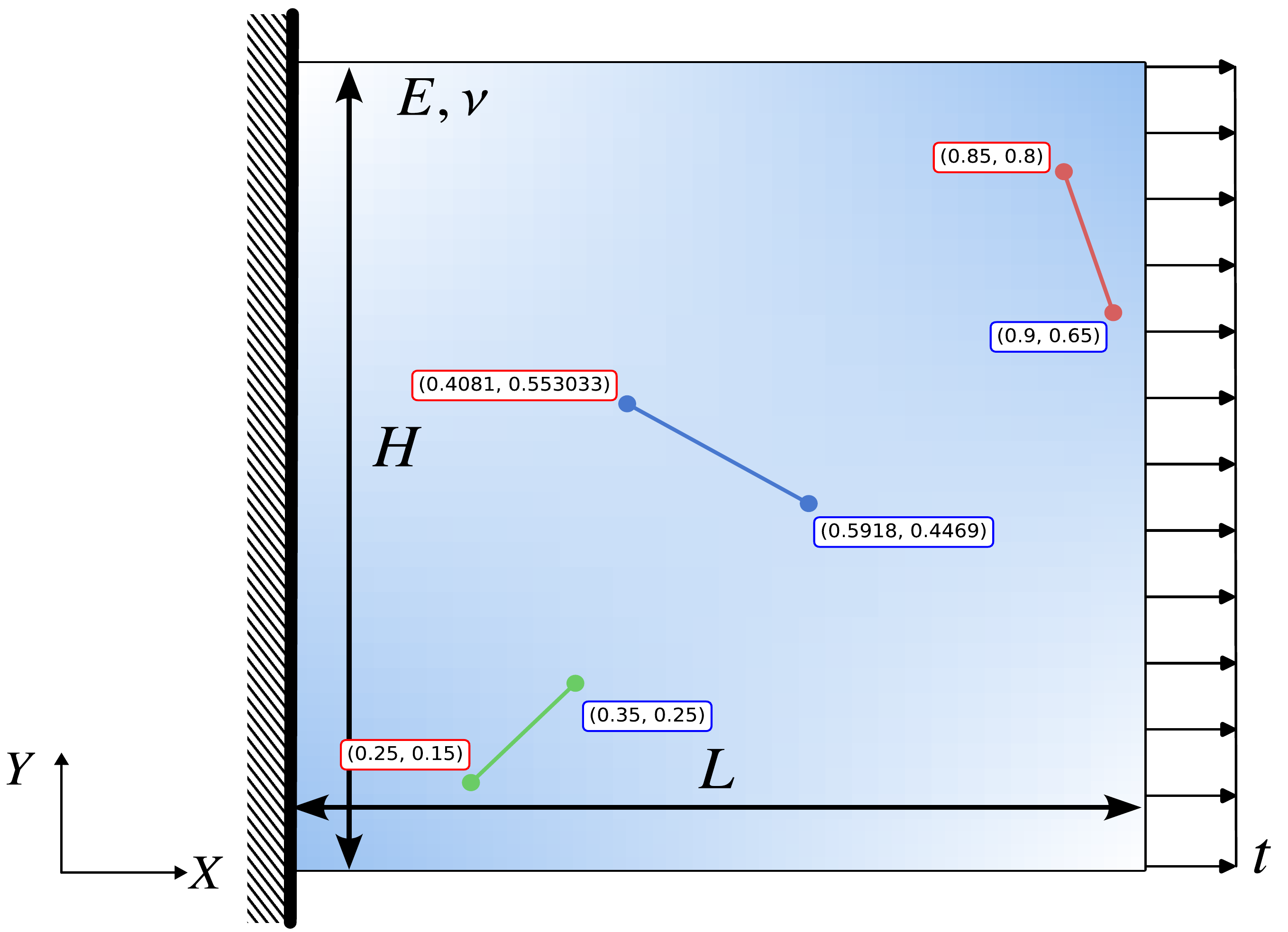}
    \caption{Geometry and Boundary Conditions for the 2D Plate with Multiple Cracks.}
    \label{fig:multipleCracks}
\end{figure}

In the final example, a 2D plate with multiple cracks is analyzed using X-PINN. The geometry and boundary conditions are illustrated in Fig.~\hyperref[fig:multipleCracks]{\ref{fig:multipleCracks}}. The figure indicates the locations of the cracks by their tip positions. For this analysis also, the dimensions of the plate, \( L \) and \( H \), as well as the modulus of elasticity, \( E \), are all set to 1. Additionally, Poisson's ratio, \( \nu \), is defined as 0.3, and the applied traction, \( t \), is specified as 0.1.

Both solution schemes~1 and ~2 (see Figs. \ref{fig:enrichedneuralNetwork} and \ref{fig:enrichedneuralNetwork_b}) are employed using 40{,}000 standard and 10{,}800 enrichment training points, respectively. Each scheme incorporates standard and enrichment neural networks: the standard networks consist of 10 layers with 20 neurons per layer, while the enrichment networks consist of 10 layers with 10 neurons per layer. As mentioned earlier, in scheme~1, a single neural network models the discontinuous component for all cracks. Conversely, scheme~2 allocates a separate enrichment network to each crack.

\begin{figure}[htbp]  
    \centering  
    \begin{minipage}{0.95\textwidth}  
        \centering  
        \begin{minipage}{0.49\linewidth}  
            \begin{minipage}{\linewidth}  
                \includegraphics[height=0.85\textwidth]{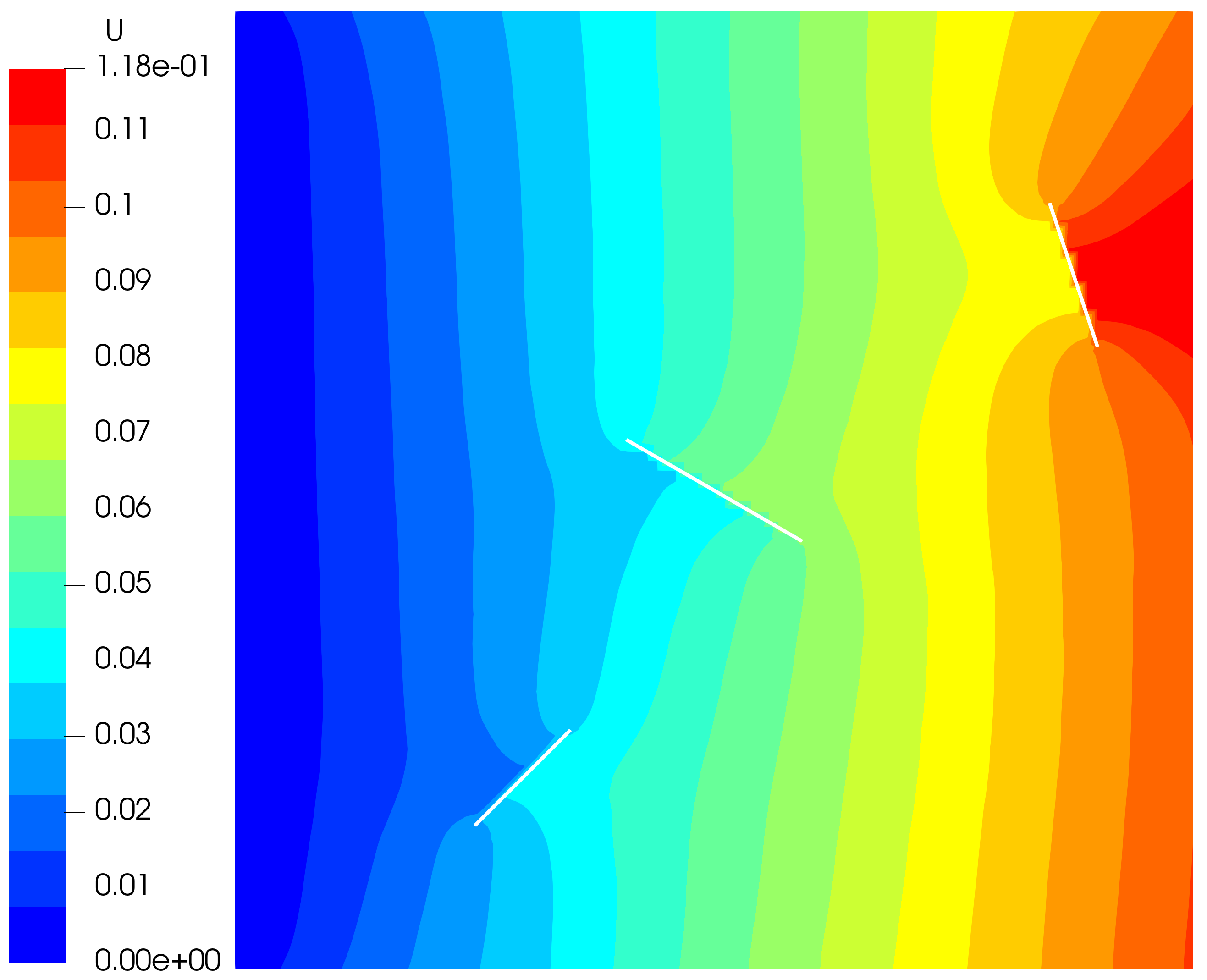}  
                \caption*{(a)}  
            \end{minipage}  
            \begin{minipage}{\linewidth}  
                \includegraphics[height=0.85\textwidth]{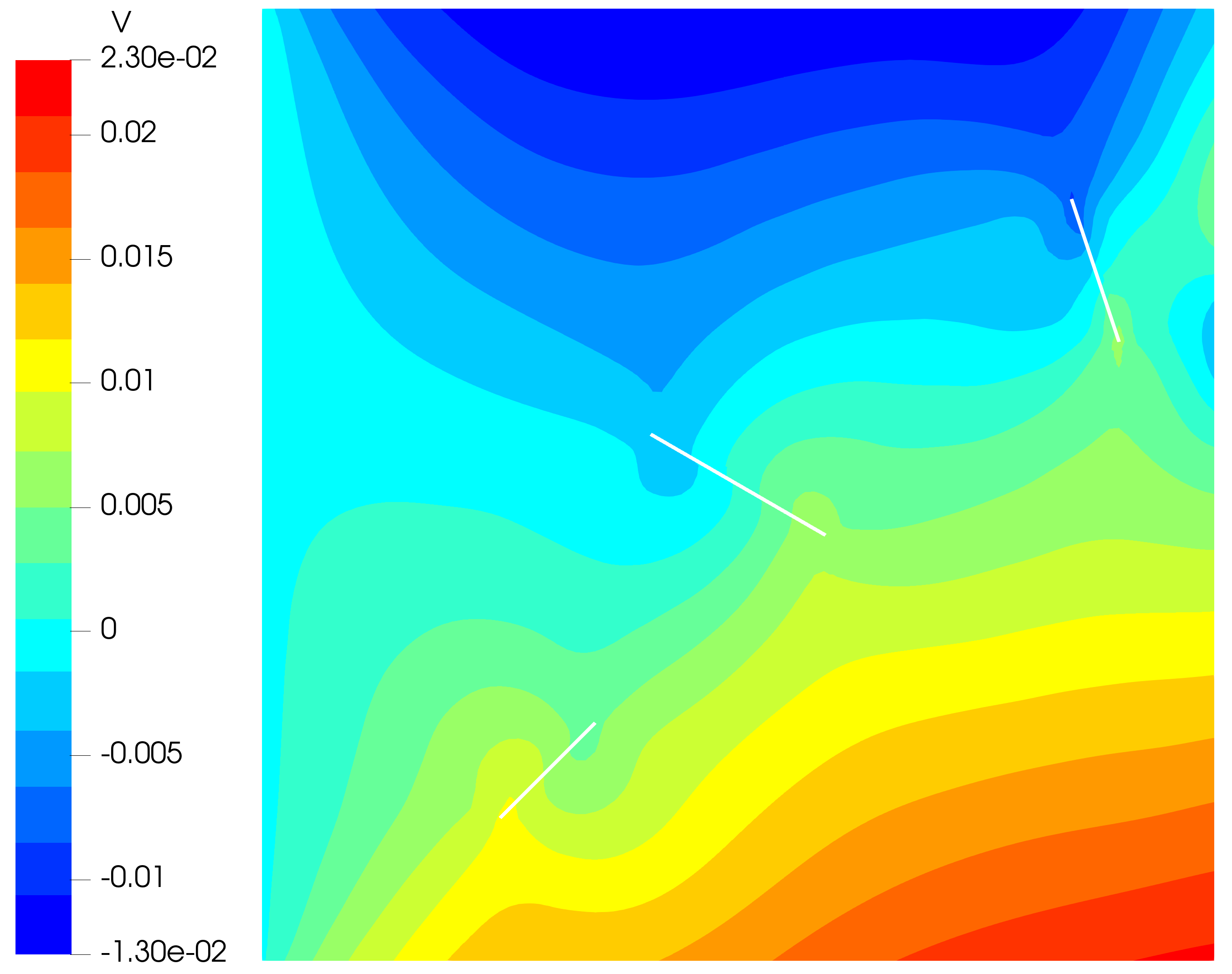}  
                \caption*{(c)}  
            \end{minipage}
        \end{minipage}  
        \begin{minipage}{0.49\linewidth}  
            \begin{minipage}{\linewidth}  
                \includegraphics[height=0.85\textwidth]{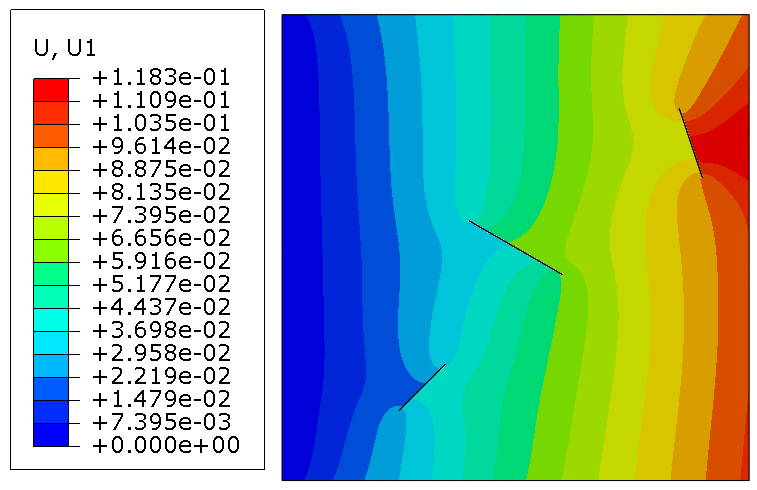}  
                \caption*{(b)}  
            \end{minipage}  
            \begin{minipage}{\linewidth}  
                \includegraphics[height=0.85\textwidth]{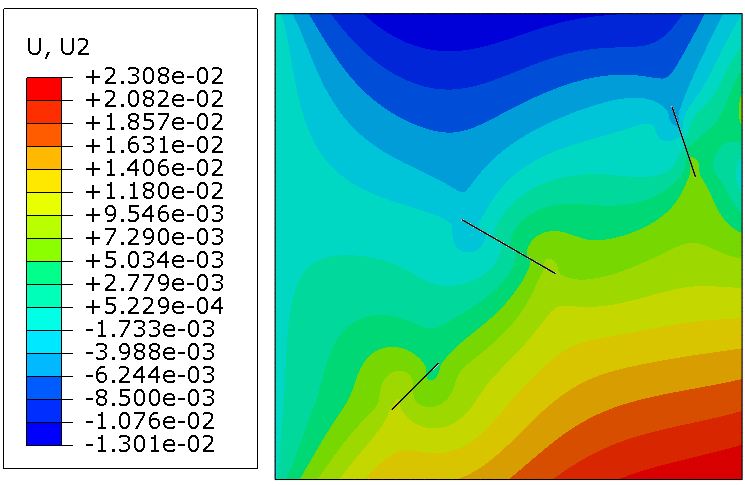}  
                \caption*{(d)}  
            \end{minipage}  
        \end{minipage}  
        \caption{Displacement Contours of the Multiple Crack problem: (a) U Using X-PINN (scheme ~1), (b) U Using Abaqus, (c) V Using X-PINN (scheme ~1), and (d) V Using Abaqus}  
        \label{fig:mutipleCrackDisp}  
    \end{minipage}  
\end{figure}

\begin{figure}[htbp]  
    \centering  
    \begin{minipage}{0.95\textwidth}  
        \centering  
        \begin{minipage}{0.49\linewidth}  
            \begin{minipage}{\linewidth}  
                \includegraphics[height=0.85\textwidth]{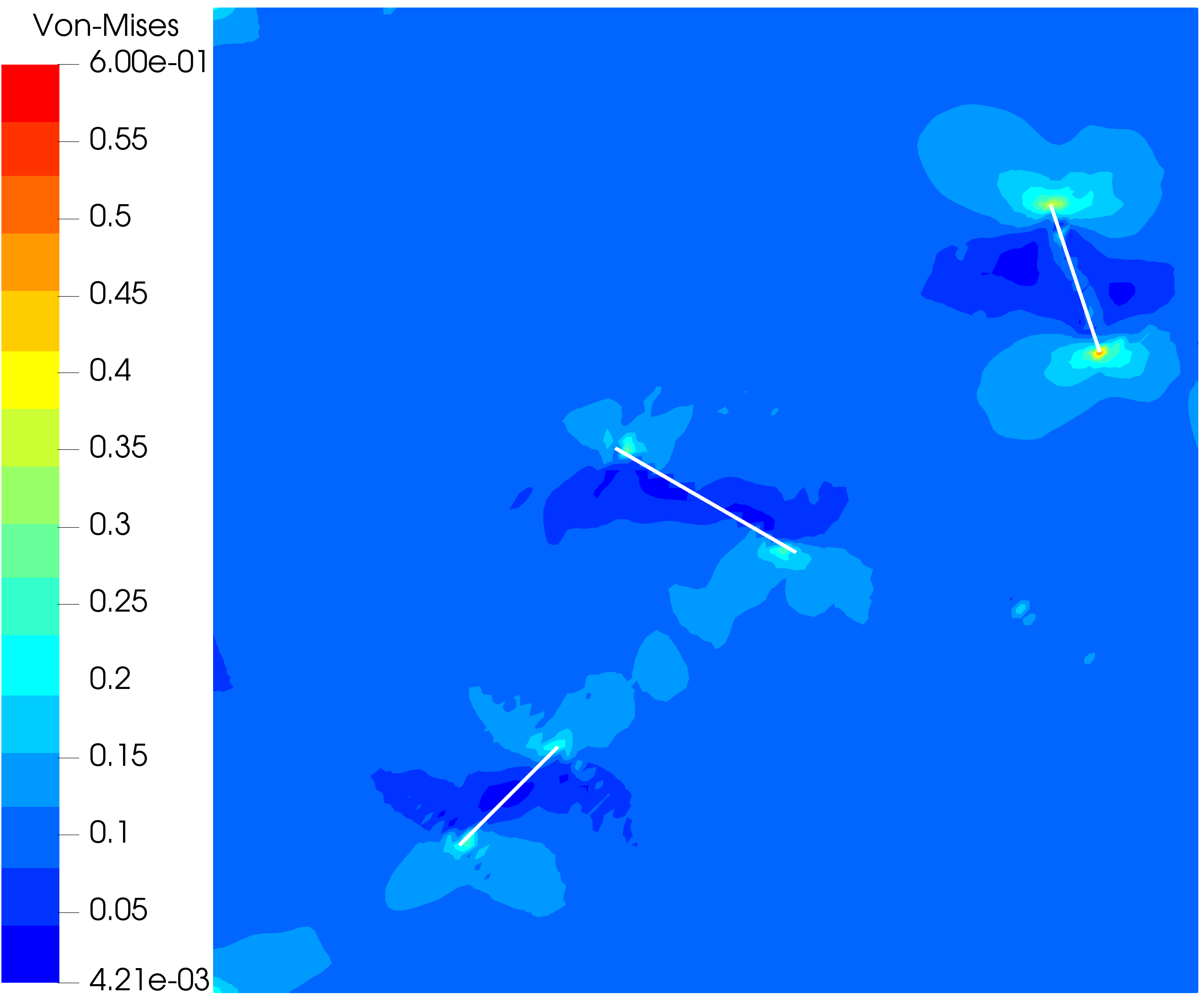}  
                \caption*{(a)}  
            \end{minipage}  
        \end{minipage}  
        \begin{minipage}{0.49\linewidth}  
            \begin{minipage}{\linewidth}  
                \includegraphics[height=0.85\textwidth]{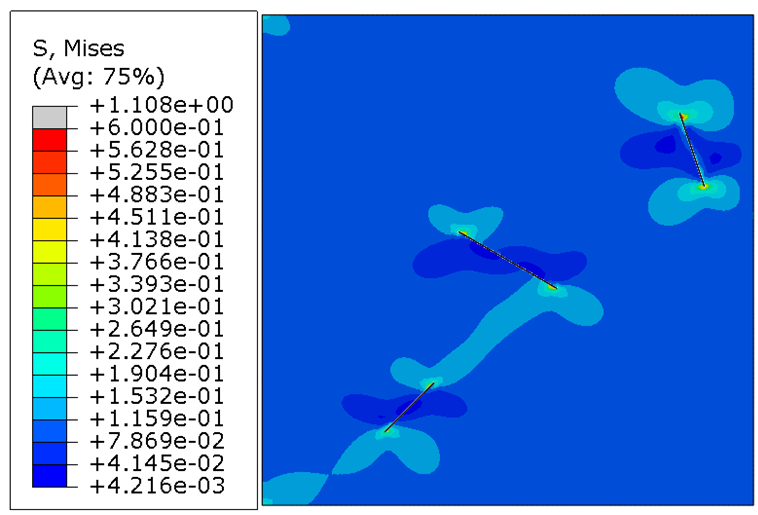} 
                \caption*{(b)}  
            \end{minipage}   
        \end{minipage}  
        \caption{von-Mises stress contour  of the Multiple Crack problem: (a) using X-PINN (scheme ~1), (b) using Abaqus}  
        \label{fig:multipleCrackVon-Mises}  
    \end{minipage}  
\end{figure}

\begin{figure}[htbp]  
    \centering  
    \begin{minipage}{0.95\textwidth}  
        \centering  
        \begin{minipage}{0.49\linewidth}  
            \begin{minipage}{\linewidth}  
                \includegraphics[height=0.85\textwidth]{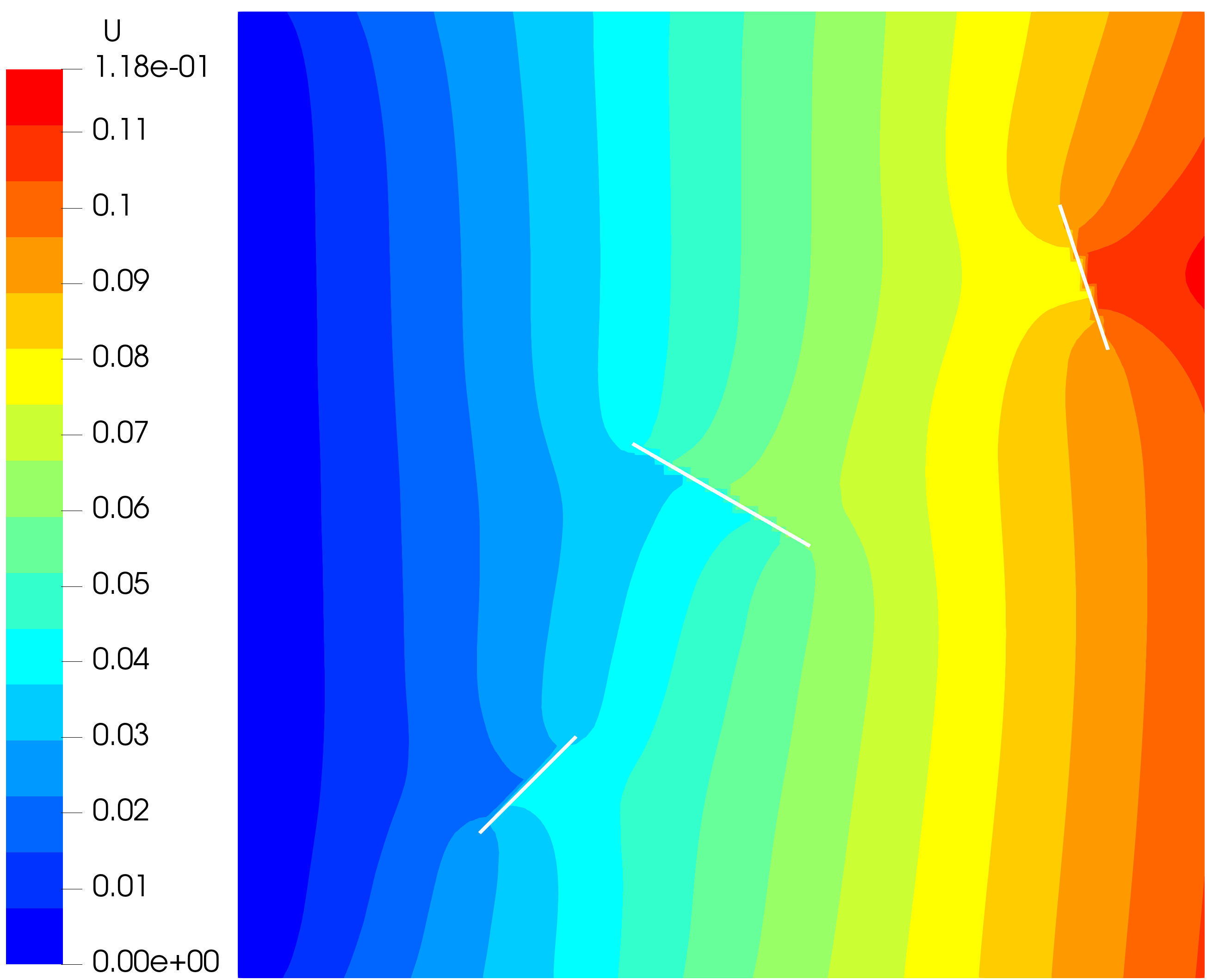}  
                \caption*{(a)}  
            \end{minipage}  
            \begin{minipage}{\linewidth}  
                \includegraphics[height=0.85\textwidth]{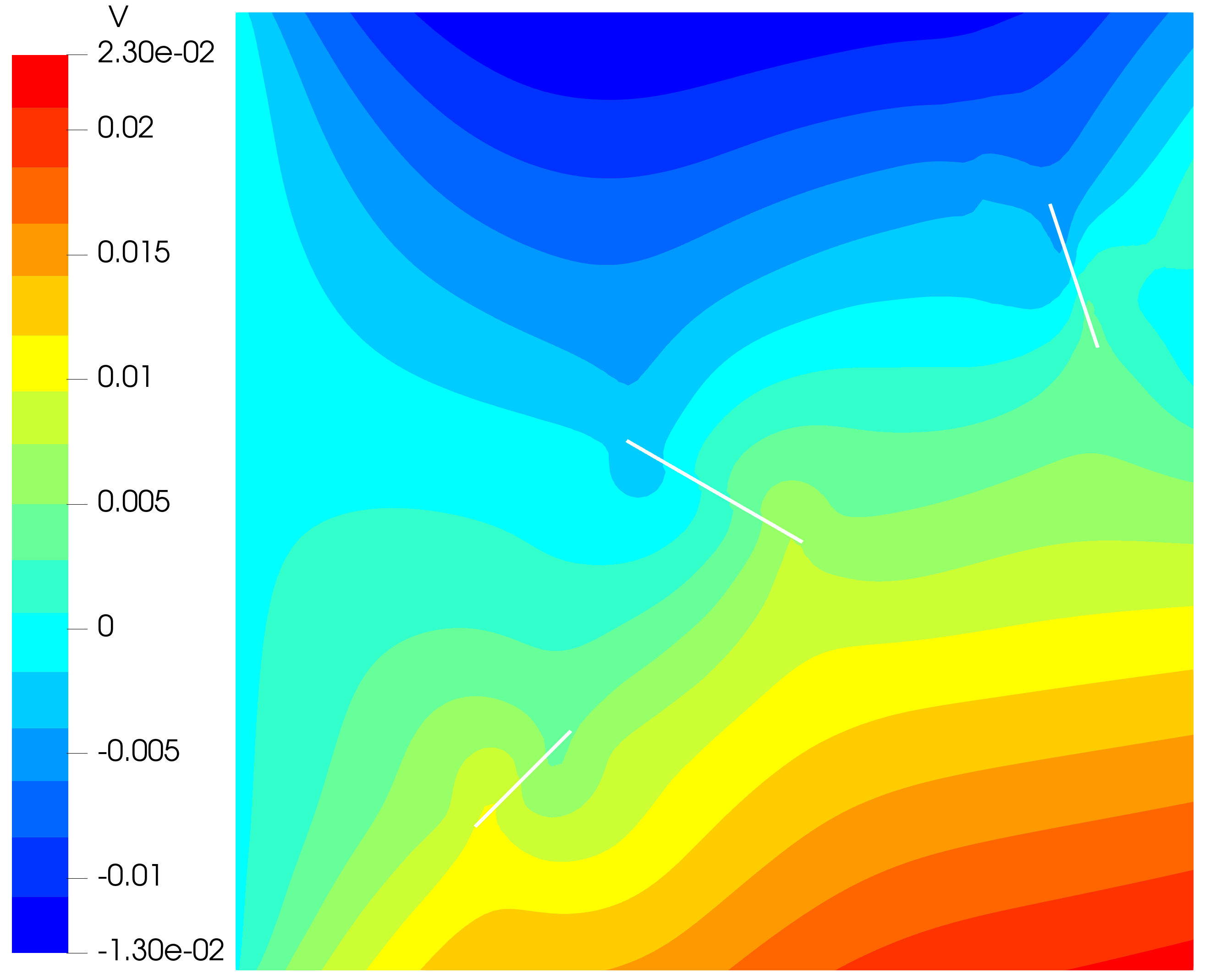}  
                \caption*{(c)}  
            \end{minipage}
        \end{minipage}  
        \begin{minipage}{0.49\linewidth}  
            \begin{minipage}{\linewidth}  
                \includegraphics[height=0.85\textwidth]{Figures/UMultiple_Abq.png}  
                \caption*{(b)}  
            \end{minipage}  
            \begin{minipage}{\linewidth}  
                \includegraphics[height=0.85\textwidth]{Figures/VMultiple_Abq.png}  
                \caption*{(d)}  
            \end{minipage}  
        \end{minipage}  
        \caption{Displacement contours of the Multiple Crack problem: (a) U using X-PINN (scheme ~2), (b) U using Abaqus, (c) V using X-PINN (scheme 2), and (d) V using Abaqus}  
        \label{fig:mutipleCrackDisp_2}  
    \end{minipage}  
\end{figure}

\begin{figure}[htbp]  
    \centering  
    \begin{minipage}{0.95\textwidth}  
        \centering  
        \begin{minipage}{0.49\linewidth}  
            \begin{minipage}{\linewidth}  
                \includegraphics[height=0.85\textwidth]{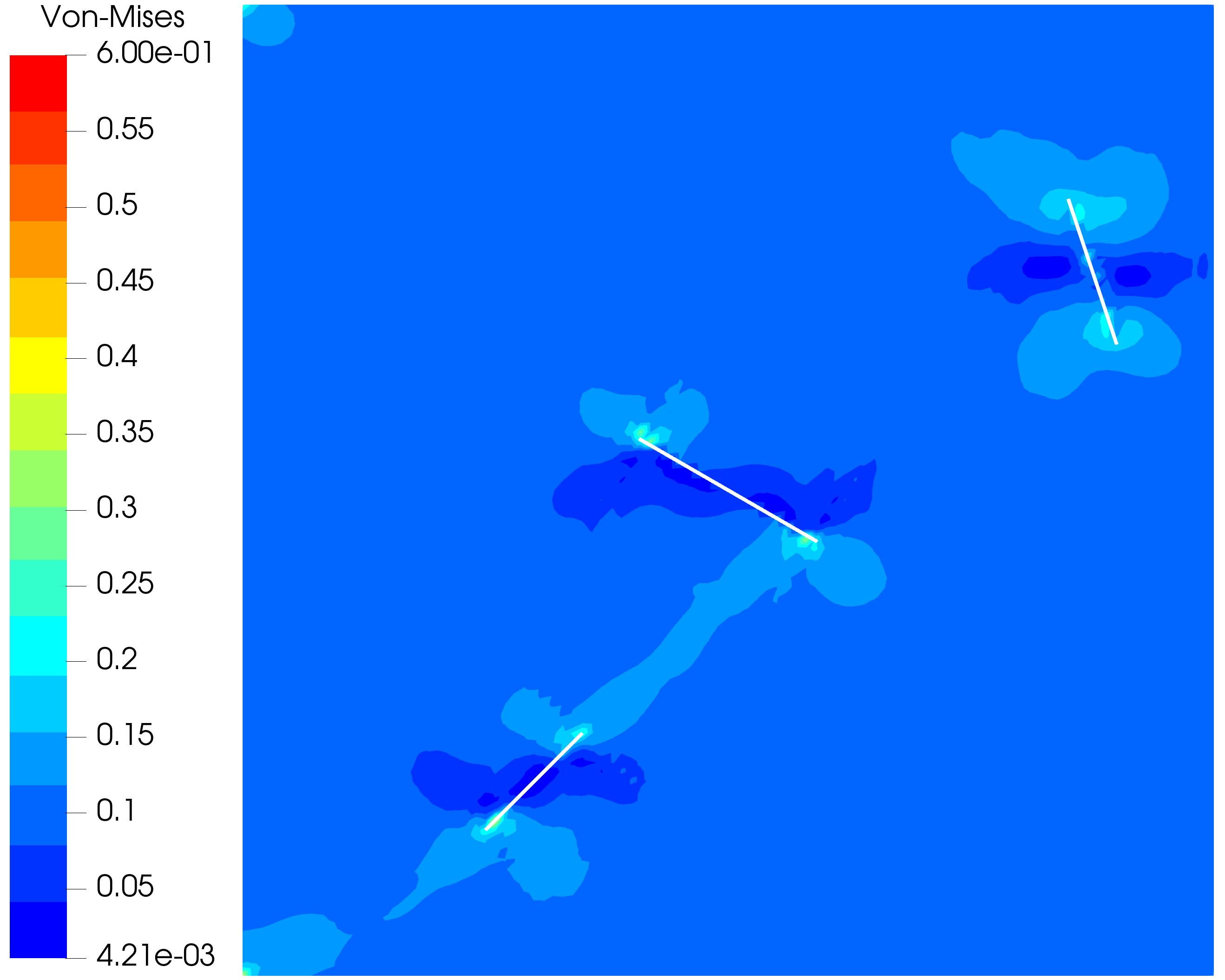}  
                \caption*{(a)}  
            \end{minipage}  
        \end{minipage}  
        \begin{minipage}{0.49\linewidth}  
            \begin{minipage}{\linewidth}  
                \includegraphics[height=0.85\textwidth]{Figures/VonMultiple_Abq.png} 
                \caption*{(b)}  
            \end{minipage}   
        \end{minipage}  
        \caption{von-Mises stress of Multiple Crack problem: (a) Using X-PINN (scheme 2), (b) Using Abaqus}  
        \label{fig:multipleCrackVon-Mises_2}  
    \end{minipage}  
\end{figure}

\begin{figure}[htbp]  
    \centering  
    \includegraphics[scale=0.6]{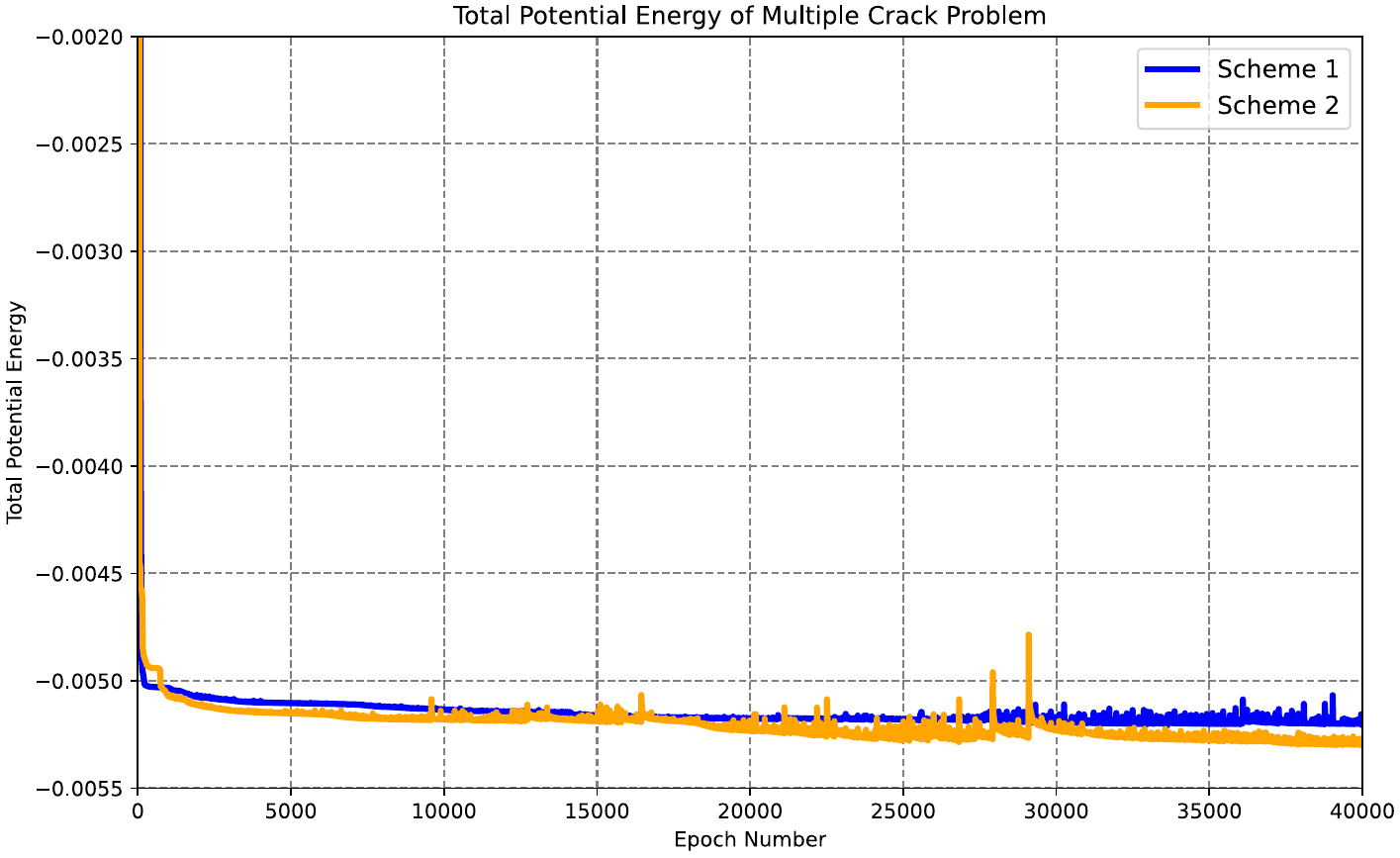}  
    \caption{Total potential energy history for Multiple Crack Problems using the two proposed solution schemes} 
    \label{fig:lossHistorymultipleCrack}  
\end{figure}

The displacement contours generated by schemes~1 and ~2 are shown in Fig.~\hyperref[fig:mutipleCrackDisp]{\ref{fig:mutipleCrackDisp}} and Fig.~\hyperref[fig:multipleCrackVon-Mises]{\ref{fig:mutipleCrackDisp_2}}, respectively, while the von Mises stress contours produced by each scheme is presented in Fig.~\hyperref[fig:multipleCrackVon-Mises]{\ref{fig:multipleCrackVon-Mises}} and Fig.~\hyperref[fig:multipleCrackVon-Mises_2]{\ref{fig:multipleCrackVon-Mises_2}}. The potential energy convergence diagrams for both schemes also also shown in Fig.~\hyperref[fig:lossHistorymultipleCrack]{\ref{fig:lossHistorymultipleCrack}}.

As can be seen, the displacements obtained by both schemes closely align with the Abaqus results, and their convergence behaviors are nearly identical. However, the von Mises stress computed by scheme 2 shows better compatibility with the Abaqus results. The potential energy convergence diagrams also indicate that scheme 2 results in lower potential energy and thus less error. However, the computational cost of the scheme ~1 is considrably lower than scheme ~2 which makes it attractive in multiple crack problems. 

\section{Conclusions} \label{summary}
This study presents the eXtended Physics-Informed Neural Network (X-PINN), a novel and robust framework for addressing fracture mechanics problems involving multiple cracks. Traditional PINNs face challenges in handling the discontinuities and singularities characteristic of fracture behavior. Drawing inspiration from the extended finite element method (XFEM), X-PINN overcomes these limitations through a hybrid neural architecture and an enriched solution strategy.

The proposed framework utilizes an energy-based loss function to enhance physical fidelity and decomposes the solution into continuous and discontinuous components, each modeled by specialized neural networks. To facilitate accurate and efficient integration, the Cartesian Transformation Method (CTM) is employed, offering a simple yet effective approach recommended for broader application within PINNs.

Crack-induced discontinuities are addressed through the use of enrichment functions, such as Heaviside and sawtooth-like functions. While this work emphasizes discontinuity enrichment, the framework is general and adaptable to various enrichment types. A domain partitioning strategy is introduced to accommodate distinct training points across cracked regions, and two training schemes are proposed to govern the learning process of the enrichment networks.

The effectiveness of X-PINN is demonstrated through validation on 1D and 2D benchmark problems involving single and multiple cracks. Both training schemes show promising results, with scheme 2 providing higher accuracy and scheme 1 offering reduced computational cost.

In conclusion, X-PINN emerges as a flexible, mesh-free, and physically grounded extension of traditional PINNs, well-suited for modeling complex fracture problems. Its adaptability and performance indicate strong potential for future extensions to 3D problems, dynamic fracture scenarios, and integration with advanced enrichment techniques.

\appendix
\section{2D Enrichment Function Constants}
The 2D enrichment function \(\mathbb{D}(\xi, \eta) \) is defined as:

\begin{equation}
    \mathbb{D}(\xi, \eta) = \Xi(\xi) \cdot \Lambda(\eta) \tag{\ref{2DEnrichment}, Repeated}
\end{equation}

\noindent where the component \(\Xi(\xi)\) is defined as:

\begin{equation}
    \Xi(\xi) =
    \begin{cases}
        a_0 + a_1 \xi + a_2 \xi^2 + a_3 \xi^3, & \text{if } \xi_1 \leq \xi \leq \xi_2 \\[6pt]
        0, & \text{otherwise} \tag{\ref{Xi}, Repeated}
    \end{cases}
\end{equation}

\noindent For \( \xi_1 \leq \xi \leq \frac{\xi_1 + \xi_2}{2} \):
\begin{subequations}
\begin{equation}
    a_0 = -\frac{4(\xi_1^3 + 3\xi_2 \xi_1^2)}{(\xi_1 - \xi_2)(\xi_1^2 - 2\xi_1 \xi_2 + \xi_2^2)}
\end{equation}

\begin{equation}
    a_1 = \frac{24(\xi_1^2 + \xi_2 \xi_1)}{(\xi_1 - \xi_2)(\xi_1^2 - 2\xi_1 \xi_2 + \xi_2^2)}
\end{equation}

\begin{equation}
    a_2 = -\frac{12(3\xi_1 + \xi_2)}{(\xi_1 - \xi_2)(\xi_1^2 - 2\xi_1 \xi_2 + \xi_2^2)}
\end{equation}

\begin{equation}
    a_3 = \frac{16}{(\xi_1 - \xi_2)(\xi_1^2 - 2\xi_1 \xi_2 + \xi_2^2)}
\end{equation}
\end{subequations}
\noindent For \( \frac{\xi_1 + \xi_2}{2} \leq \xi \leq \xi_2 \):

\begin{subequations}
\begin{equation}
    a_0 = \frac{4(\xi_2^3 + 3\xi_1 \xi_2^2)}{(\xi_1 - \xi_2)(\xi_1^2 - 2\xi_1 \xi_2 + \xi_2^2)}
\end{equation}

\begin{equation}
    a_1 = -\frac{24(\xi_2^2 + \xi_1 \xi_2)}{(\xi_1 - \xi_2)(\xi_1^2 - 2\xi_1 \xi_2 + \xi_2^2)}
\end{equation}

\begin{equation}
    a_2 = \frac{12(\xi_1 + 3\xi_2)}{(\xi_1 - \xi_2)(\xi_1^2 - 2\xi_1 \xi_2 + \xi_2^2)}
\end{equation}

\begin{equation}
    a_3 = -\frac{16}{(\xi_1 - \xi_2)(\xi_1^2 - 2\xi_1 \xi_2 + \xi_2^2)}
\end{equation}
\end{subequations}
\noindent In addition, the component \(\Lambda(\eta)\) is defined as:

\begin{equation}
    \Lambda(\eta) =
    \begin{cases}
        b_0 + b_1 \eta + b_2 \eta^2, & \text{if } \eta_1 - l_0 \leq \eta \leq \eta_1 + l_0 \\[6pt]
        0, & \text{otherwise}
    \end{cases}\tag{\ref{Lambda}, Repeated}
\end{equation}

\noindent  For \( \eta_1  \leq \eta \leq \eta_1 + l_0 \):

\begin{equation}
    b_0 = \frac{l_0^2 + 2l_0 \eta_1 + \eta_1^2}{l_0^2}, \quad
    b_1 = -\frac{2(l_0 + \eta_1)}{l_0^2}, \quad
    b_2 = \frac{1}{l_0^2}
\end{equation}

\noindent And, for \( \eta_1 - l_0 \leq \eta \leq \eta_1 \):

\begin{equation}
    b_0 = -\frac{l_0^2 - 2l_0 \eta_1 + \eta_1^2}{l_0^2}, \quad
    b_1 = -\frac{2(l_0 - \eta_1)}{l_0^2}, \quad
    b_2 = -\frac{1}{l_0^2}
\end{equation} \label{app1}

\bibliographystyle{elsarticle-num}
\bibliography{Bibliography}
\end{document}